%% file: iclr2026_conference.tex
\documentclass{article} % For LaTeX2e
\PassOptionsToPackage{numbers, compress}{natbib}
\PassOptionsToPackage{dvipsnames,svgnames,x11names,HTML}{xcolor}

\usepackage{natbib}
\usepackage{times}
\usepackage{iclr2026_conference}

\input{math_commands.tex}

\input{config}
\input{macro}

\title{RuleReasoner: Reinforced Rule-based Reasoning via Domain-aware Dynamic Sampling}

\author{Yang Liu\coauth{}~~~~Jiaqi Li\coauth{}~~~~Zilong Zheng\correspond{} \\
State Key Laboratory of General Artificial Intelligence, BIGAI \\
\texttt{\{liuyang,lijiaqi,zlzheng\}@bigai.ai} \\
}

\iclrfinalcopy % Uncomment for camera-ready version, but NOT for submission.
\begin{document}

\maketitle

\begin{abstract}
Rule-based reasoning is acknowledged as one of the fundamental problems of reasoning. While recent studies show that large reasoning models (LRMs) have remarkable reasoning capabilities enhanced by reinforcement learning (RL), real applications still face severe challenges due to variations in rule formats, types, and complexity.\iffalse It remains an open question \textit{whether models can perform rule-based reasoning effectively with robust generalization across diverse tasks and domains}.\fi\ To mitigate this issue, we introduce \textbf{\ourmethod}, an effective method for rule-based reasoning via a wide collection of curated tasks and a novel domain-aware dynamic sampling approach in RL. Specifically, \ourmethod resamples each training batch by updating the domain weights based on historical rewards. This facilitates domain balance and active learning schedules for RL, obviating static mix-training engineered by human. Evaluations on in-distribution (ID) and out-of-distribution (OOD) benchmarks reveal that \ourmethod outperforms frontier LRMs by a significant margin (\textbf{$\Delta$4.1\%} on eight ID tasks and \textbf{$\Delta$10.4\%} on three OOD tasks over OpenAI-o1). Notably, our approach also exhibits higher computational efficiency compared to prior methods. \unmarkedfootnote{\coauth Equal contribution. \correspond Corresponding author.}
\end{abstract}

\input{figure/ood_performance}

\section{Introduction}

Rule-based reasoning~\citep{xu2024faithful,wang2024can,servantez2024chain,morishita2024enhancing,wang2024symbolic} is an ability to draw new conclusions or make decisions based on provided facts and predefined logical rules, which requires a strong ability of reasoning. It emulates human reasoning and mirrors the structured deductive processes that humans employ in domains such as law, mathematics, and medical diagnostics~\citep{liu2023benchmarking, xiong2024benchmarking}. The need for rule-based reasoning increasingly grows in scenarios that require transparency, explainability, and adherence to domain constraints. Moreover, deviations from rules for different scenarios lead to significant changes in reasoning process, which requires more controllable and adaptable reasoning capabilities under ever-changing circumstances~\citep{saparov2023testing, tang2023icsr}.

Recent work has demonstrated the remarkable reasoning capabilities of large reasoning models (LRMs) with an intermediate thinking process, chain-of-thought (CoT)~\citep{wei2022chain}, notably the long thinking ability stimulated by reinforcement learning (RL)~\citep{xie2025logic, meng2025mm, guo2025deepseek, hu2503open}. However, conventional approaches rely closely on scaling to larger model sizes or supervision distilled from superior models. What's more, as the contextual window expands, language models encounter difficulties in attending to the relevant information required for task completion, experiencing issues such as the \textit{``lost in the middle''} \citep{liu-etal-2024-lost,NEURIPS2024_71c3451f}, exhibiting relatively weak instruction following abilities to understand and associate rules and facts provided in the context~\citep{srivastava2025towards, li2025small, guan2025rstar}. %In this work, we explore whether the capability of rule-based reasoning for language models can be learnt and enhanced effectively. Furthermore, we are curious whether this ability can be generalized across various tasks and reasoning paradigms with sufficient explainability.

With this in mind, we aim at investigating \textbf{whether it is feasible and effective to enhance the rule-based reasoning ability of language models}. We also want to validate that this improved capability can generalize to unseen reasoning tasks, forms, and difficulties. Therefore, we propose \textbf{\ourmethod}, which trains language models to be rule-based reasoners. It performs on-policy RL and mitigates its limitations of sample inefficiency and the application of rigid rules in dynamic contexts. Specifically, \ourmethod leverages task reward to estimate domain sampling weights without requiring human prior knowledge or high computational costs from repeated rollouts. First, \ourmethod initializes to train a model in a standard Reinforcement Learning with Verifiable Rewards (RLVR) way. Second, within a training iteration, \ourmethod updates the domain weights based on historical rewards of previous training steps. Finally, \ourmethod samples a training batch based on the domain weights to perform policy optimization.

\textbf{Our innovations} are summarized as three folds:
\begin{itemize}[leftmargin=*,topsep=0pt]
    \item \textbf{Novel Rule-centric Dataset:} We release a large and diverse dataset, \ourdata, which spans eight rule-based reasoning tasks with explicit or implicit contextual rules tied to each question. These rules vary in format, reasoning forms, and complexity, allowing training and ID/OOD evaluation in generalizable rule application rather than memorization. 
    \item \textbf{RLVR for Rule-based Reasoning:} We design a RLVR framework that introduces useful training regularization to achieve stable RL training dynamics on complex rules, even with model size under 8B. It encourages exploring and exploiting valid reasoning steps instead of imitation, improving generalization to unseen rules in the \ourdata.
    \item \textbf{Domain-aware Dynamic Sampling:} To harmonize language models' proficiencies across imbalanced domains, we present an adaptive sampling algorithm that dynamically reweights training domains based on their degree of under-optimization. This ensures balanced learning dynamics across tasks, enhancing both in-distribution (ID) and out-of-distribution (OOD) performance.%This method is intuitively compatible with a variety of RLVR algorithms for enhancement. % with varying rule complexities and mitigates catastrophic forgetting.
\end{itemize}

We perform extensive evaluations on \ourmethod, introducing two best rule-based reasoning models: \ourmodelbest and \ourmodel. Empirical results show that: (1) As depicted in \cref{fig:ood-results}, \ourmodelbest outperforms OpenAI-o1, Claude-3.7-Sonnet, and DeekSeek-R1, achieving higher performance than trivial RLVR methods. Specifically, \ourmodelbest achieves $\Delta$14\% and $\Delta$49\%  OOD pass@1 respectively over o1 and the base model. (2) \ourmodel further demonstrates that language models can effectively learn rules even with a smaller model size, achieving an average pass@1 of 78.3\% on three OOD benchmarks ($\Delta$7.3\% over o1). (3) Surprisingly, \ourmethod achieves comparable task performance with notably fewer training steps than existing RLVR methods. This suggests that \ourmethod not only enhances ID and OOD performance but also improves training efficiency.

\section{Related Work and Preliminaries}
%\rnote{Introduce related work in this paragraph and highlight current gap in the field of RL for Reasoning, such as sampling efficiency issue (DAPO's dynamic sampling cause high compute cost and SRPO's history sampling only focus on solution-level and conduct resampling in (coarse-grain) epoch level, and the inefficient trials in more domain coverage). \textbf{Hence, our sampling method, which is very efficient to compute, becomes naturally necessary in this background.}}
RL plays a critical role in improving the reasoning capabilities of large language models (LLMs)~\citep{silver2025welcome}, particularly through approaches such as RLVR~\citep{guo2025deepseek, li2025limr, zuo2025ttrl}.
% Moreover, it has also been generally applied to multimodal models~\citep{tan2025reason, li2025videochat} and agentic applications~\citep{jin2503search, zhao2025absolutezeroreinforcedselfplay, wang2025ragen}. 
In this section, we introduce the task definition of rule-based reasoning (\S\ref{subsec:preliminary}) and discuss key components of prior RLVR methods (\S\ref{subsec:online_rl}) and their limitations.

% To address the issues, we design a cognitive domain-aware dynamic sampling as an active online learning policy ($\S$\ref{subsec: dads}) and correspondingly introduce the initial data curation for RL training ($\S$\ref{subsec: data_curation}).

\subsection{Related Work} It has been demonstrated that LLMs can learn rules, including the rule grounding and implementation with promising potential~\citep{zhu2023large, servantez-etal-2024-chain}. Early works start from predefined rules in symbolic forms with an emphasis on scalability and compositionality in specific tasks~\citep{tang2023large,luo2309chatrule,jia-etal-2024-combining,gui2024logicgame,he2025large} while recent works are dedicated to perform rule-based reasoning in natural language that are more applicable for real scenarios~\citep{zhou2024rulearena, he2024idea,tang2024mars}. It is also worth noting that recent advances in logical reasoning, such as Logic-RL~\citep{xie2025logic}, are generally considered rule-free for reasoning, which differs from our task definition. % Unlike common logical reasoning, which is typically regarded as rule-free for reasoning \citep{xie2025logic},
% When more advanced models are not available due to licence and cost issues, we are curious to see how well SLMs can leverage limited data to enhance its own rule-based logical reasoning capabilities. In addition, each step of rule application can be typically decoupled into two different processes, 
These methods explore the potential of rule learning through diverse prompting methods~\citep{diallo2025rule, peng2024regenesis}, supervised distillation~\citep{wang-etal-2024-llms}, and external memory augmentation~\citep{wang-etal-2024-symbolic, wang2024symbolic}. However, they spend less effort adapting the reasoning capability of LLMs to unseen tasks with limited task types and formats. Inspired by recent advancements in RLVR methods focused on mathematical reasoning and code generation~\citep{2025arXiv250517508Z,chen2025acereason,wei2025swe,li2025reflectevo,zhao2025absolute,li2025seek}, we further optimize their limitation on data efficiency with dynamic data sampling along with a curated collection of diverse rule-centric training data. Our method improves model performance across both ID and OOD reasoning tasks with higher generalization and computational efficiency. Conventional reweighting and loss-based sampling methods \citep{luo-etal-2025-velocitune,shi-etal-2025-dids} are unsuitable for RLVR because the direction of optimization for surrogate loss of RL does not correlate with policy model performance. Moreover, compared with prior methods, we unifies domain-level dynamic sampling strategy into RL for reasoning without requiring proxy models to obtain domain weights \citep{xie2023doremi,liu2025regmix}, instead leveraging only the verifiable rewards in a self-adaptive paradigm.

\subsection{Preliminaries I: Rule-based Reasoning} \label{subsec:preliminary}

Given a question, a set of \textbf{facts}, and associated \textbf{rules} as context, the model is asked to answer the question by applying and reasoning the facts with rules. In this paper, we refer to the rules as \textit{contextual logic rules}, which are expressed in natural language and given specifically for each question. The rules provided can be generated either explicitly or implicitly as principles or premises to solve the question. %Based on the above definition, the final objective of rule-based logical reasoning is to systematically derive conclusions or decisions that are valid, consistent, and reliable by reasoning a structured set of predefined rules and facts. 
For a grouped datasets $\mathcal{D} := \{(d_i, q_i, r_i, y_i)\}_{i=1}^{n}$, where $d_i \in \{d_1, \cdots, d_n\}$ is a specific domain, $q_i$ is a question, $r_i$ is a reasoning trajectory, and $y_i$ is a verifiable answer. 

% Grouped data: $\mathcal{D} := \{(d_i, q_i, r_i, y_i)\}_{i=1}^{n}$
% consisting of various domains $\{d_1, d_2,...,d_n\}$

\subsection{Preliminaries II: On-policy Reinforcement Learning}
\label{subsec:online_rl}

%\paragraph{State.}
%\paragraph{Action.}
%\paragraph{Policy.}
\textbf{Reward Shaping.}
To teach models to learn reasoning, we design a rule-based exact match (EM) reward function to evaluate the response according to the final answer, ensuring both the correctness of the answers and the adherence to the format. We define $(q,\hat{y}) \sim \mathcal{D}$, $y \sim \pi_\theta(\cdot\lvert q)$, and % Specifically, for a correct answer, the reward is 1; for an incorrect or error-formatted answer, the reward is -1. Formally, this can be expressed as:
\begin{equation}
    \mathcal{R}_\text{EM}(\hat{y}, y) = \begin{cases} 1 & \operatorname{is\_equivalent} (\hat{y}, y) , \\ -1 & \operatorname{otherwise} . \end{cases}
    \label{eq:rule_based_verifier}
\end{equation}

%\paragraph{Critic.}
%\paragraph{Objective.}
\textbf{Policy Optimization.}
We adopt the basic form of GRPO~\citep{shao2024deepseekmath} but discard the part of KL loss, encouraging the model to explore various solutions. For each question-answer pair $(q, y)$, policy model $\pi_{\theta_\text{old}}$ samples a group of responses $\{y_1, y_2, \dots, y_G\}$ and calculates the associated rewards $\{r_1, r_2, \dots, r_G\}$, given the oracle answer $y$, using the aformentioned reward function $\mathcal{R}_\text{EM}$.
% \mathcal{J}_\text{GRPO}(\theta) & =\mathbb{E}_{(q,a) \sim \mathcal{D},\ \left\{o_i\right\}_{i=1}^G \sim \pi_{\theta_{\text {old}}}(\cdot \mid q)} \\
% & ~\frac{1}{G} \sum_{i=1}^G\left(\min \left(\frac{\pi_\theta\left(o_i \mid q\right)}{\pi_{\theta_{\text {old}}}\left(o_i \mid q\right)} A_i, \right.\right. \\
% & \left.\left.\operatorname{clip}\left(\frac{\pi_\theta\left(o_i \mid q\right)}{\pi_{\theta_\text{old}}\left(o_i \mid q\right)}, 1-\varepsilon, 1+\varepsilon\right) A_i\right)\right),

\begin{small}
\begin{equation}
\begin{aligned}
% \mathcal{J}_\text{GRPO}(\theta) & =\mathbb{E}_{(q,a) \sim \mathcal{D},\ \left\{o_i\right\}_{i=1}^G \sim \pi_{\theta_{\text {old}}}(\cdot \mid q)} \\
% & ~ \left[ \frac{1}{G} \sum_{i=1}^G \frac{1}{\lvert o_i \rvert} \sum_{t=1}^{\lvert o_i \rvert} \left( \min \left( \frac{\pi_\theta\left(o_{i,t} \mid q, o_{i,<t}\right)}{\pi_{\theta_{\text {old}}}\left(o_{i,t} \mid q, o_{i,<t}\right)} A_i, \right. \right. \right.
% \\ & \left. \left. \left. \operatorname{clip}\left(\frac{\pi_\theta\left(o_{i,t} \mid q, o_{i,<t}\right)}{\pi_{\theta_{\text {old}}}\left(o_{i,t} \mid q, o_{i,<t}\right)}, 1-\varepsilon, 1+\varepsilon\right) A_i \right) \right) \right] ,
\mathcal{J}(\theta) =\ &\mathbb{E}_{(q, y) \sim \mathcal{D},\ \left\{y_i\right\}_{i=1}^G \sim \pi_{\theta_{\text{old}}}(\cdot \mid q)} \\& \Bigg[ \frac{1}{G}\sum_{i=1}^{G} \frac{1}{|y_i|}\sum_{t=1}^{|y_i|} \Bigg( 
\min \Big( r_{i,t}(\theta) A_{i,t}, \operatorname{clip} \Big( r_{i,t}(\theta), 1 - \varepsilon, 1 + \varepsilon \Big) A_{i,t} \Big) 
\Bigg) \Bigg] ,
\end{aligned}
\end{equation}
\end{small}

where $r_{i,t}(\theta)$ is the rate of importance sampling for domain $d_i$ at the $t$-th token for $y_i$, and $A_i$ is the advantage as the critic obtained by normalizing the rewards within each group. We strictly follow the on-policy training method, performing only one gradient update after the policy model $\pi_{\theta_\text{old}}$ generates a group of $G$ rollouts, to enable stable RL training and prevent entropy collapse.

\begin{equation}
    r_{i,t}(\theta)=\frac{\pi_{\theta}(y_{i,t} \mid q, y_{i,<t})}{\pi_{\theta_{\text{old}}}(y_{i,t} \mid q,y_{i,<t})} ,\    A_i := \widetilde{r}_i
               = \frac{r_i-\operatorname{mean}\left(\left\{r_1, r_2, \cdots, r_G\right\}\right)}{\operatorname{std}\left(\left\{r_1, r_2, \cdots, r_G\right\}\right)} .
\end{equation}

\textbf{Limitations of RLVR on Training Efficiency.}  Though current RLVR elicits the long chain-of-thought reasoning ability based on the policy gradient RL algorithm like PPO~\citep{schulman2017proximal} and GRPO~\citep{shao2024deepseekmath}, the efficiency of training data for RLVR remains relatively unexplored. Existing works like DAPO~\citep{yu2025dapo} oversamples and filters out prompts with the accuracy equal to 1 and 0 to enhance training efficiency. However, it does not push the limits of training efficiency due the large recompute cost in the rollout stage. SRPO~\citep{zhang2025srpo} shows the gains via epoch-level re-sampling wtih RLVR without exploring the agile sampling methods for fine-grain control. Moreover, \textsc{AdaRFT}~\citep{shi2025efficient} explores an efficient batch-level sampling method using curriculum learning; however, it relies on human priors or an empirical success rate by models on sample difficulty. In the following sections, we expend great effort to further leverage training examples to achieve higher reasoning performance effectively.

\section{Domain-aware Policy Optimization with Dynamic Sampling}\label{sec:dapo}

% \yang{Revise, Reorg, and Simplify.}

\input{figure/training_recipe}

In data sampling, increasing the number of samples in an individual domain potentially harms the performance of other domains without timely control~\citep{albalak2024a} or causes obvious tradeoff across domains~\citep{xie2023doremi}.

To address similar issues in RLVR, as shown in Alg.~\ref{alg:algo1} and \cref{fig:training_recipe}, we propose \textbf{D}omain-\textbf{a}ware \textbf{D}ynamic \textbf{S}ampling (\ouralgo), an effective sampling method for RLVR aiming to improve the performance of a policy $\pi_\theta$ for solving multi-domain rule-based reasoning tasks. \ouralgo dynamically adjusts the probability of sampling data from different domains based on their 
historical rewards. By prioritizing domains 
that yield lower verifiable rewards or those lagging behind a target reward, \ouralgo enhances sample efficiency in training batch $\mathcal{B}_s$ and mitigates the domain imbalance issue, leading to faster and more stable learning 
of policies that satisfy reward specifications.  We instantiate \ourmethod with the gradient policy algorithm of GRPO variant in this work to demonstrate its effectiveness and efficiency.
% Specifically, \ouralgo iteratively estimates the average reward for each domain, calculates weights that favor high-potential domains, and then uses these weights to re-sample the training batch $\mathcal{B}_s$. We specifically instantiate \ourmethod with GRPO in this work to demonstrate its effectiveness and efficiency under the limitation of computational resources.
 
\subsection{Domain-aware Dynamic Sampling (\textsc{Dads})} \label{subsec: dads}

\input{table/domain_dynamic_sampling}

\paragraph{Domain-aware Rewards.}
At each training step $s$, to evaluate the proficiency for a domain $d_i \in \mathcal{D}$, we define $\bar{r}_{d_i,s}$ as the algebraic mean, calculated by domain rewards $\{r_{s,d_i,j}\}_{j=1}^m \sim \mathcal{R}_\text{EM}(y, \hat{y})\iffalse: \{r_{s,d_i,1}, r_{s,d_i,2}, \dots, r_{s,d_i,m}\}\fi$ of $m$ previous samples in the domain, which correspond to genersations $\mathcal{Y}_{s-1}: \{y_{s,d_i,j}\}_{j=1}^m \sim \pi_\theta(\cdot\lvert q)$ and the set of ground truth $\mathcal{\hat{Y}}_{s-1}$. Note that $m$ may vary across different domains and training steps due to the batch-level domain sampling strategy. Domain-aware rewards calculation over batch (Alg.~\ref{alg:algo1}, line 6 and 7) is computed as: $\bar{r}_{s,d_i} = \frac{1}{m}\sum_{j=1}^m r_{s,d_i,j}$.  We employ a target reward, $r_{\text{target}} \triangleq 1$, to define the upper bound for the underoptimization estimation, $v_{s,d_i}$, of a domain. Thus, we have the estimation: $1 - \widetilde{r}_{s, d_i}$, where target reward quantifies the extent to which a domain performance lags behind the desired level. 

% Furthermore, given that the utilisation of past $k$ rewards for domain $d_i$, we introduce a decayed importance sampling strategy measured by domain-specific rewards, which employs a historical reward buffer $\mathcal{Q_\text{history}}$ that is implemented by a queue in practice. The past $k$ rewards $\{\bar{r}_{d_i,j}\}_{j=t-k}^{t-1}$ are maintained through $\mathcal{Q_\text{history}}$ and eliminate the oldest one if $\mathcal{Q_\text{history}}$ reaches the maximum size.
\paragraph{Decaying Importance Sampling.} Furthermore, given the utilization of past rewards for domain $d_i$, we introduce a decaying importance-sampling strategy, which employs the exponentially weighted moving average~\citep{holt2004forecasting} that considers both current and the historical estimated rewards. The historical rewards $\{\widetilde{r}_{s-1,d_i}\}_{i=1}^n$ are involved with the smoothing factor $\alpha$ to produce normalized rewards $\{\widetilde{r}_{s,d_i}\}_{i=1}^n$.  We have $\widetilde{r}_{s,d_i} = \alpha \widetilde{r}_{s-1,d_i} + (1-\alpha)\bar{r}_{s,d_i}$, where $\alpha \in [0, 1]$ serves as a smoothing factor that creates a more stable estimate of the performance for a domain over time, rather than relying solely on the most recent reward $\bar{r}_{s,d_i}$. 

\paragraph{Domain Re-weighting.} Consequently, we establish a domain weight, $w_{s, d_i}$, which is then normalized using a standard softmax function (as detailed in Alg. \ref{alg:algo1}, lines 11 to 13). In this normalization, hyperparameters $\tau$ and $\epsilon$ are used: $\epsilon$ ensures a minimum sampling weight for all domains, even well-learned ones, and $\tau$ adjusts how strongly the sampling prioritizes domains based on their rewards. We discuss the sensitivity of $\alpha$, $\tau$, and $\epsilon$ in \S\ref{subsec:robustness_analyses}.
After obtaining the re-sampling weights $\mathcal{W}_s := \{w_1, w_2, \dots, w_n\}$ across domains, we use $\mathcal{W}_s$ to construct a new batch $\mathcal{B}_s$ for the subsequent policy optimization iteration. The process will be implemented iteratively during the training and more details are described in Algorithm \ref{alg:algo1}.

\subsection{Training Regularization}
We want to ensure that the model follows, matches, and implements the rules and does not just learn to identify specific datasets and perform correctly on them. To do this, we introduce different kinds of noise during RL training. This stops the models from recognizing particular datasets, recalling specific rules, or memorizing only the similar seen facts in the context.

\textbf{Disabling Entropy Bonus.} We discard the entropy bonus employed by \cite{schulman2017proximal} and \cite{shao2024deepseekmath}, to avoid the issue of entropy explosion in RL without the coldstart bootstrap.

\textbf{Discarding KL Divergence.} Similary to \cite{liu2025understanding}, we eliminate the KL term since the concerns on distributional shift of $\pi_\theta$ is eliminated by rule-based reward function defined in Eq. \ref{eq:rule_based_verifier}. This saves memory and computation in training while encouraging more exploration required by $\pi_{\theta}$.

\textbf{Rules Order Shuffling.} To prevent memorization of ordered rules appeared in \ourdata, the order of contextual logical rules are randomly shuffled for each training sample.

\subsection{\ourdata: Logical Rules Data Curation} \label{subsec: data_curation}
We follow the following \textbf{principles} to collect our training data, namely \ourdata.
\begin{itemize}[leftmargin=*, topsep=0pt, noitemsep]
    \item \textbf{Varying Depths.} We collect 0-7 hop reasoning data for curriculum learning~\citep{bengio2009curriculum} across complexity levels and forms (deductive, inductive, analytical).
    \item \textbf{Different Formats.} Collected data includes explicit or implicit rules as premises or constraints, enhancing the model’s flexibility in recognizing, parsing, and applying rules in diverse contexts. 
    \item \textbf{Multiple Inference Rules.} We focus the diversity of rules of inference, which implies in the dataset facilitates learning dynamic rule employment and length generalization further.
    \item \textbf{Context Dependency.} We focus on applying contextual rules adaptively for different questions. Correctly answering requires more than memorizing rules, reasoning, or using common sense.
    \item \textbf{Robust Evaluation.} We prioritize boolean and multiple-choice questions over free text to make it more conducive to obtaining rule-based outcome rewards and precise evaluation results.
\end{itemize}
\input{figure/training_data_examples}
These principles are critical to ensure training data captures the complexity and diversity inherent in rule-based reasoning tasks. The statistics of training data are presented in \cref{tab:data_stat}. \textbf{We highlight the importance and necessity of \ourdata with further analyses in \S\ref{subsec:data_curation_impact}.}

\section{Empirical Results}

\subsection{Experimental Setup}
\label{subsec:experiment_setup}

\input{table/ood_bench_statistics}

\textbf{Datasets and Benchmarks.} We assess the generalization of models on unseen tasks using subsets from BigBench Hard~\citep{suzgun-etal-2023-challenging}, BigBench Extra Hard~\citep{kazemi2025big}, and ProverQA~\citep{qi2025large}, as detailed in \cref{tab:ood_statistics}. We also employ AIME 2025 \citep{aime25}, GPQA (Diamond) \citep{rein2024gpqa}, and Coin Flip \citep{wei2022chain} as additional evaluation to assess whether \ourmethod can extrapolate to more general reasoning tasks. More details on \ourdata (Figure \ref{fig:training_data_examples}) are presented in \S\ref{sec:data_details}.

\textbf{Compared Baselines.}  We include five types of baselines: \textbf{(1) Prior rule-based reasoners (RBRs):} Hypotheses-to-Theories~\citep{zhu2023large}, Logic-of-Rule~\citep{servantez-etal-2024-chain}, and Rule-Guided Feedback~\citep{diallo2025rule}; \textbf{(2) Frontier reasoners:} OpenAI-o1 (\textit{o1-2024-12-17})~\citep{jaech2024openai}, o3-mini (\textit{o3-mini-2025-01-31})~\citep{ZhangOpenAIOS}, DeepSeek-R1~\citep{guo2025deepseek}, and Claude-3.7-Sonnet (\textit{claude-3-7-sonnet-20250219} with thinking mode)~\citep{anthropic-claude-3-7-sonnet} with standard zero-shot CoT prompting~\citep{wei2022chain}; \textbf{(3)  Behavioral cloning}~\citep{NIPS1988_812b4ba2}\footnote{Following RL literature nomenclature, we refer to models trained with the negative log-likelihood loss as behavioral cloning and perform task-focused supervised training to maximize baseline performance.}\textbf{:} SFT without CoT~\citep{wei2022finetuned}, SFT with short CoT~\citep{yeo2025demystifying}, and SFT with distilled long CoT~\citep{yeo2025demystifying} from o3-mini; \textbf{(4) Advanced RLVRs:} we compare RLVR approaches including GRPO~\citep{shao2024deepseekmath}, Dr. GRPO~\citep{liu2025understanding}, and DAPO~\citep{yu2025dapo} ; \textbf{(5) Curriculum Learning:} we also introduce recent curriculum learning approaches for RL: \textsc{AdaRFT} \citep{shi2025efficient}, and data-balance and easy-to-hard strategies from \cite{parashar2025curriculum}. %Due to \ourmethod only adding negligible wall-clock time during batch sampling, we use DAPO without its dynamic sampling to exclude additional computational overhead for fair comparison purposes.

% \begin{itemize}
%     \item \textbf{Prior Rule-based Methods:} We include the following work focused on logical reasoning with rules, 
%     \textbf{1) Training-free:} Hypotheses-to-Theories~\citep{zhu2023large}, Logic-of-Rule~\citep{servantez-etal-2024-chain}, Rule-Guided Feedback~\citep{diallo2025rule}. \textbf{2) Training-based: } \yang{TODO}. \iffalse Meta-RFFT~\citep{hu2025training}, WM-Neurosymbolic~\citep{wang-etal-2024-symbolic}, Distilled Inference Engine~\citep{wang-etal-2024-llms}.\fi
%     \item \textbf{Frontier Reasoners:} We evaluate strong proprietary reasoning models (OpenAI-o1, o3-mini~\citep{jaech2024openai}, DeepSeek-R1~\citep{guo2025deepseek}, Gemini-2.5-Pro~\citep{google_gemini_2025}, Claude-3.7-Sonnet~\citep{anthropic-claude-3-7-sonnet}) with standard step-by-step CoT prompting to maximize their potentials~\citep{wei2022chain}. % and open reasoning models (QwQ-32B~\citep{qwen2.5} and DeepSeek-R1-Distill-Qwen-32B~\citep{guo2025deepseek}).
%     \item \textbf{Supervised Methods:} In addition to the representative models above, we introduce three supervised variants that serve as simple yet strong baselines: 1) SFT without CoT, 2) SFT with CoT, and 3) SFT with Long CoT.
%     \item \textbf{Advanced RLVR Algorithms:} To demonstrate the robustness of our training method, we include the recent RLVR approaches: Naive GRPO~\citep{shao2024deepseekmath}, Dr. GRPO~\citep{liu2025understanding}, DAPO~\citep{yu2025dapo}, SRPO~\citep{zhang2025srpo}, and SSR~\citep{wang2025vl}.
% \end{itemize}

\textbf{Evaluation Metrics.}  All tasks in the work are evaluated using the algebraic mean of hard exact match, which is also equivalent to pass@1 accuracy under strict extraction and comparison. % For the evaluation of test-time scaling in \S\ref{para:test_time_scaling}, we also include pass@k to assess model's potential in multiple generation configuration such as Best-of-N~\citep{cobbe2021training} and sequential test-time scaling~\citep{deepscaler2025}.

\textbf{Implementation Details.}  \textbf{Training Setup:}
We use Qwen3 (4B and 8B base)~\citep{yang2025qwen3} as the base models and employ veRL~\citep{sheng2024hybridflow} for RL post-training. We set train and mini batch sizes to 64 for strict on-policy updates, and a rollout size of 64 per question. For hyperparameters in \ouralgo, we use a $\tau$ of 0.5 for moderately frequent domain weight updates and an $\epsilon$ of 0.1 for minimum sampling probability per domain, with a smoothing factor $\alpha$ of 0.5. %Our code and trained models will be released soon.
\textbf{Inference Setup:} We employ random sampling (temperature $\tau = 0.6$ and $top\text{-}p=0.95$ with a maximum output length of 2,048. For the rest of baselines, we use Qwen3-8B-Base for full-parameter SFT or RL. We perform five runs per test set and report the mean and standard deviation of the performance. %We also report one-sided p-values for statistical significance.

\subsection{\ourmethod Improves RLVR Performance and Efficiency}
\label{sec:results_and_analysis}

\input{table/main_results}

\textbf{In-Distribution Performance.}
As shown in \cref{tab:id_performance}, we report the ID tasks performance to depict the effectiveness of \ourmethod. Compared with the frontier LRMs, \ourmodelbest surprisingly outperforms with a large performance gap. Notably, \emph{on eight ID tasks}, OpenAI-o1 lags behind \ourmodelbest with $4.1\%$ point, whereas Claude-3.7-Sonnet underperforms with $4.5\%$ point. Also, \ourmodelbest outperforms prior strong RBRs such as HtT and Chain-of-Logic, which are built directly on top of OpenAI o3-mini for all tasks, except for AR-LSAT and Logical Deduction. This implies that \textbf{\ourmethod benefits from RLVRs to obtain higher improvement in rule understanding and utilization}. In addition, \ourmodelbest also outperforms recent RLVR methods which are trained with lower intra-task performance variance in \emph{eight tasks}, for instance, higher performance of 84.0\% \textcolor{OliveGreen}{(+4.8\%)} yet with a lower variance of 3.1\% \textcolor{OliveGreen}{(-0.5\%)}, comparing to DAPO (79.2\% with a variance of 3.6\%). \input{table/improvement_comparison}This demonstrates that \textbf{\ourmethod not only develops impressive task performance, but also maintains the domain performance balance}.

%\textbf{Held-in Performance.}\\
\textbf{Out-of-Distribution Performance.} As illustrated in~\cref{fig:ood-results}, \ourmodelbest surpasses frontier LRMs across \emph{three OOD benchmarks}. Specifically, it shows a remarkable 10.4\% improvement compared to OpenAI-o1. As depicted in \cref{tab:comparison_improvement}, \ourmodelbest consistently increases performance across the three OOD benchmarks, achieving the highest average performance gains of $\Delta 56.0\%$, including $\Delta 71.4\%$ on BBH, $\Delta 48.4\%$ on ProverQA, and $\Delta 48.2\%$ on BBEH. These findings highlight the effectiveness of \ourmethod in enhancing the general rule-based reasoning capabilities of models. As shown in \cref{tab:comparison_improvement}, the SFT baseline lags behind \ourmethod in both ID and OOD evaluations. Notably, while SFT improves ID performance to closely match \ourmethod (81.9\% versus 84.0\%), its OOD performance remains significantly lower (34.4\% versus 54.5\%). We conclude that, in contrast to RLVR, SFT does not effectively generalize to unseen rules or OOD scenarios, which is also aligned to~\citet{chu2025sft}. Extended evaluations on other reasoning tasks (AIME 2025, GPQA, and Coin Flip) and the test-time scaling are discussed in \S\ref{subsec:extrapolation_to_unseen_tasks} and \S\ref{subsec:test_time_scaling}, respectively. Moreover, a relative OOD demonstration in \cref{tab:case-study} reveals that \textbf{\ourmethod elicits models to extrapolate to new tasks by applying unseen rules, through a concise and logically rigorous reasoning trajectory.} 

\input{figure/domain_weight_rewards_dynamics}

\iffalse
\paragraph{RL Surpasses SL.}
On the conventional wisdom, we choose reinforcement learning instead of superivsed learning, we dissect the key difference and implication bewteen the two training paradigm with respect to logical reasoning.

\paragraph{Compatibility with Response Filtering.}
%\input{table/sampling_method_comparison}
To analysis the compatibility between our method and recent proposed methods. Unlike our method, they are tailored in the response re-sampling level, such as DAPO~\citep{yu2025dapo} and SRPO~\citep{zhang2025srpo}. We ablate the compositional effect of our method and them. The results are shown in Table %\ref{tab:sampling_method_comparison},
\fi

\iffalse
\textbf{Test-time Scalability.}\label{para:test_time_scaling}
In the cutting-edge discussion on the essence and usefulness of a longer thinking process,~\citep{fatemi2025concise} find that the extra generated tokens do not help improve the final prediction accuracy, while~\citep{yeo2025demystifying,yang2025towards} hold the opposite positions which claim that accurate results are not necessary with the long reasoning process. To investigate the interesting question, we study two experimental context scaling strategies on \ourmethod, including 1) training-time iterative scaling and 2) test-time scaling. For test-time scaling, we report sequential and parallel strategies, respectively. \yang{Plot a line chart. Add evalution results.}
\fi

%\paragraph{[TBD] Progress Reward Hacking.}

\section{Analyses}
% We conduct ablation studies on two major aspects of our training recipe to understand their contributions. 

\subsection{Advantages of Domain-aware Dynamic Sampling}
As depicted in \cref{fig:weight_rewards_dynamics}, we find that \ourmethod enhances task performance across all domains without tradeoff. Specifically, \ourmethod takes care of the underperformed task such as AR-LSAT and increases its domain weights consistently. Even for the low-portion domains (\eg, ProofWriter) and the decreased domains such as Clutrr, \ourmethod still steadily improves their training rewards and validation pass@1 without reaching a plateau. Interesting, analogous to the phenomenon described by~\citet{zucchet2025language}, the knowledge acquisition period during pre-training is accelerated fast on transition, but led to overfitting by the imbalanced data distributions. As shown in Table \ref{tab:more_baselines_comparison}, we compare the OOD performance with DAPO and \textsc{AdaRFT} to directly demonstrate that \ouralgo achieves superior generalization without requiring external difficulty estimation (as in \textsc{AdaRFT}) or additional rollout compute (as in DAPO).

Furthermore, we perform curriculum learning baselines as stated in \S\ref{subsec:curriculum_baselines}. In summary, \ouralgo surpasses these baselines by a significant margin (see Table \ref{tab:id_performance} and \ref{tab:offline_curriculumn_sft_rl_results}). Static curriculum methods fail because they treat easy and hard domains equally or experience vast distribution shift during training, while \ouralgo acts as an online scheduler, shifting compute resources from converged (easy) domains (\eg, ProntoQA) to under-optimized (hard) domains (\eg, AR-LSAT). This prevents over-optimization on solved tasks while preventing under-fitting on complex ones. In summary, we conclude that \ouralgo serves as an online data scheduling strategy, stablizing the dynamics of on-policy RL training and mitigates over-optimization.

%\paragraph{RQ2: How Well Varying Reward Buffer Size Works?}

%\paragraph{RQ5: How Different Training Task Compositions Perform?}

\subsection{Impact of Task Mixing Recipe}
\input{figure/task_mixing_recipe}
\cref{fig:task_mixing_recipe} illustrates the impact of mixing recipes of incremental tasks in OOD tasks across models of different sizes. The incremental mixing strategies show consistent gains as the model size increases, indicating that a larger model capacity generally contributes to better generalization. The mixed collection of various reasoning types enhances the OOD performance nearly to 80\% while the baseline “w/o All” at around 25\%. It further validates the benefit of our training data curation principles and the effectiveness of task mixing for learning generalization.

% \subsection{Advantages of Utilizing DADS for Multi-task RLVR}

\subsection{Investigation on Training Sample Efficiency}
We assess OOD performance by scaling training sample sizes among three methods at step 500, sampling data with varying domain proportions on a log scale. It is evident that for all methods (Figure \ref{fig:sample_efficiency}), as the proportion of each domain increases, \ourmethod consistently outperforms GRPO and Mix SFT at every proportion level. For the same proportion of data, \ourmethod achieves higher performance than Mix SFT as well as a slight advantage over GPRO.  For example, when the domain proportion is 1\%, \ourmethod achieves an OOD performance of about 60\% point, while the other two is around 59\% point and 37\% point. \iffalse Similarly, our method is more effective with high training sample efficiency to achieve the same OOD performance,\fi This implies the effectiveness of the online data scheduling stated in \cref{sec:dapo}.

As shown in \cref{fig:rlvr_comparison}, we also prove that \ourmethod improves sample efficiency with fewer training steps than the existing RLVR methods. It achieves comparable OOD performance with $\sim$66.8\% on average, roughly fewer 72 steps and approximately 1.4$\times$ acceleration compared to DAPO within the same training steps. Similarly, GRPO requires more steps to match \ourmethod's performance, which indicates that the dynamic batch-level data curricula in \ourmethod benefit RLVR and accelerate it to achieve higher performance with fewer costs.

\subsection{Case Study: \ourmethod Improves Quality of Rule Application}
We show examples from \ourmodelbest in various test sets in \cref{tab:case-study} to demonstrate how our training recipe improves the performance of the model. Before training, the model fails to solve the problems because of various reasons, such as poor reasoning logic, reasoning distracted by the noisy rules, \etc. After training, the model is able to derive a valid reasoning chain and answer rigidly. Additionally, we provide more successful and failed cases with their trajectories in Table \ref{tab:positive-case-study} and \ref{tab:negative-case-study}.
\input{table/case_studies}

\subsection{Generalization Mechanism for Compositional Rules}
\label{subsec:generalization_mechanism}
Motivated by the efficacy gap in \cref{tab:comparison_improvement}, we analyze the origin of \ourmethod's generalization capability. The OOD performance gains stem not just from the dynamic curriculum via \ouralgo but also from the emergence of meta introspection during training, fostering self-simulation and self-verification abilities. \ourmethod explores candidate paths and verifies logical consistency against inferred rules before concluding. This mirrors abstract introspection, where the model critiques intermediate outcomes and corrects flawed steps, enabling generalization beyond seen tasks. 

In the first ProverQA example (Table \ref{tab:ood_case_studies}), the untrained model makes unverified and speculative assumptions (\eg, ``that good intentions imply being lovable''). In contrast, the trained model adheres strictly to objective facts, noting that ``lovable'' is not explicitly defined or linked to those attributes in context. It analyzes available clues contextually, performing self-verification and consistency checks throughout the reasoning process. This rigorous behavior enables model to eliminate cognitive biases and logical fallacies, arriving at well-supported conclusions that align with established facts.

\section{Conclusion}
\label{sec:conclusion}
% \paragraph{Conclusion.}
% \label{sec:conclusion}
We introduce \ourmethod, a training framework tailored to enhance both effectiveness and efficiency of multi-domain training for RLVR. It harmonizes diverse rule-based reasoning capabilities across various tasks, resulting in higher performance efficiently compared to existing rule-based reasoners, frontier LRMs, strong supervised baselines, and prior RLVR methods. We aim to explore further research into data-centric approaches to facilitate reasoning efficiency in future work.

\paragraph{Limitations \& Future Work.}
% \begin{itemize}
%     \item Limitation on Rule Coverage.
%     \item Limitation on Rule Quality Filtering.
%     \item Limitation on Larger Scale Modeling.
%     \item Extrapolating to Long Logical Reasoning.
% \end{itemize}
In this study, due to the scarce and imbalanced nature of rule-based reasoning data, current methods may not cover all rule formats and complexities found in real-world applications, which hinders task generalization. Besides, our method is constrained by the quality of rule filtering, particularly when dealing with noisy or redundant rules that can negatively impact reasoning. Furthermore, while effective with smaller models (4B and 8B), its scalability to large-scale modeling remains unverified due to computational limitations, despite potentially higher effectiveness in complex scenarios. These limitations highlight areas for future improvement in expanding rule diversity and robustness to support longer reasoning trajectories.

\section*{Ethics Statement}\label{sec:ethics}
We adhere to ethical principles to ensure the responsible development and application of our proposed techniques. The research conducted in the article is in every respect in accordance with the ICLR code of ethics guidelines. Our work focusses on enhancing the rule-based reasoning abilities of models without directly involving human subjects or sensitive information. The study acknowledges ethical implications, such as the transparency of rule-based systems being advantageous for interpretability but raising concerns about accountability if rules are misapplied in high-stakes domains. We advocate for rigorous validation of rules against diverse datasets to avoid human biases in manually crafted rules. We also recognize the potential broader impacts, including the environmental and computational costs of LLM training, and strive to optimize our methods for efficiency.

\section*{Reproducibility Statement}\label{sec:reproducibility}
To ensure the reproducibility of our research, we provide detailed information regarding our methodology and experimental setup. The exact prompt templates utilized for datasets with both explicit and implicit rules during training and evaluation are detailed in Appendix \ref{sec:prompts}. Furthermore, Appendix \ref{sec:data_details} provides a comprehensive list of the data sources for both training and OOD evaluation, along with specifics on how the datasets were curated for the \ourdata. Our evaluation metrics are defined in \S\ref{subsec:experiment_setup}, with additional evaluation results and analyses presented in Appendix \ref{sec:additional_eval_results}. Finally, a complete breakdown of the computational infrastructure and all hyperparameter assignments for both training and inference can be found in Appendix \ref{sec:compute_infra}. The code, model, and data will be made publicly available upon completion of the review process.

\section*{Acknowledgments}
\label{sec:acknowledgments}
We would like to thank Zixia Jia, Hengli Li, and Xubo Qin from BIGAI for their valuable contributions in discussing the project, and Tong Wu from BIGAI for his initial trials, helpful discussions on RLVR training, and the infra-system operation and maintenance.

\bibliography{iclr2026_conference}
\bibliographystyle{iclr2026_conference}

\appendix
\clearpage
\section{Prompts}
\label{sec:prompts}
In this work, we use the same prompt template for each dataset for model training and evaluation.

\subsection{Prompts for Dataset with Explicit Rules.}
\begin{tcolorbox}[colback=pppurple!15, colframe=white, width=14cm,arc=3mm, auto outer arc, boxrule=1pt,top=10pt, bottom=10pt, left=10pt, right=10pt]
\textbf{Instruction:} Please answer the question based on the given rules and facts using either of [A/B/C/D] (or [True/False/Unknown]). Fill in the answer between $<$answer$>$ and $<$/answer$>$. Provide your step-by-step reasoning process between $<$think$>$ and $<$/think$>$.  \\\\
\textbf{Input:}   \\
- Rules: \{\{Rules\}\}\\
- Facts: \{\{Facts\}\}\\\\
\textbf{Question:}  \{\{Question\}\} \\\\
\textbf{Options:}  \{\{Options\}\} (OPTIONAL)
\end{tcolorbox}

\subsection{Prompts for Dataset with Implicit Rules.}
\begin{tcolorbox}[colback=pppurple!15, colframe=white, width=14cm,arc=3mm, auto outer arc, boxrule=1pt,top=10pt, bottom=10pt, left=10pt, right=10pt]
\textbf{Instruction:} Please answer the question based on the given contexts using either of [A/B/C/D] (or [True/False/Unknown]). Fill in the answer between $<$answer$>$ and $<$/answer$>$. Provide your step-by-step reasoning process between $<$think$>$ and $<$/think$>$.  \\\\
\textbf{Input:}   \\
- Context: \{\{Context\}\}\\\\
\textbf{Question:}  \{\{Question\}\} \\\\
\textbf{Options:}  \{\{Options\}\} (OPTIONAL)
\end{tcolorbox}

\section{Data Details}
\subsection{Data Sources}
\label{sec:data_details}
We list the training and evaluation data sources associated with the urls used in the paper as below.
The followings are the training and validation data sources:

\begin{itemize}
    \item \hyperlinkcite{tafjord-etal-2021-proofwriter}{\textcolor{black}{ProofWriter (2021)}}: \url{https://allenai.org/data/proofwriter}
    \item \hyperlinkcite{saparov2023language}{\textcolor{black}{ProntoQA (2023)}}: \url{https://github.com/asaparov/prontoqa}
    \item \hyperlinkcite{sinha-etal-2019-clutrr}{\textcolor{black}{Clutrr (2019)}}: \url{https://github.com/SiyuanWangw/RuleApplication/blob/master/Data/clutrr}
    \item \hyperlinkcite{zhong-etal-2022-analytical}{\textcolor{black}{AR-LSAT (2022)}}: \url{https://github.com/SiyuanWangw/RuleApplication/blob/master/Data}
    \item \hyperlinkcite{han-etal-2024-folio}{\textcolor{black}{FOLIO (2024)}}: \url{https://github.com/Yale-LILY/FOLIO/blob/main/data/v0.0}
    \item \hyperlinkcite{tian-etal-2021-diagnosing}{\textcolor{black}{LogicNLI (2021)}}: \url{https://github.com/omnilabNLP/LogicNLI/blob/main/dataset}
    \item \hyperlinkcite{xu-etal-2024-faithful}{\textcolor{black}{LogicalDeduction (2024)}}: \url{https://github.com/Aiden0526/SymbCoT/tree/main/data}
    \item \hyperlinkcite{logiqa2023}{\textcolor{black}{LogiQA (2023)}}: \url{https://github.com/csitfun/LogiQA2.0/blob/main/logiqa/DATA/LOGIQA}
\end{itemize}
The followings are the OOD test data sources:
\begin{itemize}
    \item \hyperlinkcite{suzgun-etal-2023-challenging}{\textcolor{black}{BigBench-Hard (2023)}}: \url{https://huggingface.co/datasets/lukaemon/bbh}
    \item \hyperlinkcite{qi2025large}{\textcolor{black}{ProverQA (2025)}}: \url{https://huggingface.co/datasets/opendatalab/ProverQA}
    \item \hyperlinkcite{kazemi2025big}{\textcolor{black}{BigBench-Extra-Hard (2025)}}: \url{https://github.com/google-deepmind/bbeh}
\end{itemize}

\subsection{Dataset Curation Details}

\input{table/data_stat}
For ProntoQA, we randomly negate some of the proof questions to avoid learning the shortcut of answer ``True''. For ProofWriter, we randomly sample ten percent of the orginal source data considering the imbalance nature of the whole training data. Then we use DeepSeek-R1 to generate the reasoning process including short CoT and long CoT sequences for each QA sample. For LogiQA, we use data with the reasoning type both categorical reasoning and conjunctive reasoning that leverages the implicit rule application and reasoning. For BigBench-Hard, we use the subset of ``logical\_deduction'' with three, five, and seven objects with varing levels of difficulties and select BoardgameQA and ZebraPuzzles from BigBench-Extra-Hard to keep consistent with our task definition for OOD evaluation.
\label{sec:dataset_curation_details}
% \subsection{Task Descriptions}
% \label{subsec:task_descriptions}
% We supply the target and requirements of each tasks with the aformentioned datasets as below.
% \begin{itemize}
%     \item \textbf{ProofWriter:} xxx.
%     \item \textbf{ProntoQA:} xxx.
%     \item \textbf{Clutrr:} xxx.
%     \item \textbf{AR-LSAT:} xxx.
%     \item \textbf{FOLIO:} xxx.
%     \item \textbf{LogicNLI:} xxx.
%     \item \textbf{LogicalDeduction:} xxx.
%     \item \textbf{LogiQA:} xxx.
%     \item \textbf{Bigbench-Hard:} xxx.
%     \item \textbf{Bigbench-Extra-Hard:} xxx.
%     \item \textbf{ProverQA} xxx.
% \end{itemize}

\section{Additional Evaluation Results}
\label{sec:additional_eval_results}

\subsection{Challenges of Different Rule Settings}

\input{figure/challenging_rule_settings}

We investigate the task performance on Clutrr, with three-level rule settings of \cref{fig:rule_settings} in the following: 1) Ordered Rules: rules are arranged in their application order; 2) Shuffled Rules: rules are provided in a random order; 3) Noisy Rules: rules are shuffled and include irrelevant ones. aligns with real-world scenarios, rules may contain distractors. To our expectation, the task with ordered rules achieves the best performance among them likely due to the logical sequence aiding in task execution. Shuffled Rules, while still contain only the relevant rules but in a random order, show a moderate performance drop. Noisy rules result in the most significant performance reduction with the added complexity of redundant rules as distractors, highlighting the negative effect on task performance.

\subsection{\ourmethod can Adapt to Varying Rule Complexity}
\input{figure/varying_task_complexity}
As depicted in \cref{fig:varying-complexity-results}, we present the extended OOD evaluation results, with test sets separated by rule complexity (\ie task difficulty). The BBH, ProverQA, and BBEH benchmarks consist of questions requiring reasoning up to various difficulties that hinge on the diverse factors of query complexity.

Specifically, for BBH, we divide the original test set into three difficulty levels based on the multi-hop number of the query. For ProverQA, we adopt the original difficulty levels from its source, which is separated by the number of reasoning steps. For BBEH, the test set was categorised into three levels according to the length of the query (in tokens): Easy [0, 1068), Medium [1068, 2175), and Hard [2175, 2741). Thus, we test the generalization capabilities on the three subsets of each OOD benchmark and report their performance on the higher difficulty questions. Not surprisingly, performance in easy subset exhibit substantially stronger than corresponding medium- and hard-level subsets, with an average pass@1 of 86.7\% compared to the 63.0\% \textcolor{BrickRed}{(-23.7\%)} of hard subset across benchmarks. Interestingly, we notice that \ourmodel drops significantly along with subsets in different difficulties, while we maintain the still performance in BBH. One possible explanation is that the base model, Qwen3-8B-Base, might have encountered a test set leakage, given that BBH was published in late 2022 but Qwen3 models were released in 2025~\citep{qwen3}. Therefore, we suggest assessing our models to more challenging benchmarks to achieve more reasonable results. We leave this direction for future work.

\subsection{Test-time Scalability}\label{subsec:test_time_scaling}
In the cutting-edge discussion on the essence and usefulness of a longer thinking process,~\cite{fatemi2025concise} find that the extra generated tokens do not help improve the final prediction accuracy, while~\cite{yeo2025demystifying,yang2025towards} hold the opposite positions which claim that accurate results are not necessary with the long reasoning process. To investigate this interesting question for rule-based reasoning, we study parallel test-time scaling strategies with \ourmethod. We perform repeated sampling to investigate the upper limit of performance for each collection of rule-based reasoning problems in the way of~\citet{brown2024large}, illustrated in \textcolor{deepbrown}{brown curves}. Concretely, we take the \textcolor{lightbrown}{majority vote} and \textcolor{lightpurple}{normalized weighted sum} methods in the way of~\citet{wang2023selfconsistency}.\input{figure/test_time_scaling} As depicted in Figure \ref{fig:test-time-scaling}, test-time scaling demonstrates different effectiveness across benchmarks. For BigBench Hard, all Pass@K methods achieve near-perfect performance (close to 100\%) with minimal scaling, indicating the limited complexity of the testbed to distinguish scaling benefits. In ProverQA, Pass@K Coverage consistently outperforms both the majority and weighted approaches, with performance gaps widening as the sample size increases ($\sim$1.7\% at 128 samples). BigBench Extra Hard reveals the most substantial scaling benefits, where the coverage method achieves 73.1\% pass@k performance compared to $\sim$72\% for alternative approaches at 128 samples. The consistent superiority of Coverage sampling across challenging benchmarks (ProverQA and BigBench Extra Hard) suggests that diverse solution exploration outweighs consensus-based aggregation for complex reasoning tasks. These findings support the position that extended reasoning processes, when properly sampled, enhance the prediction accuracy of difficult problems in rule-based reasoning.

\subsection{Comparison with More Baselines}\label{subsec:comparison_more_baselines}
We compare the differences and advantages of \ouralgo versus DAPO and \textsc{AdaRFT} on three OOD benchmarks. DAPO's dynamic sampling employs an online filtering scheme, which repeatedly samples and discards rollout generations until rewards fall within a target range, ensuring a mix of partially correct and incorrect answers. \ouralgo is more efficient than DAPO with its dynamic sampling. \ouralgo samples training problems before responses are generated, preventing wasted computation of creating and then discarding unwanted responses. \textsc{AdaRFT} relies on an opaque deterministic difficulty estimation based on Qwen2.5-Math-7B \citep{yang2024qwen2}, which limits the precision of initial difficulty judgement and introduces inductive bias from the LLM-as-a-judge model \citep{gu2024survey}. For direct comparison, \ouralgo only needs coarse-grained metadata labels from original data sources (it can be the form of ``dataset-as-a-domain'') for domain partitioning, reducing the need for prior dependencies like domain annotation or problem difficulty scoring.

\input{table/more_baseline_comparison}

\subsection{Robustness Analyses on Hyperparameters of \ouralgo.}\label{subsec:robustness_analyses}
To investigate the robustness and sensitivity of the hyperparameters, we conduct concise analyses of the sensitivity and effect of the hyperparameters $\alpha$, $\tau$, and $\epsilon$ in \ouralgo. Specifically, we keep two of the three hyperparameters constant and incrementally adjust the remaining hyperparameter. For each hyperparameter combination, we observe the model's performance on the BBH, ProverQA, and BBEH, three OOD benchmarks after convergence.

\input{table/hyperparameters_analyses}

As depicted in Table \ref{tab:hyperparameter_analysis}. We list the observations and effects as below. 
\begin{itemize}
    \item \textbf{Smoothing factor $\alpha$:} Performance shows an inverse ``U-shape curve'', initially increasing and then decreasing as $\alpha$ increases. Optimal performance is observed around $\alpha=0.5$ with 70.4\% pass@1. Larger $\alpha$ means more dependence on historical rewards. An excessively high or low $\alpha$ can degrade performance, suggesting a sweet spot for balancing historical and current rewards.
    \item \textbf{Temperature $\tau$ for magnitude control:} Performance generally improves as $\tau$ increases, with a slight dip at $\tau=1.0$. The best performance is observed at $\tau=0.8$ with 70.4\% pass@1. Smaller $\tau$ indicates more sensitivity to reward fluctuations. Higher $\tau$ (up to a point) seem to improve performance, suggesting that less sensitivity to individual fluctuations can be beneficial.
    \item \textbf{Minimum sampling weight $\epsilon$:} Observation: Performance generally decreases as $\epsilon$ increases. The best performance is at $\epsilon=0.1$ with 70.4\% pass@1. Larger $\epsilon$ indicates a more average sampled number for each domain. This suggests that increasing the minimum sampling weight leads to a slight degradation in performance, implying that a lower $\epsilon$ (allowing for more varied sampling) is generally better.
\end{itemize}

\subsection{Extrapolating to Unseen Reasoning Tasks.}\label{subsec:extrapolation_to_unseen_tasks}
As presented in Table \ref{tab:additional_ood_results}, we further explore the generalization to other verifiable domains, such as math, multi-disciplinary, and symbolic reasoning. We complement expanded inference-only OOD evaluation on more general reasoning tasks and show the extended OOD evaluation results including: (1) math (AIME 2025) \citep{aime25}; (2) multi-disciplinary (GPQA-Diamond) \citep{rein2024gpqa}; and (3) symbolic (Coin Flip) \citep{wei2022chain} reasoning benchmarks based on \ourmodelbest which is only trained on the rule-based reasoning tasks described in \S\ref{subsec: data_curation}.

\input{table/more_unseen_tasks_eval}

\subsection{Effectiveness of \ouralgo on Extremely High Difficulty.}\label{subsec:effectiveness_on_high_difficulty}
To investigate whether \ouralgo, the dynamic sampling approach, can still be a highly difficult learnable set of tasks by assigning a lot of computations to learn these samples, we identify the existence of an unlearnable set of extremely difficult tasks within our training domains, such as AR-LSAT. As Figure 4 illustrates, the training rewards for AR-LSAT start at approximately 0.11 and steadily increase to approximately 0.42. This improvement correlates with an increase in the training sample size, indicating a more assigned training computation.

Interestingly, even after Step 120, when the AR-LSAT domain was allocated nearly 20\% of the computation in each training batch, other domains maintained their upward slope in reward/validation pass@1 curves. This leads us to conclude that the DADS algorithm achieves an optimal balance in computation assignment within multi-domain reinforcement learning dynamics. Although the experimental dashboard (Figure 4) might suggest an occasional over-emphasis on one domain, overall performance indicates an effective allocation strategy.

To empirically demonstrate it, we provide an additional comparison experiment between \ourmethod with \ouralgo and without \ouralgo on the data mixing training (infusing the two-curve comparison of AR-LSAT).

We perform GRPO-style RLVR training with or without \ouralgo on our curated rule-based reasoning dataset. We conduct interval evaluations every 50 training steps to assess the reasoning accuracy (Pass@1) on AR-LSAT, which we assume is an extremely hard domain to optimize with \ouralgo. Finally, we compare the results of two setups and showcase the step-wise gains of validation performance. The comparison results between ``w/o \ouralgo'' and ``w/ \ouralgo'' assessed on AR-LSAT are shown in Table \ref{tab:high_difficulty_results}.

\input{table/high_difficulty_tasks}

\subsection{Comparison with Data Scheduling and Balancing Baselines}
\label{subsec:curriculum_baselines}
We provide additional simple data scheduling baselines including Easy-to-hard SFT \citep{sun2024easytohard}, Easy-to-hard RL \citep{parashar2025curriculum}, and Data-balance RL \citep{parashar2025curriculum} for comprison purposes with \ouralgo.

\paragraph{Baselines setup.}
\textbf{Easy-to-hard SFT / RL:} We perform progressive curriculum training using SFT or GRPO-style RL on partitioned data subsets, respectively. First, we adopt LLM-as-a-judge with a Qwen3-8B-Base to obtain difficulty scores of problems. This allows us to partition the training data into eight domains based on difficulty. We then train a Qwen3-8B-Base model without \ouralgo on this offline-partitioned data, progressively moving from easier to harder problems with the same portion of training epochs for each data partition.
\textbf{Data-balance RL:} We perform GRPO-style policy gradient-based RL training using a balanced domain setup. First, based on the training data statistics in Table \ref{tab:data_stat}, we calculate the average samples per domain across all eight domains as 3,571 problem samples. Second, we apply repeated domain-balanced sampling to ensure 3,571 sampled problems for each domain. This involves down-sampling domains with more samples than the average and up-sampling those with fewer. Finally, we disable \ouralgo and initiate RL training from scratch with this data-balanced setup.

\paragraph{Results.} As shown in Table \ref{tab:offline_curriculumn_sft_rl_results}, we demonstrate that training with \ouralgo outperforms other three baselines of Easy-to-hard SFT / RL and Data-balance RL recipes both the three OOD rule-based reasoning benchmarks and the ODD general reasoning benchmarks, such as AIME 2025, GPQA (Diamond), and Coin Flip. 

\input{table/offline_curriculum_sft_rl_results}

\subsection{Impact of Rule-based Reasoning Data Curation}\label{subsec:data_curation_impact}
To the best of our knowledge, there is no accessible dataset used for training rule-based reasoners. We argue that current approaches over-specialize in math and code \citep{yu2025dapo}, omitting the scalability of general rule-based and natural language reasoning. This narrow data focus hinders progress on broader reasoning capabilities, a critical bottleneck we aim to address. Moreover, our collected multi-domain dataset, \ourdata (\S\ref{subsec: data_curation}), is integral to evaluate \ouralgo, which specifically mitigates online training problem scheduling for such data in RLVR for LLMs.

\input{table/mix_math_rule_rl}

Without these intrinsically linked multi-domain datasets, the full benefits of \ouralgo can not be comprehensively demonstrated. As depicted in Table \ref{tab:mix_math_rule_rl}, additional experiments (RL on math-only data, RL on rule-only data, and RL jointly on both of them) using a naive GRPO objective (without \ouralgo) further underscore the irreplaceable necessity of our curated data, proving it can not be substituted by training solely on existing math problems such as AIME 1983-2024 \citep{aime_1983_2024}.

\section{Computational Infrastructure \& Hyperparameters}
\label{sec:compute_infra}

We list the details of the computational infrastructure and hyperparameters of training and inference used in this work in Table \ref{tab:hyperparams}.
\input{table/hyperparameters}

% \section{Computational Cost Comparison}
% \label{sec:compute_cost}

\section{Example Trajectories}
\label{sec:example_trajectories}
We provide additional positive and negative examples (\ie, failed cases) with fine-grained trajectories from the test sets of various tasks to demonstrate clear step-by-step logical reasoning patterns and well-organized thought processes of \ourmethod after training in \cref{tab:positive-case-study} and \cref{tab:negative-case-study}.

\input{table/positive_case_studies}
\input{table/negative_case_studies}
\input{table/ood_case_studies}

\end{document}

%% file: math_commands.tex
\usepackage{amsmath,amsfonts,bm}

\def\eqref#1{equation~\ref{#1}}
\def\1{\bm{1}}

\DeclareMathAlphabet{\mathsfit}{\encodingdefault}{\sfdefault}{m}{sl}
\SetMathAlphabet{\mathsfit}{bold}{\encodingdefault}{\sfdefault}{bx}{n}

\newcommand{\R}{\mathbb{R}}

%% file: config.tex
\usepackage{hyperref}
\usepackage{url}

\usepackage{setspace}
\usepackage[most]{tcolorbox}
\usepackage{array}
\usepackage{booktabs}
\usepackage{tabularx}
\usepackage{placeins}

\usepackage{xcolor}
\usepackage{colortbl}
\usepackage{multirow,makecell}
\usepackage[capitalise,noabbrev]{cleveref}
\usepackage{epigraph}
\usepackage{pifont}
\usepackage{ragged2e}
\usepackage{arydshln}
\usepackage{tikz}
\usetikzlibrary{decorations.shapes}
\usepackage{amsfonts}
\usepackage{amssymb}
\usepackage{enumitem}
\usepackage[normalem]{ulem}
\usepackage{hypcap}
\usepackage{capt-of}
\usepackage{fontawesome}
\usepackage{bm}
\usepackage{atbegshi}
\usepackage{hyperref}
\usepackage{nameref}
\usepackage{colortbl}
\usepackage{subfigure}
\usepackage{xspace,mfirstuc,tabulary}
\usepackage{newunicodechar}
\usepackage{atbegshi}
\usepackage{mdframed}
\usepackage{algorithm}
\usepackage[noend]{algpseudocode}
\usepackage{soul}
\usepackage{bbm}
\usepackage{rotating}
\usepackage[outline]{contour}
\usepackage{ifthen}
\usepackage{framed}
\usepackage{amsmath}
\usepackage{mathtools}

\usepackage{subfigure}
\usepackage{wrapfig}
\usepackage{caption}
\usepackage{float}
\usepackage{tcolorbox}
\usepackage{xspace}
\usepackage[misc,geometry,weather]{ifsym}
\usepackage[hang,flushmargin]{footmisc}

%% file: macro.tex
\definecolor{OliveGreen}{rgb}{0.0, 0.5, 0.0}
\definecolor{ggreen}{HTML}{f0f5ef}
\definecolor{rred}{HTML}{f9efed}
\definecolor{bblue}{HTML}{eef2f9}
\definecolor{light_purple}{HTML}{dfccfa}
\definecolor{ppurple}{HTML}{9f9ac8}
\definecolor{pppurple}{HTML}{7159a6}
\definecolor{ppppurple}{HTML}{431280}
\definecolor{gray}{gray}{0.9}
\definecolor{gray2}{gray}{0.7}
\definecolor{pale_red}{rgb}{0.90,0.61,0.58}
\definecolor{pale_green}{rgb}{0.55,0.75,0.60}
\definecolor{codepurple}{rgb}{0.58,0,0.82}
\definecolor{codegray}{rgb}{0.5,0.5,0.5}
\definecolor{darkgreen}{rgb}{0.09, 0.46, 0.16}
\definecolor{darkred}{rgb}{0.72, 0.13, 0.07}
\definecolor{deepbrown}{HTML}{8b4515}
\definecolor{lightbrown}{HTML}{c19a68}
\definecolor{lightpurple}{HTML}{b29cda}
\definecolor{MK_Three_Two}{RGB}{175,141,195}
\tcbset{on line, 
        boxsep=1pt, left=0pt,right=0pt,top=0pt,bottom=0pt,
        colframe=white,
        colback=light_purple,  
        highlight math style={enhanced}
}
\newtcbox{\greenbox}[1][]{
  on line,
  boxsep=2pt,left=0pt,right=0pt,top=0pt,bottom=0pt,
        colframe=white,
        colback={green!10},  
        highlight math style={enhanced}, #1
}
\newtcbox{\redbox}[1][]{
  on line,
  boxsep=2pt,left=0pt,right=0pt,top=0pt,bottom=0pt,
        colframe=white,
        colback={red!10},  
        highlight math style={enhanced}, #1
}
\newcommand{\yang}[1]{\textcolor{OliveGreen}{[Yang] #1}}

\newcolumntype{C}{>{\arraybackslash}X}

\newcommand{\ourmethod}{\textsc{RuleReasoner}\xspace}
\newcommand{\ouralgo}{\textsc{Dads}\xspace}
\newcommand{\ourmodel}{\textsc{RuleReasoner-4B}\xspace}
\newcommand{\ourmodelbest}{\textsc{RuleReasoner-8B}\xspace}
\newcommand{\ourdata}{\textsc{RuleCollection-32K}\xspace}

\newcommand\coauth{$^\star$}
\newcommand{\correspond}{$^\dag$}

\newcommand{\RequireIndent}{\item[]~~~~~~}

\usetikzlibrary{shapes} % for ellipse node
\usetikzlibrary{shadows.blur}
\colorlet{myred}{red!85!black!65}
\colorlet{mycyan}{blue!40!cyan!85!black!55}
\colorlet{myblue}{blue!70!cyan!85!black!55}
\colorlet{mydarkblue}{myblue!90!black!90}
\colorlet{mydarkerblue}{myblue!80!black!90}
\colorlet{mydarkestblue}{myblue!70!black!90}
\colorlet{mylightblue}{blue!50!cyan!80!black!45}
\colorlet{mypink}{magenta!95!red!100!black!72}
\colorlet{mypurple}{blue!50!red!90!black!70}
\colorlet{mydarkpurple}{blue!50!red!70!black!70}
\colorlet{mygreen}{green!60!black!75}
\colorlet{mylightgreen}{green!80!red!90!black!50}
\colorlet{myorange}{orange!95!black!70}
\colorlet{mybrown}{brown!80!orange!95!black!80}
\colorlet{myyellow}{yellow!80!orange!95!black!95}
\colorlet{mylightyellow}{yellow!80!orange!95!black!45}

\def\R{1} % default pie radius
\tikzstyle{pin}=[very thin,line cap=round]
\tikzset{
  slice/.style={fill=#1,draw=#1!80!black,line join=round,line width=0.4,
                blur shadow={shadow blur steps=20,shadow xshift=0.5,
                             shadow yshift=-1,shadow opacity=40}},
  slice/.default=myblue,
  pics/piechart/.style n args={2}{ % SIMPLE PIECHART 1
    code={
      \foreach \frac/\name/\col [
        count=\i,
        remember=\angb as \anga (initially #1), % start angle
        evaluate={\angm=\anga-\frac*1.8;  % middle angle
                  \angb=\anga-\frac*3.6;  % final angle
                  \exp=\R*max(0.02,0.2*(1-\frac/100)^8); % explode (radial shift)
                  \r=\exp+\R*max(0.5,(0.8-\frac/100));} % label radial positon
      ] in {#2}{ % loop over slices
        \coordinate (P\i) at (\angm:\exp+\R);
        \draw[slice=\col] % slice / circle sector
          (\angm:\exp) --++ (\anga:\R) arc(\anga:\angb:\R) -- cycle;
        \node[white,scale=0.9]
          at (\angm:\r) {\contour{\col!75!black}{\bm{$\frac\mathbf{\%}$}}};
        \node[\col!70!black,anchor=180+\angm,inner sep=4]
          at (P\i) {\bf\bm{$\name$}};
      }
    }
  },
  pics/piechart2/.style n args={2}{ % SIMPLE PIECHART 2
    code={
      \foreach \frac/\name/\col [
        remember=\angb as \anga (initially #1), % start angle
        evaluate={\angm=\anga+\frac*1.8;  % middle angle
                  \angb=\anga+\frac*3.6;  % final angle
                  \exp=\R*max(0.02,0.14*(1-\frac/100)^3); % explode (radial shift)
                  \r=\exp+\R*max(0.5,(1-\frac/100)^2);} % label radial positon
      ] in {#2}{ % loop over slices
        \message{^^J frac=\frac: anga=\anga -> angb=\angb}
        \draw[slice=\col] % slice / circle sector
          (\angm:\exp) --++ (\anga:\R) arc(\anga:\angb:\R) -- cycle;
        \ifdim \frac pt > 10pt % label inside slice
          \node[white,align=center]
            at (\angm:\r) {\boldlabel{$\name$}{\frac}};
        \else % label outside slice
          \node[\col!5!black,align=center,anchor=180+\angm,inner sep=3]
            at (\angm:\exp+\R) {\boldlabel{$\name$}{\frac}};
        \fi
      }
    }
  },
}

\newcommand\boldlabel[2]{\bf\bm{#1}\\[-1]\small\bm{$#2\mathbf{\%}$}}

\newcommand\unmarkedfootnote[1]{%
  \begingroup
  \renewcommand\thefootnote{}% Remove footnote number
  \footnote{#1}% Insert footnote
  \addtocounter{footnote}{-1}% Prevent incrementing footnote counter
  \endgroup
}

\newmdenv[
  backgroundcolor=purple!10,
  skipabove=1em,
  skipbelow=0em,
  leftline=true,
  topline=false,
  bottomline=false,
  rightline=false,
  linecolor=purple!88,
  linewidth=4pt
]{rquote}

\newmdenv[
  backgroundcolor=OliveGreen!8,
  skipabove=1em,
  skipbelow=0em,
  leftline=true,
  topline=false,
  bottomline=false,
  rightline=false,
  linecolor=OliveGreen!88,
  linewidth=4pt
]{gquote}

\hypersetup{
  colorlinks,
  citecolor=MK_Three_Two,
  linkcolor=red,
  urlcolor=blue}

\makeatletter
\newcommand*{\hyperlinkcite}[1]{\hyper@link{cite}{cite.#1}}
\makeatother

\definecolor{lightpink}{HTML}{FFDEEF}
\definecolor{lightgreen}{HTML}{E2FEC3}
\definecolor{lightblue}{HTML}{ECF4FF}

\newtcbox{\redtextbox}{on line, box align=center, colback=red!10,colframe=white,size=fbox,arc=3pt}
\newtcbox{\greentextbox}{on line, box align=center, colback=green!10,colframe=white,size=fbox,arc=3pt}

\definecolor{darkgreen}{rgb}{0.09, 0.46, 0.16}
\definecolor{darkred}{rgb}{0.72, 0.13, 0.07}
\definecolor{iccvblue}{rgb}{0.21,0.49,0.74}
\definecolor{sbblue}{HTML}{5273AD}
\definecolor{sborange}{HTML}{D78357}
\definecolor{sbgreen}{HTML}{5fA86C}
\definecolor{sbred}{HTML}{BD4C53}
\definecolor{mypurple}{HTML}{431280}

\newcounter{takeaway}
\newcounter{conclusion}
\newcommand{\coloreduline}[3][black]{\bgroup\markoverwith{\textcolor{#1}{\rule[-0.5ex]{#2}{0.4pt}}}\ULon{#3}\egroup}

\makeatletter
\DeclareRobustCommand\onedot{\futurelet\@let@token\@onedot}
\def\@onedot{\ifx\@let@token.\else.\null\fi\xspace}
\def\eg{\emph{e.g}\onedot} 

\def\ie{\emph{i.e}\onedot}

\def\etc{\emph{etc}\onedot}

\makeatother

%% file: figure/ood_performance.tex
\begin{figure*}[hb]
  \centering
  % \vspace{-0.3em}
  \resizebox{1.0\linewidth}{!}{\includegraphics{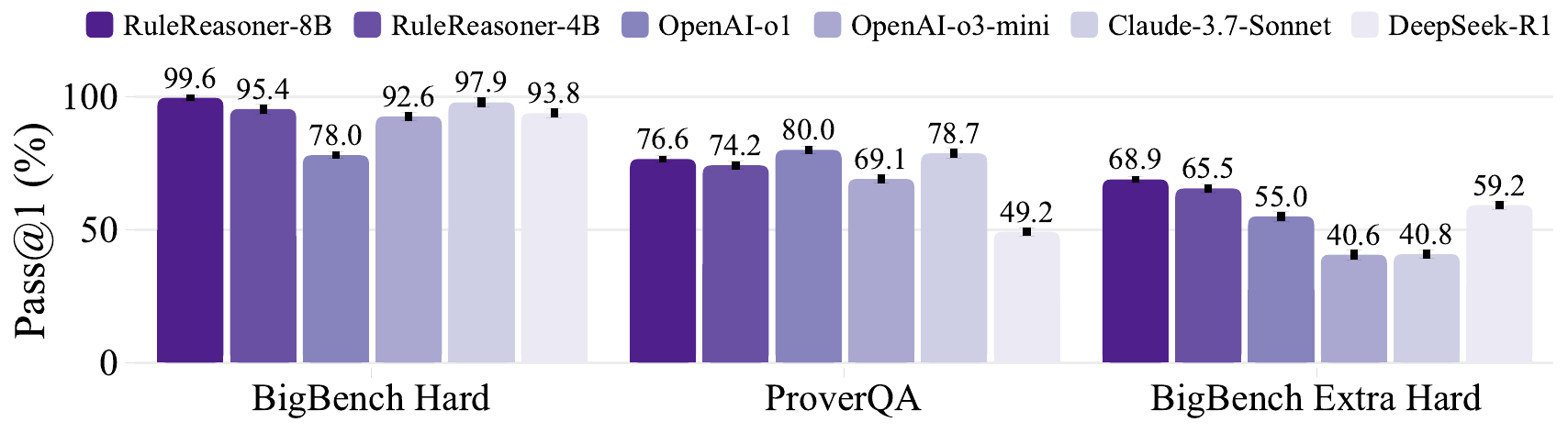}}
  \vspace{-6mm}
  \caption{Out-of-distribution performance comparison between \ourmethod (8B and 4B) and other frontier reasoning models on challenging rule-based reasoning benchmarks.}
  \label{fig:ood-results}
  %\vspace{-3mm}
\end{figure*}

%% file: figure/training_recipe.tex
% \iffalse
\begin{wrapfigure}{r}{0.55\textwidth}
    %\centering
    % \vspace{-6mm}
    % \begin{minipage}{0.51\textwidth}
        \centering
        \includegraphics[width=1\linewidth]{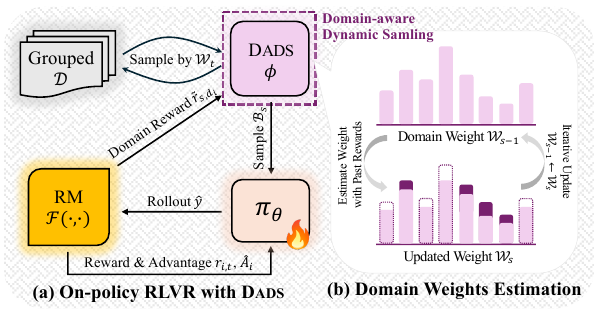}
        \caption{\ourmethod{} training recipe.}
        \label{fig:training_recipe}
        % \vspace{-10mm}
    % \end{minipage}
    \vspace{-3mm}
\end{wrapfigure}
% \fi

\iffalse
\begin{figure*}[ht]
    \centering
    \resizebox{0.9\linewidth}{!}{\includegraphics{figure/raw/training_recipe.pdf}}
    \caption{Diagram of the \ourmethod training recipe.}
    \label{fig:training_recipe}
\end{figure*}
\fi

%% file: table/domain_dynamic_sampling.tex
\begin{wrapfigure}{R}{0.52\textwidth}
    \vspace{-8mm}
    \begin{minipage}{0.52\textwidth}
        \begin{algorithm}[H]
        \caption{Domain-aware Dynamic Sampling}\label{alg:algo1}
        \begin{algorithmic}[1]
        \Require Policy model: $\pi_\theta: \mathcal{X} \rightarrow \mathcal{Y}$;
        \RequireIndent Reward model: $\mathcal{R}_\text{EM}(\cdot, \cdot): \mathcal{Y, \hat{Y}} \rightarrow \{0, 1\}$;
        \RequireIndent Last weight: $\mathcal{W}_{s-1} := \{w_1, w_2, \dots, w_n\}$;
        \RequireIndent Grouped data: $\mathcal{D} := \{(d_i, q_i, r_i, y_i)\}_{i=1}^{n}$\RequireIndent where domain: $d_i \in \{d_1, \cdots, d_n\}$;
        %\RequireIndent Historical reward buffer: $\mathcal{Q}_\text{history} \subseteq \{0, 1\}$;
        \RequireIndent Hyperparameters: $\{\alpha, \epsilon, \tau\} \subset \mathbb{R}^+$.
        \Ensure  Constructed batch of samples: $\mathcal{B}_s$.
        % \Ensure  Re-weighted batch: $\mathcal{B}_t^\text{weight}$.
        \Procedure{Train Step $s$ Sampling}{}
        %\For{$i = 1, 2, \dots, n$}
        % \State \textbf{Initialize:} $\mathcal{B}_{t-1}^\text{raw} =$ $\mathcal{W}_{t-1} \times \mathcal{D}$; $r_\text{target} = \mu$
        \State \textbf{Initialize:} $\mathcal{B}_{s-1} \leftarrow$ $\mathcal{W}_{s-1} \times \mathcal{D}$; $\tilde{r}_{0, d_i} \gets 0$. % $r_\text{target} \leftarrow 1$;
        % \State $\mathcal{A}_{t-1} \leftarrow$ $\pi_\theta(\mathcal{B}_{t-1}^\text{raws})$ 
        \State $\mathcal{Y}_{s-1} \leftarrow$ $\pi_\theta(\mathcal{B}_{s-1})$ 
         %\Comment{\colorbox{green!20}{\textsc{Rollout}}}
          \Comment{\colorbox{ppurple}{\textcolor{white}{\textbf{\textsc{Rollout}}}}}
        \State $\{\{r_{s,d_i,j}\}_{j=1}^{m_i}\}_{i=1}^n \leftarrow \mathcal{R}_\text{EM}(\mathcal{Y}_{s-1}, \hat{\mathcal{Y}}_{s-1})$
        \State \textcolor{mypurple}{/* Update estimated rewards */}
        %\State \textcolor{BrickRed}{$\{\bar{r}_{d_i}\}_{i=1}^n \leftarrow$
        %$\operatorname{CalcRW}$\text{($\mathcal{R}_\text{acc}, \mathcal{A}, \mathcal{Q}_\text{history}$)}}
        %\State \textcolor{BrickRed}{$r_\text{target} \leftarrow \alpha \cdot %\sum_{i=1}^{\lvert\mathcal{Q_\text{history}}\rvert}\frac{1}{i}$}
        \State $\{\bar{r}_{s,d_i}\}_{i=1}^n \leftarrow \{\frac{1}{m_i}\sum_{j=1}^{m_i} r_{s,d_i,j}\}_{i=1}^n$
        \State $\{\widetilde{r}_{s,d_i}\}_{i=1}^n \leftarrow$ $\{\alpha \widetilde{r}_{s-1,d_i} + (1-\alpha)\bar{r}_{s,d_i}\}_{i=1}^n$
        %\State $r_\text{target} \leftarrow \mu$
        \State \textcolor{mypurple}{/* Calculate weights by rewards */}
        \For{$i = 1, 2, \dots, n$} %\Comment{\colorbox{red!20}{\textsc{Reweighting}}}
            \State $v_{s,d_i} \leftarrow 1 - \widetilde{r}_{s, d_i}$ 
            %\State $w_{s, d_i} \leftarrow \exp((v_{s, d_i} + \epsilon) / \tau)$
            \State $w_{s,d_i} \leftarrow \exp(v_{s,d_i}/\tau) + \epsilon$
        \EndFor
        \State $\mathcal{W}_s := \{w_{s,d_i}^\text{norm}\}_{i=1}^n$ \Comment{\colorbox{pppurple}{\textcolor{white}{\textbf{\textsc{Normalizing}}}}}
        \State $~~~~~~~~ = \{w_{s,d_i} / \sum_{j=1}^n w_{s,d_j}\}_{i=1}^n$ 
        %\EndFor
        \State \textcolor{mypurple}{/* Re-sample w.r.t. optimized weights */}
        \State $\mathcal{B}_s \leftarrow \mathcal{W}_s \times \mathcal{D}$ \Comment{\colorbox{ppppurple}{\textcolor{white}{\textbf{\textsc{Sampling by} $\mathcal{W}_s$}}}}
        % \State $\mathcal{B}_t^\text{weight} \leftarrow \mathcal{W}_t \times \mathcal{D}$ \Comment{\colorbox{SkyBlue!50}{\textsc{Sampling by} $\mathcal{W}_t$}}
        \State \textbf{return} $\mathcal{B}_s$
        % \State \textbf{return} $\mathcal{B}_t^\text{weight}$
        \EndProcedure
        \end{algorithmic}
        \end{algorithm}
\end{minipage}
\vspace{-8mm}
\end{wrapfigure}

%% file: figure/training_data_examples.tex
\begin{figure*}[t]
  \centering
  \resizebox{0.9\linewidth}{!}{\includegraphics{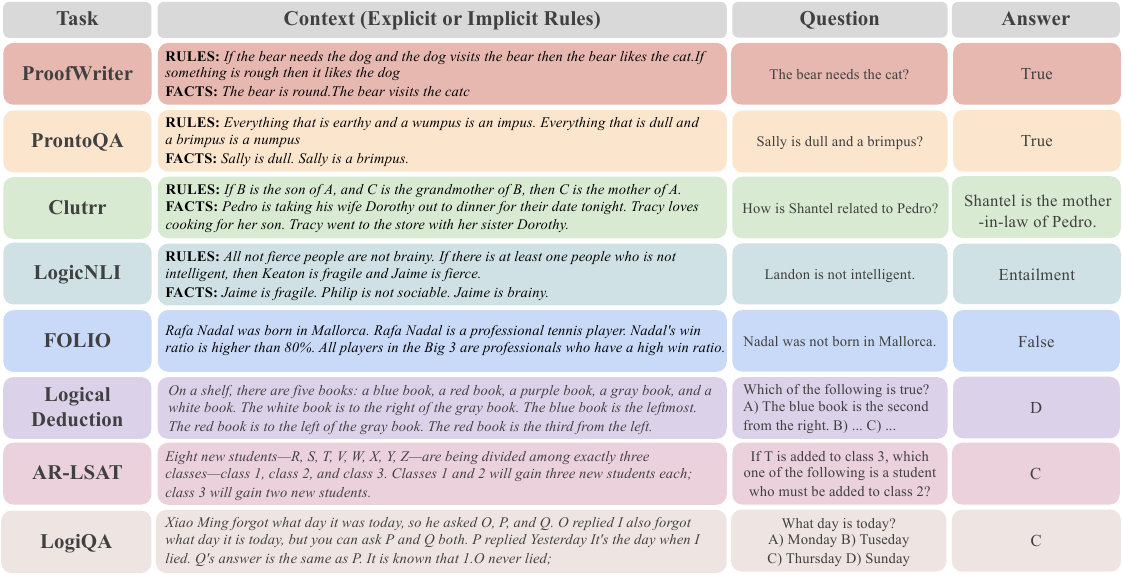}}
  \caption{Demonstration overview of \ourdata.}
  \label{fig:training_data_examples}
  \vspace{-10pt}
\end{figure*}

%% file: table/ood_bench_statistics.tex
\begin{wraptable}{r}{0.375\textwidth}
    \small
    % \vspace{-1mm}
    \caption{OOD benchmarks statistics.}
    \begin{tabular}{ccc}
        \toprule
        \textbf{OOD Test} & \textbf{Examples} & \textbf{Levels} \\
        \midrule
        BBH & 750 & \textcolor{darkgreen}{\ding{51}} \\
        BBEH & 400 & \textcolor{darkgreen}{\ding{51}} \\
        ProverQA & 1,500 & \textcolor{darkgreen}{\ding{51}} \\
        \bottomrule
    \end{tabular}
    \vspace{-4mm}
    \label{tab:ood_statistics}
\end{wraptable}

%% file: table/main_results.tex
\begin{table}[t!]
\caption{Comparison with all baselines on eight ID benchmarks. \ourmethod significantly outperforms most of other methods. Average is the macro mean across all samples of domains.}
\setlength{\tabcolsep}{2pt}
\resizebox{\linewidth}{!}{
\begin{tabular}{@{}lcccccccccc@{}}
\toprule
 & \multicolumn{1}{c}{Induction} & \multicolumn{2}{c}{Deduction} & \multicolumn{2}{c}{FOL} & \multicolumn{3}{c}{Others} & \multicolumn{2}{c}{\multirow{-1.9}{*}{\multirowcell{3.2}{\hspace{0.1cm}Avg. \\\hspace{0.1cm} Results}}} \\\cmidrule(lr){2-2}\cmidrule(lr){3-4}\cmidrule(lr){5-6}\cmidrule(lr){7-9}
 & \multicolumn{1}{c}{Clutrr} & \multicolumn{1}{c}{ProntoQA} & \multicolumn{1}{c}{ProofWriter} & \multicolumn{1}{c}{FOLIO} & \multicolumn{1}{c}{LogicNLI} & \multicolumn{1}{c}{AR-LSAT} & \multicolumn{1}{c}{Logic. Dedu.} & \multicolumn{1}{c}{LogiQA} \\\midrule
\textbf{\textsc{Prior RBRs}} & \multicolumn{1}{l}{} & \multicolumn{1}{l}{}    & \multicolumn{1}{l}{}   & \multicolumn{1}{l}{} & \multicolumn{1}{l}{}    & \multicolumn{1}{l}{}   & \multicolumn{1}{l}{}  & \multicolumn{1}{l}{} \\ 
HtT \citep{zhu2023large}    & 40.3    &   92.0     &  88.0     &  71.0  & 54.0  &   97.0    &   100.0   &  79.1   & $\hspace{0.20cm}\text{77.7}$   \\
RGFB \citep{diallo2025rule} &  31.3      &  94.0    &   88.0   &  74.0      & 55.0     & 95.0     &  100.0   & 79.1 &  $\hspace{0.20cm}\text{77.1}$   \\
Chain-of-Logic \citep{servantez-etal-2024-chain} & 44.8    &   91.0     &   92.0    &  80.0   &   54.0     & 97.0      & 100.0    &  80.6  &  $\hspace{0.20cm}\text{80.0}$   \\\midrule
\textbf{\textsc{Frontier Reasoners}} & \multicolumn{1}{l}{} & \multicolumn{1}{l}{}    & \multicolumn{1}{l}{}   & \multicolumn{1}{l}{} & \multicolumn{1}{l}{}    & \multicolumn{1}{l}{}   & \multicolumn{1}{l}{}  & \multicolumn{1}{l}{}  \\
OpenAI o1 \citep{jaech2024openai} & 52.2 &  91.0   &   91.0   &  77.0      &  60.0     &  98.0    & 88.0    &  82.1 & $\hspace{0.20cm}\text{79.9}$ \\
OpenAI o3-mini \citep{ZhangOpenAIOS} & 40.3 & 94.0 & 93.0 & 74.0 & 55.0 & 96.3 & 100.0 & 77.6 & $\hspace{0.20cm}\text{78.8}$ \\
Claude-3.7-Sonnet \citep{anthropic-claude-3-7-sonnet} & 65.7 & 92.8 & 90.0 & 
74.7 & 58.0 & 76.2 & 97.0 & 81.5 & $\hspace{0.20cm}\text{79.5}$ \\
DeepSeek-R1 \citep{guo2025deepseek} & 71.6 & 40.0 & 27.0 & 72.7 & 
49.0 & 89.7 & 98.3 & 85.0 & $\hspace{0.20cm}\text{66.7}$ \\\midrule
\textbf{\textsc{Behavioral Cloning}} & \multicolumn{1}{l}{} & \multicolumn{1}{l}{}    & \multicolumn{1}{l}{}   & \multicolumn{1}{l}{} & \multicolumn{1}{l}{}    & \multicolumn{1}{l}{}   & \multicolumn{1}{l}{}  & \multicolumn{1}{l}{}  \\
SFT w/o CoT \citep{wei2022finetuned} & 37.5 & 96.0 & 88.8 & 73.4 & 74.8 & 37.5 & 85.9 & 76.1 & $\hspace{0.20cm}\text{71.2}$ \\
SFT w/ Short CoT \citep{yeo2025demystifying} & 77.6 & 92.6 & 87.0 & 82.9 & 73.8 & 54.8 & 87.6 & 88.0 & $\hspace{0.20cm}\text{80.9}$ \\
SFT w/ Long CoT \citep{yeo2025demystifying} & 83.5 & 95.6 & 89.2 & 83.4 & 76.6 & 68.6 & 79.6 & 79.1 & $\hspace{0.20cm}\text{81.9}$ \\\midrule
\textbf{\textsc{Advanced RLVRs}}& \multicolumn{1}{l}{} & \multicolumn{1}{l}{}    & \multicolumn{1}{l}{}   & \multicolumn{1}{l}{} & \multicolumn{1}{l}{}    & \multicolumn{1}{l}{}   & \multicolumn{1}{l}{}  & \multicolumn{1}{l}{}  \\
GRPO \citep{shao2024deepseekmath} & 73.1 & 95.4 & 96.4 & 72.3 & 66.6 & 36.3 & 90.3 & 70.1 & $\hspace{0.20cm}\text{75.0}$ \\
Dr. GRPO \citep{liu2025understanding} & 68.6 & 96.0 & 95.6 & 73.9 & 75.4 & 32.1 & 84.3 & 65.6 & $\hspace{0.20cm}\text{73.9}$ \\
DAPO \citep{yu2025dapo} & 86.5 & 96.0 & 94.8 & 80.9 & 65.8 & 40.0 & 95.3 & 74.6 & $\hspace{0.20cm}\text{79.2}$ \\\midrule
\textbf{\textsc{Curriculum Learning}} & \multicolumn{1}{l}{} & \multicolumn{1}{l}{}    & \multicolumn{1}{l}{}   & \multicolumn{1}{l}{} & \multicolumn{1}{l}{}    & \multicolumn{1}{l}{}   & \multicolumn{1}{l}{}  & \multicolumn{1}{l}{}  \\
Data-balance RL \citep{parashar2025curriculum} & 86.5 & 95.8 & 95.6 & 76.8 & 64.4 & 45.6 & 95.3 & 73.1 & $\hspace{0.20cm}\text{79.1}$ \\
Easy-to-hard RL \citep{parashar2025curriculum} & 88.0 & 96.2 & 96.8 & 78.9 & 66.6 & 46.3 & 96.0 & 74.6 & $\hspace{0.20cm}\text{80.4}$ \\
\textsc{AdaRFT} \citep{shi2025efficient} & 92.5 & 96.0 & 97.4 & 81.8 & 64.4 & 44.6 & 96.6 & 80.5 & $\hspace{0.20cm}\text{81.7}$ \\\midrule
\textbf{\ourmethod (Ours)}& \multicolumn{1}{l}{} & \multicolumn{1}{l}{}    & \multicolumn{1}{l}{}   & \multicolumn{1}{l}{} & \multicolumn{1}{l}{}    & \multicolumn{1}{l}{}   & \multicolumn{1}{l}{}  & \multicolumn{1}{l}{}  \\
% \textsc{TrippleR-1.7B} &  &  &  &  &  &  &  & &  \\
\ourmodel & $\hspace{0.38cm}\text{82.0}_{0.4}$ & $\hspace{0.38cm}\text{95.0}_{0.6}$ & $\hspace{0.38cm}\text{96.3}_{0.3}$ & $\hspace{0.38cm}\text{78.9}_{0.8}$ & $\hspace{0.38cm}\text{66.6}_{0.4}$ & $\hspace{0.38cm}\text{38.6}_{0.5}$ & $\hspace{0.38cm}\text{96.3}_{0.2}$ & $\hspace{0.38cm}\text{80.5}_{0.7}$ & $\hspace{0.54cm}\text{79.2}_{0.6}$ \\
\ourmodelbest & $\hspace{0.4cm}\textbf{95.5}_{0.3}$ & $\hspace{0.38cm}\textbf{96.4}_{0.4}$ & $\hspace{0.38cm}\textbf{97.0}_{0.2}$ & $\hspace{0.38cm}\textbf{84.7}_{0.6}$ & $\hspace{0.38cm}\text{70.4}_{0.1}$ & $\hspace{0.38cm}\text{46.8}_{0.3}$ & $\hspace{0.38cm}\text{98.3}_{0.4}$ & $\hspace{0.38cm}\text{83.5}_{0.3}$ & $\hspace{0.54cm}\textbf{84.0}_{0.5}$ \\\bottomrule
\end{tabular}}
\vspace{-3mm}
\label{tab:id_performance}
\end{table}

%% file: table/improvement_comparison.tex
\begin{wraptable}{r}{0.43\textwidth}
    \small
    % \vspace{3mm}
    \caption{Comparison of average improvement. \greenbox{\%} denotes ID and \redbox{\%} denotes OOD performance, respectively. Unlike the task-focused settings in \S\ref{subsec:experiment_setup}, $\dagger$ indicates a full-set mix supervised training to obtain stronger OOD performance for SFT.}
    \begin{tabular}{lcc}
        \toprule
        \textbf{Model} & \textbf{Pass@1} & \textbf{Avg. $\Delta$} \\
        \midrule
        Qwen3-8B & \greenbox{27.4}/\redbox{34.2} & $-$ \\
        \midrule
        \hspace{0.1cm}$\text{+}$ SFT$^\dagger$ & \greenbox{81.9}/\redbox{66.6}  & \greenbox{54.5}/\redbox{34.4}  \\
        \hspace{0.1cm}$\text{+}$ GRPO & \greenbox{75.0}/\redbox{75.8} & \greenbox{47.6}/\redbox{41.6} \\
        \hspace{0.1cm}$\text{+}$ \textbf{Ours} & \greenbox{\textbf{84.0}}/\redbox{\textbf{81.7}} & \greenbox{\textbf{56.6}}/\redbox{\textbf{47.5}} \\
        \bottomrule
    \end{tabular}
    \vspace{-3mm}
    \label{tab:comparison_improvement}
\end{wraptable}

%% file: figure/domain_weight_rewards_dynamics.tex
\begin{figure*}[!ht]
  \centering
  \resizebox{1.0\linewidth}{!}{\includegraphics{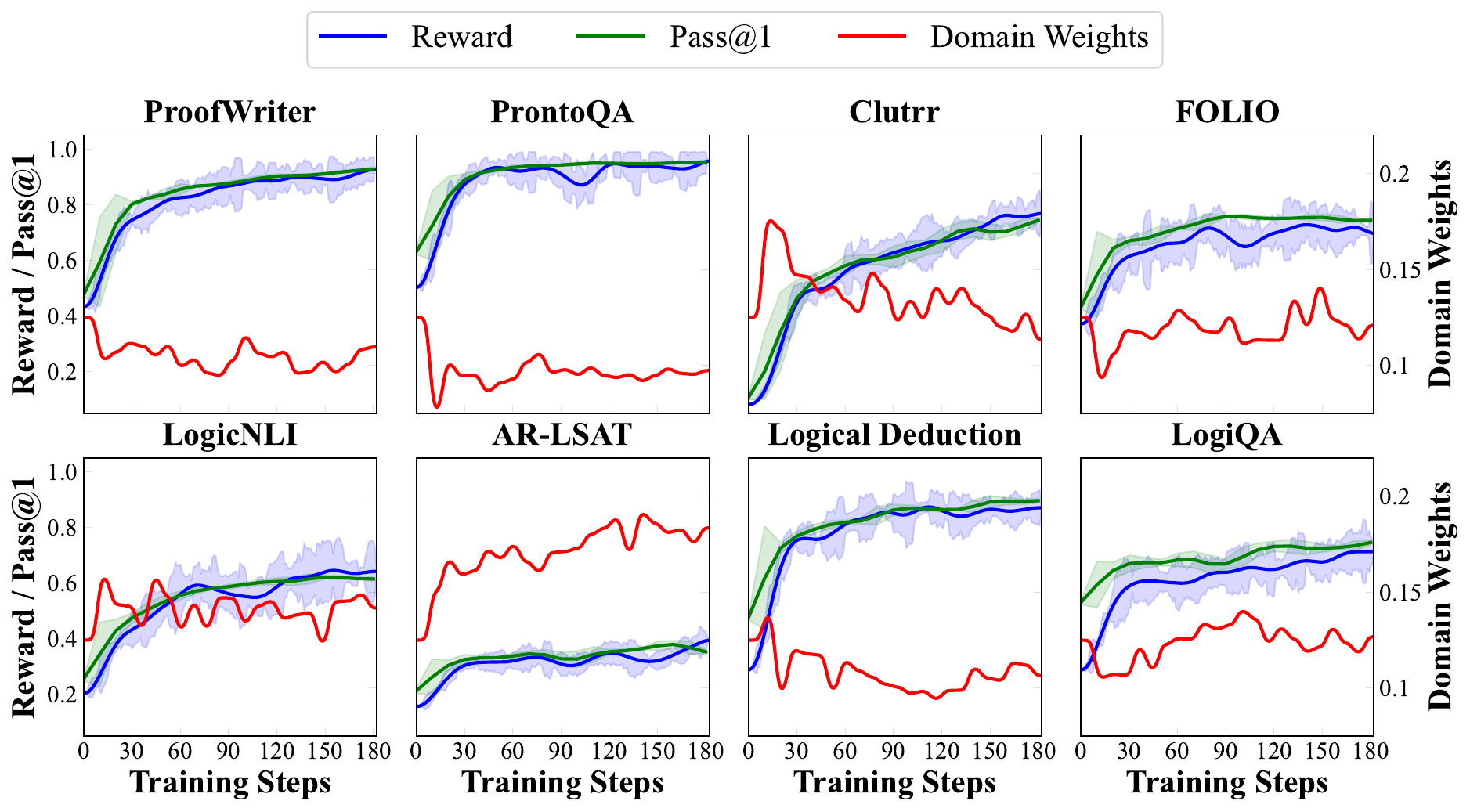}}
  \caption{Learning dynamics by domains. ``\setulcolor{blue}\ul{Reward}'' represents the training reward obtained from tasks and ``\setulcolor{OliveGreen}\ul{Pass@1}'' denotes validation pass@1 performance. We employ exponential moving average smoothing for clearly displaying the curves ``\setulcolor{blue}\ul{Reward}'', ``\setulcolor{OliveGreen}\ul{Pass@1}'', and ``\setulcolor{red}\ul{Domain Weights}''.}
  \label{fig:weight_rewards_dynamics}
\end{figure*}

%% file: figure/task_mixing_recipe.tex
% \begin{figure}[t]
%   \centering
%   \resizebox{1\linewidth}{!}{\includegraphics{figure/raw/task_mixing_recipe.pdf}}
%   \caption{Comparison of performance on different task mixing recipes.}
%   \label{fig:task_mixing_recipe}
% \end{figure}
% \begin{figure}[t]
%   \centering
%   \resizebox{1\linewidth}{!}{\includegraphics{figure/raw/training_datasize_impact.pdf}}
%   \caption{Comparison of the impact on scaling training data size.}
%   \label{fig:training_datasize_impact}
% \end{figure}

\begin{figure}[!ht]
    \vspace{-3mm}
    \centering
    \begin{minipage}[t]{0.33\linewidth}
        \centering
        \includegraphics[width=0.98\linewidth]{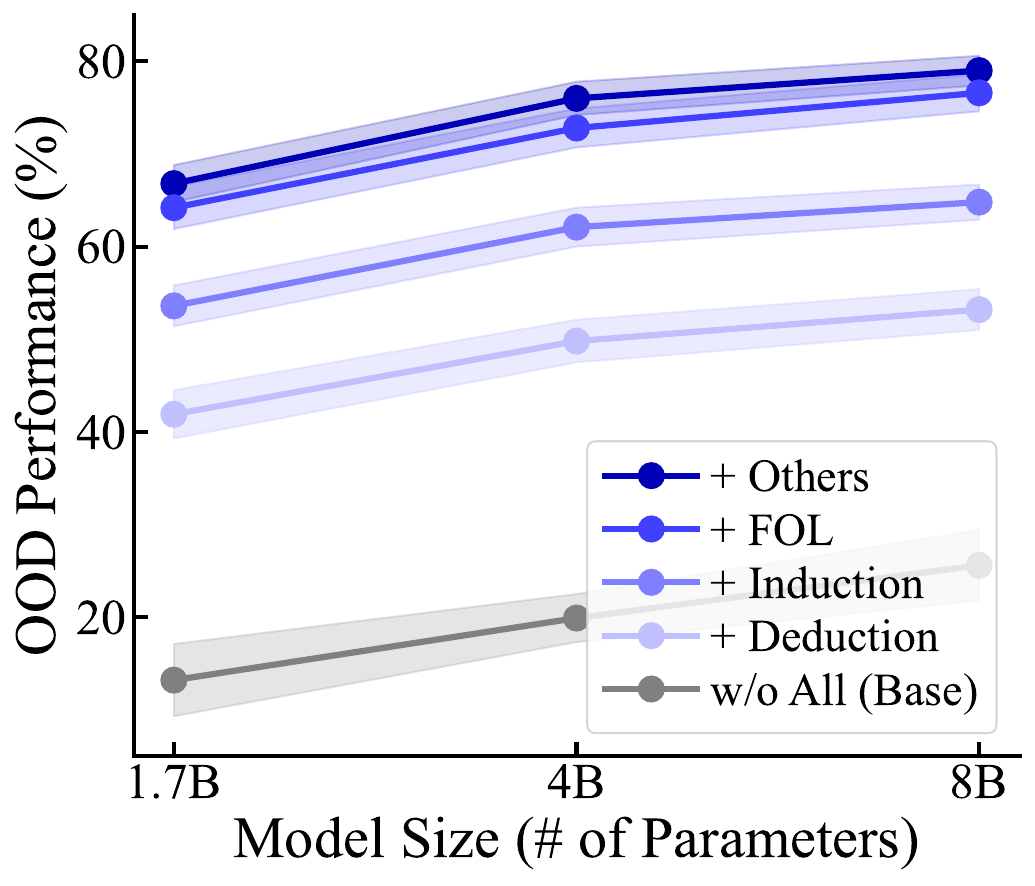}
        \captionsetup{width=0.9\linewidth}
        \caption{Impact on incremental task mixing recipes.}
        \label{fig:task_mixing_recipe}
    \end{minipage}
    \begin{minipage}[t]{0.33\linewidth}
        \centering
        \includegraphics[width=0.98\linewidth]{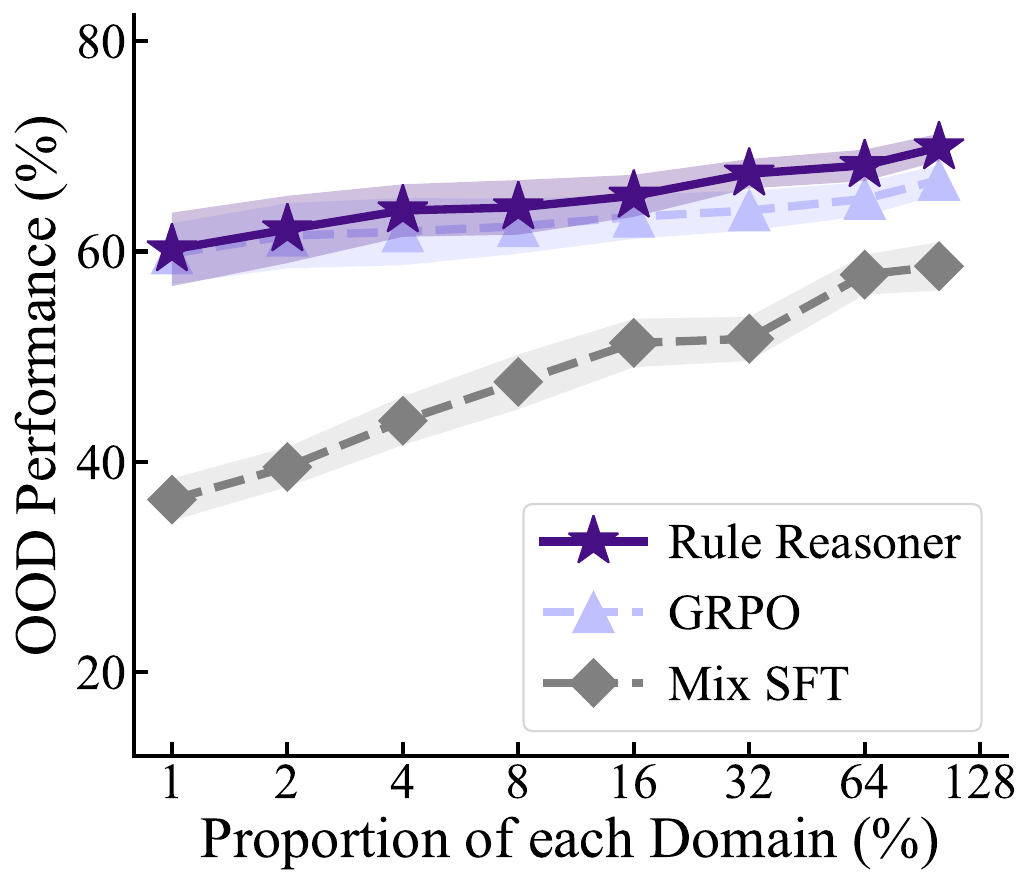}
        \captionsetup{width=0.9\linewidth}
        \caption{Impact on training sample efficiency.}
        \label{fig:sample_efficiency}
    \end{minipage}
    \begin{minipage}[t]{0.32\linewidth}
        \centering
        \includegraphics[width=1\linewidth]{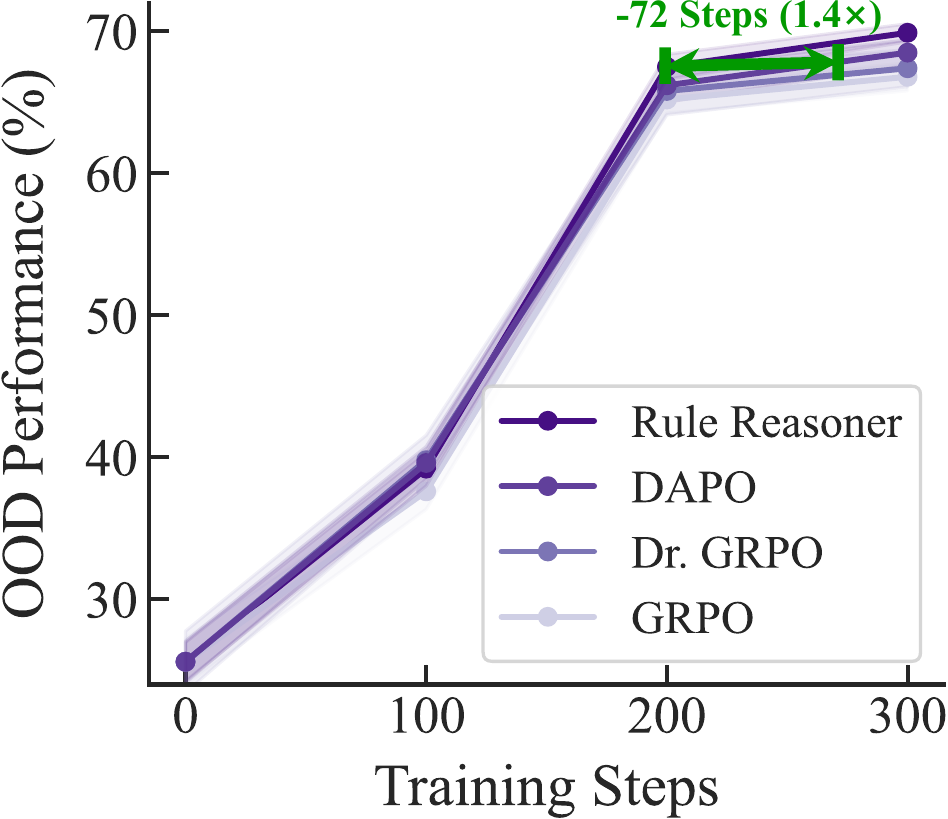}
        \captionsetup{width=0.9\linewidth}
        \caption{Comparisons on different RLVR methods.}
        \label{fig:rlvr_comparison}
    \end{minipage}
     \vspace{-1mm}
\end{figure}

%% file: table/case_studies.tex
\begin{table}[ht]
\vspace{-1em}
\caption{Reasoning chains before and after training by \ourmethod for comparison.}
\label{tab:case-study}
 \setlength\tabcolsep{2.5pt}
\begin{center}
\begin{tabular}{m{2.2cm}m{2.5cm}m{4.3cm}m{4.3cm}}
\toprule
\multicolumn{1}{l}{\bf\small Case Type} & \bf\small \makecell[c]{Input}  & \bf\small \makecell[c]{Before Training}  & \bf\small \makecell[c]{After Training}\\ \midrule
\footnotesize \makecell[c]{Explicit Rules \\ (ProofWriter)} & \tiny \textbf{Rules}: If something is smart then it is furry. If something is cold it is green. Cold, smart things are green. All furry things are white. If something is green, it is smart. All furry things are green. All cold, green things are rough. Cold things are green. \newline \textbf{Facts}: Dave is blue. Dave is cold. Dave is green. \newline \textcolor{DodgerBlue2}{\textbf{Question}: Dave is furry?} & \tiny Step 1: Analyze the given rules and facts. Rule 1: If something is smart, then it is furry.  ... Fact 1: Dave is blue. Fact 2: Dave is cold. ... Step 2: Determine if Dave is furry based on the given rules and facts. From Fact 2 and Rule 2, we know that Dave is green. From Fact 2 and Rule 8, we know that Dave is green. 
\textcolor{BrickRed}{However, we also have Rule 4 stating that all furry things are white, and Dave is blue (Fact 1). This creates a contradiction, as Dave cannot be both furry and blue.} Since we have a contradiction, we cannot determine if Dave is furry. The answer is Unknown. \newline \textcolor{BrickRed}{Answer: Unknown.} \textcolor{darkred}{\ding{55}} & \tiny 
Step-by-Step Reasoning: 1. Dave is cold. From Rule 8: If something is cold, then it is green. Since Dave is cold, Dave is green. This is consistent with the given fact that Dave is green. \textcolor{OliveGreen}{From Rule 5: If something is green, then it is smart. Since Dave is green, Dave is smart. From Rule 1: If something is smart, then it is furry. Since Dave is smart, Dave is furry.} From Rule 6: All furry things are green. This is consistent with the given fact that Dave is green. 3. Conclusion: Since Dave is smart (from Rule 5), and from Rule 1, if something is smart, then it is furry, we can conclude Dave is furry. \newline \textcolor{OliveGreen}{Answer: True.} \textcolor{darkgreen}{\ding{51}} \\
\midrule
\footnotesize \makecell[c]{Implicit Rules \\ (BBH)} & \tiny \textbf{Context}: The following paragraphs each describe a set of three objects arranged in a fixed order. The statements are logically consistent within each paragraph. On a branch, there are three birds: a robin, an owl, and a hawk. The robin is to the right of the hawk. The owl is the leftmost. \newline \textbf{Options}: (A) The robin is the rightmost (B) The owl is the rightmost (C) The hawk is the rightmost. \newline \textcolor{DodgerBlue2}{\textbf{Question}: Which of the following options is true?} & \tiny To answer this question, let's analyze the information provided in the context step by step: 1. There are three birds on a branch: ... Now, let's evaluate the options: (A) The robin is the rightmost: \textcolor{BrickRed}{This option is not true because the robin is to the right of the hawk, but not necessarily the rightmost bird.} (B) The owl is the rightmost: This option is not true because the owl is the leftmost bird, not the rightmost. 
\textcolor{BrickRed}{Based on the information provided, none of the options (A, B, or C) are true.} However, since we must choose one of the given options, we can conclude that the correct answer is: (B) Please note that this answer is not entirely accurate, as the owl is actually the leftmost bird. The question might have a mistake or missing information. \newline \textcolor{BrickRed}{Answer: B.} \textcolor{darkred}{\ding{55}} & \tiny To determine the correct order of the three birds on the branch, let's analyze the information provided: 1. The owl is the leftmost. This means the owl is in the first position on the branch. 2. The robin is to the right of the hawk. This means the robin comes after the hawk in the order. Now, let's determine the positions of the hawk and the robin: Since the owl is the leftmost, the owl must be in the first position. 
The robin is to the right of the hawk, so the hawk must be in the second position, and the robin must be in the third position. \textcolor{OliveGreen}{So, the order from left to right is: 1. Owl 2. Hawk 3. Robin} Now, let's evaluate the options: (A) The robin is the rightmost. 
So, this statement is true. (B) The owl is the rightmost. ... So, this statement is false. 
%Therefore, the correct option is (A) The robin is the rightmost. 
\newline \textcolor{OliveGreen}{Answer: A.} \textcolor{darkgreen}{\ding{51}} \\
\bottomrule
\end{tabular}
\end{center}
% \vspace{-2em}
\end{table}

%% file: table/data_stat.tex
\begin{table*}[htbp]
\setlength\tabcolsep{1.8pt}
\begin{center}
\caption{Data statistics of curated tasks. $\dagger$ denotes it can be deemed as deduction reasoning since we provide rules explicitly. The abbr. in the table indicate Modus Ponens (MP), Universal Instantiation (UI), Hypothetical Syllogism (HS), Disjunctive Syllogism (DS), Modus Tollens (MT), respectively. ``FOL'', ``AR'', ``CS'', and ``CCR'' denotes First-Order Logic, Analytical Reasoning, Constraint Satisfaction, and Categorical \& Conjunctive Reasoning, respectively. ``MC'' represents multiple choice.}
%The table contains the most representative rules of inference for each dataset.}
\begin{tabular}{lcccccc}
\toprule
\small \textbf{Dataset} & \bf\small \makecell{\# Train/Test} & \bf\small \makecell{Task\\ Format} & \bf\small \makecell{Reasoning \\ Form} & \bf\small \makecell{Reasoning \\ Depth} & \bf\small \makecell{Fiction \\ Rule}  & \bf\small \makecell{Rule of \\ Inference} \\
\midrule
\small ProofWriter \citep{tafjord-etal-2021-proofwriter} & \small 7,997/500  & \small Boolean & \small Deduction & \small [0, 5] &  \textcolor{darkgreen}{\ding{51}} & \small \makecell{MP, UI} \\
\small ProntoQA \citep{saparov2023language} & \small 8,000/500  & \small Boolean & \small Deduction  &  \small \{1, 3, 5\}  &  \textcolor{darkgreen}{\ding{51}} &  \small \makecell{UI, Conjunction \\  Simplification}  \\
% \small RuleTakers & \small  & \small Boolean & \small Synthetic & \small  Deduction &  & \textcolor{darkgreen}{\ding{51}}  &  \\
\small Clutrr \citep{sinha-etal-2019-clutrr} & \small 268/67  & \small Free Text & \small Induction$^\dagger$ & $-$ & \textcolor{darkred}{\ding{55}} & \small  HS \\
\small FOLIO \citep{han-etal-2024-folio} & \small 1,208/242  & \small MC & \small FOL &  \small [0, 7] &  \textcolor{darkgreen}{\ding{51}} & \small MT, DS, UI  \\
\small LogicNLI \citep{tian-etal-2021-diagnosing} & \small 8,000/500  & \small MC & \small FOL  & \small [1, 5] &  \textcolor{darkgreen}{\ding{51}} & \small MP, MT\\
\small AR-LSAT \citep{zhong-etal-2022-analytical} & \small 1,636/410  & \small  MC & \small AR & $-$ &  \textcolor{darkgreen}{\ding{51}} & \small MP, MT \\
\small Logic. Dedu. \citep{xu-etal-2024-faithful} & \small 1,200/300  & \small  MC & \small CS  & \small \{1, 3, 5\}  & \textcolor{darkgreen}{\ding{51}}  & \small MP, MT \\
\small LogiQA \citep{logiqa2023} & \small 264/67  & \small MC & \small  \makecell{CCR}  & $-$ & \textcolor{darkgreen}{\ding{51}}  & \small MP, MT \\
%\midrule
%\small Total & & & & & \\
\bottomrule
\end{tabular} 
\label{tab:data_stat}
% \vspace{-3mm}
\end{center}
\end{table*}

%% file: figure/challenging_rule_settings.tex
% Note: The wrapfigure may float if placed at the top of a page, causing a warning. This is harmless.
 \begin{wrapfigure}{r}{0.35\textwidth}
    %\centering
    \vspace{-8mm}
    \centering
    \includegraphics[width=1\linewidth]{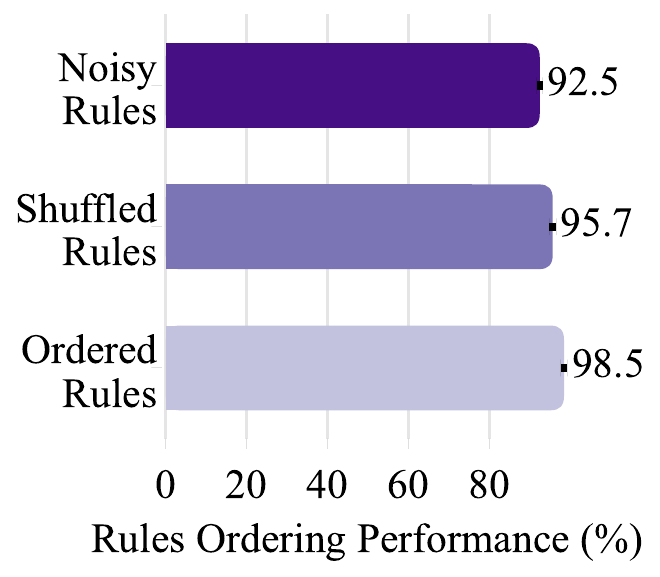}
    \captionsetup{width=1\linewidth}
    \caption{Comparison of performance on challenging rule settings.}
    \label{fig:rule_settings}
    % \end{minipage}
    \vspace{-3mm}
\end{wrapfigure}

%% file: figure/varying_task_complexity.tex
% \begin{figure}[t]
%   \centering
%   \resizebox{1\linewidth}{!}{\includegraphics{figure/raw/varying_task_complexity_performance.pdf}}
%   \caption{Comparison of performance on varying task complexity.}
%   \label{fig:varying-complexity-results}
% \end{figure}

% \begin{figure}[t]
%   \centering
%   \resizebox{1\linewidth}{!}{\includegraphics{figure/raw/challenging_rule_settings.pdf}}
%   \caption{Comparison of performance on different challeging rule settings.}
%   \label{fig:challenging-rule-settings}
% \end{figure}

% \begin{figure}[!ht]
%     \centering
%     \begin{minipage}[t]{0.45\linewidth}
%         \centering
%         \includegraphics[width=0.9\linewidth]{figure/raw/varying_task_complexity_performance.pdf}
%         \captionsetup{width=0.9\linewidth}
%         \caption{Comparison of performance on varying task complexity. \yang{Add baseline for comparison.}}
%         \label{fig:varying-complexity-results}
%     \end{minipage}
%     \begin{minipage}[t]{0.45\linewidth}
%         \centering
%         \includegraphics[width=0.9\linewidth]{figure/raw/challenging_rule_settings.pdf}
%         \captionsetup{width=0.9\linewidth}
%         \caption{Comparison of performance on different challeging rule settings.}
%     \end{minipage}
%  \end{figure}

\begin{wrapfigure}{r}{0.35\textwidth}
  \vspace{-3mm}
  \centering
  \resizebox{1\linewidth}{!}{\includegraphics{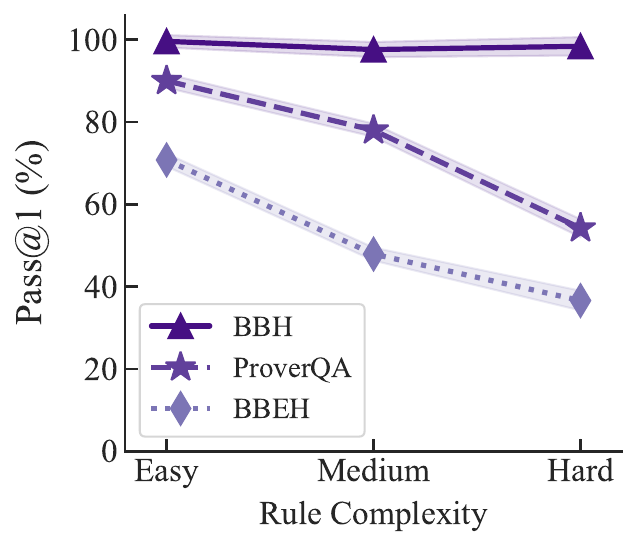}}
  \caption{Comparison of performance on varying task complexity.}
  \label{fig:varying-complexity-results}
  \vspace{-3mm}
\end{wrapfigure}

%% file: figure/test_time_scaling.tex
\begin{figure*}[!ht]
  \centering
  \resizebox{1\linewidth}{!}{\includegraphics{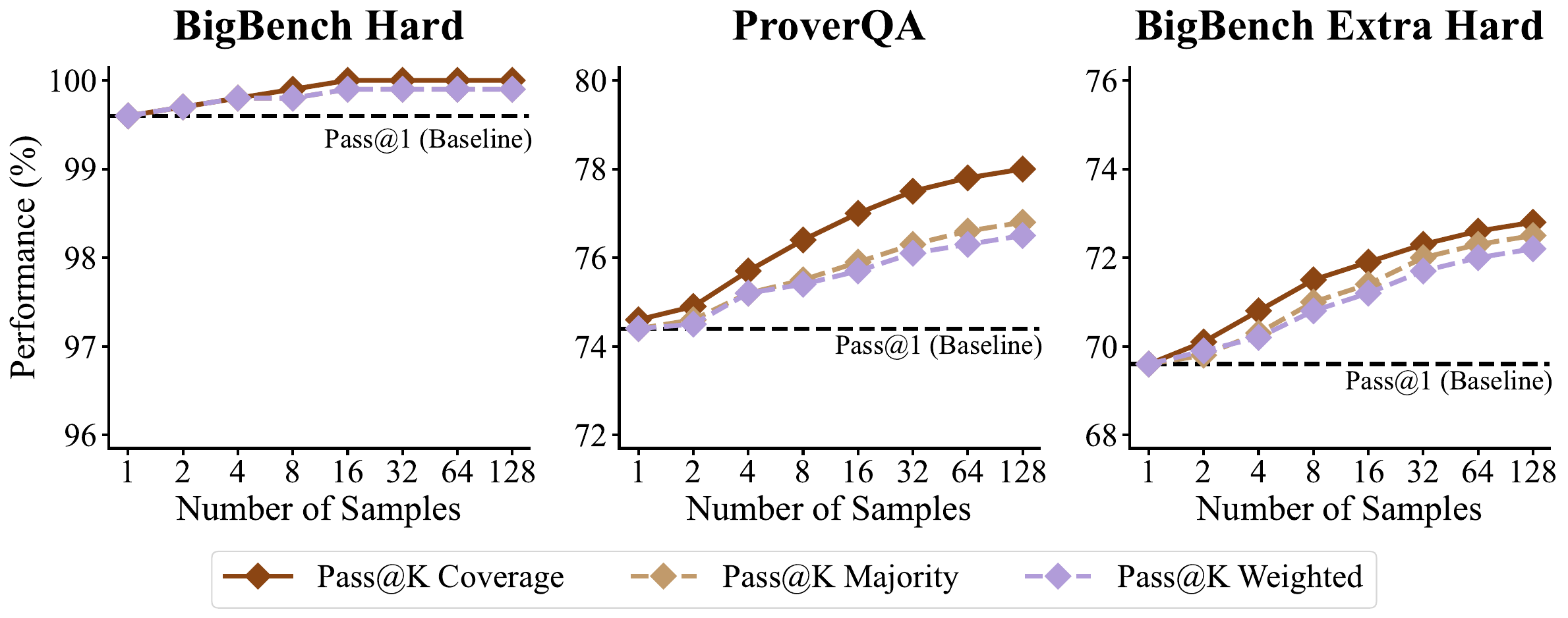}}
  \caption{Comparison of OOD performance in parallel test-time scaling methods.}
  \label{fig:test-time-scaling}
\end{figure*}

%% file: table/more_baseline_comparison.tex
\begin{table}[h]
\centering
\begin{tabular}{lcccc}
\toprule
\textbf{Model} & \textbf{BBH} & \textbf{ProverQA} & \textbf{BBEH} & \textbf{Average ($\Delta$)} \\
\midrule
Qwen3-8B-Base & 21.2 & 13.4 & 8.0 & 14.2 (-) \\
\midrule
\hspace{0.1cm}+ DAPO (w/ dynamic sampling) & 95.4 & 68.8 & 62.0 & 75.4 (+61.2) \\
\hspace{0.1cm}+ \textsc{AdaRFT} & 96.4 & 73.4 & 64.5 & 78.1 (+63.9) \\
\hspace{0.1cm}+ \ouralgo (Ours) & \textbf{99.6} & \textbf{76.6} & \textbf{68.9} & \textbf{81.7 (+67.5)} \\
\bottomrule
\end{tabular}
\caption{Performance comparison across different methods on BBH, ProverQA, and BBEH.}
\label{tab:more_baselines_comparison}
\end{table}

%% file: table/hyperparameters_analyses.tex
\begin{table*}[!ht]
    \centering
    \caption{Hyperparameter sensitivity analysis for different parameters}
    \small
    \begin{tabular}{c}
      \toprule
      \textbf{Hyperparameter Sensitivity Analyses}\\
      \midrule \\
    \vspace{0mm}
    \begin{tabular}{cc}
\midrule \textbf{$\alpha$} & \textbf{Avg. Pass@1 (OOD)} \\
\midrule 0.1 & 68.6 \\
    0.3 & 69.3 \\
    0.5 & 70.4 \\
    0.8 & 70.1 \\
    1.0 & 69.8 \\
\midrule
\end{tabular}
    \vspace{3mm}
    \begin{tabular}{cc}
    \midrule \textbf{$\tau$} & \textbf{Avg. Pass@1 (OOD)} \\
    \midrule 0.1 & 67.6 \\
    0.3 & 69.4 \\
    0.5 & 70.1 \\
    0.8 & 70.4 \\
    1.0 & 69.7 \\
    \midrule
    \end{tabular}
    \vspace{-1mm}
    \begin{tabular}{cc}
    \midrule \textbf{$\epsilon$} & \textbf{Avg. Pass@1 (OOD)} \\
    \midrule 0.1 & 70.4 \\
    0.3 & 70.2 \\
    0.5 & 69.9 \\
    0.8 & 69.3 \\
    1.0 & 68.9 \\
    \midrule
    \end{tabular}
    \vspace{1mm} \\
    \bottomrule
    \end{tabular} 
    \label{tab:hyperparameter_analysis}
\end{table*}

%% file: table/more_unseen_tasks_eval.tex
\begin{table}[ht]
\centering
\caption{Additional evaluation results on AIME 2025, GPQA (Diamond), and Coin Flip.}
\label{tab:additional_ood_results}
\begin{tabular}{lcccc}
\toprule
\textbf{Model} & \textbf{AIME 2025} & \textbf{GPQA (Diamond)} & \textbf{Coin Flip} & \textbf{Average ($\Delta$)} \\
\midrule
Qwen3-8B-Base & 3.3 & 16.6 & 44.7 & 21.5 (-) \\
\midrule
\hspace{0.1cm}+ \ouralgo (Ours) & \textbf{23.3} & \textbf{44.9} & \textbf{95.1} & \textbf{54.4 (+32.9)} \\
\bottomrule
\end{tabular}
\end{table}

%% file: table/high_difficulty_tasks.tex
\begin{table}[ht]
\centering
\caption{Performance comparison of Pass@1 with and without \ouralgo across different training steps.}
\label{tab:high_difficulty_results}
\begin{tabular}{lcc}
\toprule
\textbf{Training Steps} & \textbf{w/o \ouralgo} & \textbf{w/ \ouralgo ($\Delta$)} \\
\midrule
50 & 30.0 & 32.6 (+2.6) \\
100 & 30.9 & 36.8 (+5.9) \\
150 & 34.3 & 37.0 (+2.7) \\
200 & 35.1 & 38.5 (+3.4) \\
250 & 34.8 & 39.0 (+4.2) \\
300 & 38.0 & 41.7 (+3.7) \\
\bottomrule
\end{tabular}
\end{table}

%% file: table/offline_curriculum_sft_rl_results.tex
\begin{table}[ht]
\centering
\caption{Performance comparison across different benchmarks and training methods}
\label{tab:offline_curriculumn_sft_rl_results}
\begin{tabular}{lcccc}
\toprule
& \textbf{BBH} & \textbf{ProverQA} & \textbf{BBEH} & \textbf{Average ($\Delta$)} \\
\midrule
Qwen3-8B-Base & 21.2 & 13.4 & 8.0 & 14.2 (-) \\
+ Easy-to-hard SFT & 89.2 & 31.8 & 17.7 & 46.2 (+32.0) \\
+ Easy-to-hard RL & 96.0 & 72.2 & 63.5 & 77.2 (+63.0) \\
+ Data-balance RL & 94.8 & 69.4 & 60.5 & 74.9 (+60.7) \\
+ \ouralgo (Ours) & \textbf{99.6} & \textbf{76.6} & \textbf{68.9} & \textbf{81.7 (+67.5)} \\
\midrule
& \textbf{AIME 2025} & \textbf{GPQA (Diamond)} & \textbf{Coin Flip} & \textbf{Average ($\Delta$)} \\
\midrule
Qwen3-8B-Base & 3.3 & 16.6 & 44.7 & 21.5 (-) \\
+ Easy-to-hard SFT & 6.6 & 33.8 & 48.7 & 29.7 (+8.2) \\
+ Easy-to-hard RL & 20.0 & 39.8 & 91.4 & 50.4 (+28.9) \\
+ Data-balance RL & 16.7 & 38.3 & 92.8 & 49.2 (+27.7) \\
+ \ouralgo (Ours) & \textbf{23.3} & \textbf{44.9} & \textbf{95.1} & \textbf{54.4 (+32.9)} \\
\bottomrule
\end{tabular}
\end{table}

%% file: table/mix_math_rule_rl.tex
\begin{table}[ht]
\centering
\caption{Performance comparison on AIME 2025 and ProverQA across different training datasets.}
\label{tab:mix_math_rule_rl}
\begin{tabular}{lcc}
\toprule
& \textbf{AIME 2025} & \textbf{ProverQA} \\
\midrule
Qwen3-8B-Base & 3.3 (-) & 13.4 (-) \\
\midrule
+ AIME 1983-2024 & 63.3 (+60.0) & 38.8 (+25.4) \\
+ \ourdata (Ours) & 23.3 (+20.0) & 73.6 (+60.2) \\
+ AIME 1983-2024 + \ourdata (Ours) & \textbf{66.7 (+63.4)} & \textbf{80.4 (+67.0)} \\
\bottomrule
\end{tabular}
\end{table}

%% file: table/hyperparameters.tex
\begin{table*}[!ht]
    \centering
    \caption{Hyper-parameters of \ourmodel and \ourmodelbest on-policy RL training and inference.}
    \small
    \begin{tabular}{c}
      \toprule
      \textbf{Computational Infrastructure}\\4 \texttimes\ A100-SXM4-80GB GPU \\ 
      \midrule \\
    \vspace{3mm}
    \begin{tabular}{ll}
\midrule \textbf{Hyperparameter} & \textbf{Assignment} \\

\midrule Base model & \ourmodel \\& (Qwen3-4B-Base) \\
    Training strategy & \textsc{FSDP} \citep{pytorchfsdp} \\
    Training epochs & 3 \\
    Training precision & bfloat16 \\
    Mini batch size & 65,536 tokens \\
    Micro batch size & 16,384 tokens \\
    Max input length & 1,024 \\
    Max output length & 2,048 \\
    Max learning rate & $1\mathrm{e}-6$ \\
    Optimizer & AdamW \\
    Adam beta weights & $0.9,0.95$ \\
    KL loss coefficient & 0 \\
    Rollout per sample & 64 \\
    Clip ratio & 0.2 \\
    Gradient clipping & 1.0 \\
    Evaluation steps & 10 \\
    Save steps & 10 \\
    Training engine & PyTorch \cite{pytorch} \\
    \textcolor{ppppurple}{Decay factor $\alpha$} & \textcolor{ppppurple}{0.5} \\
    \textcolor{ppppurple}{Base reward $\epsilon$} & \textcolor{ppppurple}{0.1} \\
    \textcolor{ppppurple}{Scaling factor $\tau$} & \textcolor{ppppurple}{0.8} \\
    Rollout engine & vLLM \citep{kwon2023efficient} \\
    Rollout temperature & 0.8 \\
    Rollout top-p & 1 \\
    Random seeds & $\{21, 42, 84\}$ \\
\midrule
\end{tabular}
    \vspace{-3mm}
    \begin{tabular}{ll}
    \midrule \textbf{Hyperparameter} & \textbf{Assignment} \\
    \midrule Base model & \ourmodelbest \\& (Qwen3-8B-Base) \\
    Training strategy & \textsc{FSDP} \citep{pytorchfsdp} \\
    Training epochs & 3 \\
    Training precision & bfloat16 \\
    Mini batch size & 65,536 tokens \\
    Micro batch size & 16,384 tokens \\
    Max input length & 1,024 \\
    Max output length & 2,048 \\
    Max learning rate & $1\mathrm{e}-6$ \\
    Optimizer & AdamW \\
    Adam beta weights & $0.9,0.95$ \\
    KL loss coefficient & 0 \\
    Rollout per sample & 64 \\
    Clip ratio & 0.2 \\
    Gradient clipping & 1.0 \\
    Evaluation steps & 10 \\
    Save steps & 10 \\
    Training engine & PyTorch \citep{pytorch} \\
    \textcolor{ppppurple}{Decay factor $\alpha$} & \textcolor{ppppurple}{0.5} \\
    \textcolor{ppppurple}{Base reward $\epsilon$} & \textcolor{ppppurple}{0.1} \\
    \textcolor{ppppurple}{Scaling factor $\tau$} & \textcolor{ppppurple}{0.5} \\
    Rollout engine & vLLM \citep{kwon2023efficient} \\
    Rollout temperature & 0.8 \\
    Rollout top-p & 1 \\
    Random seeds & $\{21, 42, 84\}$ \\
    \midrule
    \end{tabular}
    \vspace{3mm} \\
    \bottomrule
    \end{tabular} 
    \label{tab:hyperparams}
\end{table*}

%% file: table/positive_case_studies.tex
\begin{table}[ht]
\caption{Positive examples of \ourmethod on various datasets.}
\label{tab:positive-case-study}
 \setlength\tabcolsep{2.5pt}
\begin{center}
\begin{tabular}{m{2.2cm}m{4cm}m{8cm}}
\toprule
\multicolumn{1}{l}{\bf\small Dataset} & \bf\small \makecell[c]{Input}  &\bf\small \makecell[c]{\ourmethod}\\ 
\midrule
\footnotesize \makecell[l]{Clutrr} & \tiny \textbf{Context}: James took his son Francisco out on a hunting trip. Natasha, Francisco's daughter, wanted to come, but they thought she was too young. Instead, James's wife Frances stayed home with her. \newline \textcolor{DodgerBlue2}{\textbf{Question}: How is Natasha related to Frances? } & \tiny  To determine how Natasha is related to Frances, let's analyze the information provided and trace the family relationships step by step. Given Information:1. James took his son Francisco out on a hunting trip. This means James is Francisco's father ... Step-by-Step Reasoning: 1. James is Francisco's father, Francisco is James's son. 2. Since Frances is James's wife, she is also Francisco's mother. 3. Since Francisco is James's son and Frances's son, Natasha is Francisco's daughter, making Natasha James's granddaughter and Frances's granddaughter.4. Conclusion: Natasha is the granddaughter of Frances. \newline \textcolor{darkgreen}{Answer: granddaughter} \textcolor{darkgreen}{\ding{51}} \\
\midrule
\footnotesize \makecell[l]{Folio} & \tiny \textbf{Context}: Diamond Mine is a professional wrestling stable, formed in WWE. Roderick Strong leads Diamond Mine. Diamond Mine includes the Creed Brothers, and Ivy Nile. Imperium has a feud with Diamond Mine. \newline \textcolor{DodgerBlue2}{\textbf{Question}: Roderick strong leads a professional wrestling stable.}  & \tiny To determine whether the statement "Roderick Strong leads a professional wrestling stable" is true, false, or unknown based on the given context, let's analyze the information provided: 1. The context states: ... 3. From the context, we know that Roderick Strong leads Diamond Mine, and Diamond Mine is a professional wrestling stable. 4. Therefore, \textcolor{darkgreen}{since Roderick Strong leads Diamond Mine, and Diamond Mine is a professional wrestling stable, it follows that Roderick Strong leads a professional wrestling stable.} \newline \textcolor{darkgreen}{Answer: true } \textcolor{darkgreen}{\ding{51}} \\
\midrule
\footnotesize \makecell[l]{Logical \\ Deduction} & \tiny \textbf{Context}: A fruit stand sells five fruits: apples, loquats, cantaloupes, watermelons, and kiwis. The watermelons are more expensive than the loquats. The watermelons are less expensive than the kiwis. The loquats are more expensive than the apples. The cantaloupes are the cheapest.  \newline \textbf{Options}: A) The apples are the third-most expensive. B) The loquats are the third-most expensive. C) The cantaloupes are the third-most expensive. D) The watermelons are the third-most expensive. E) The kiwis are the third-most expensive.\newline \textcolor{DodgerBlue2}{\textbf{Question}: Which of the following is true? } & \tiny To determine the correct order of the fruits from most expensive to least expensive, let's analyze the information provided: 1. The cantaloupes are the cheapest.  This means cantaloupes are the least expensive. 2. The loquats are more expensive than the apples. Loquats > Apples 3. The watermelons are more expensive than the loquats. Watermelons > Loquats 4. The watermelons are less expensive than the kiwis.  Kiwis > Watermelons \textcolor{darkgreen}{Now, let's piece this information together to establish the order from most expensive to least expensive: the order from most expensive to least expensive is: Kiwis > Watermelons > Loquats > Apples > Cantaloupes} Now, let's evaluate the options: A) The apples are the third-most expensive. Apples are fourth in the order, so this is false. B) The loquats are the third-most expensive. Loquats are third in the order, so this is true. C) The cantaloupes are the third-most expensive. Cantaloupes are fifth in the order, so this is false. D) The watermelons are the third-most expensive. Watermelons are second in the order, so this is false. E) The kiwis are the third-most expensive. Kiwis are first in the order, so this is false \newline \textcolor{darkgreen}{Answer: B} \textcolor{darkgreen}{\ding{51}} \\
\midrule
\footnotesize \makecell[l]{ProverQA} & \tiny \textbf{Context}: Every protozoa is either parasitic or free-living, but not both. Frederick lives in a host. Any protozoa that lives in a host or feeds on nutrients is parasitic. Queenie lives in a host. Queenie does not feed on nutrients. \newline  \textbf{Options}: A) True B) False C) Uncertain \newline \textcolor{DodgerBlue2}{\textbf{Question}: Queenie is free-living.} & \tiny To determine whether the statement "Queenie is free-living", let's analyze the information provided and the options step by step. Given Information: 1. Every protozoa is either parasitic or free-living, but not both. This means if a protozoa is parasitic, it cannot be free-living, and vice versa ... Step-by-Step Reasoning: 1. Queenie lives in a host. From the given information, any protozoa that lives in a host or feeds on nutrients is parasitic. Since Queenie lives in a host, she falls into the category of protozoa that lives in a host. 2. Any protozoa that lives in a host or feeds on nutrients is parasitic. Since Queenie lives in a host, she is parasitic. 3.  Every protozoa is either parasitic or free-living, but not both.  If Queenie is parasitic, she cannot be free-living. 4. Conclusion: Since Queenie is parasitic, she is not free-living.  \newline \textcolor{darkgreen}{Answer: B} \textcolor{darkgreen}{\ding{51}} \\
\midrule
\footnotesize \makecell[l]{LogiQA} & \tiny \textbf{Context}:  Customs is a state organ that supervises, checks and collects customs duties on all goods and articles entering and leaving the country. \newline \textbf{Options}: A) Check other people's luggage B)  Check the body of others C) Take portraits of others D)Detain others for one day \newline \textcolor{DodgerBlue2}{\textbf{Question}: According to the above definition, when on duty, customs officer can?} & \tiny To determine the correct answer, we need to analyze the role of a customs officer based on the given context and see which option aligns with their duties. Reasoning: 1. The primary role of a customs officer is to supervise, check, and collect customs duties on goods and articles entering and leaving the country.  2. To ensure compliance with customs regulations, customs officers may need to inspect luggage or packages to verify the contents and ensure they are properly declared.  3. Checking the body of others (Option B) is not a standard duty of a customs officer, as it would involve invasive procedures that are not typically part of customs inspection.  4. Taking portraits of others (Option C) is unrelated to the duties of a customs officer.  5. Detaining others for one day (Option D) is a more severe action and would require specific legal grounds, which are not mentioned in the context of customs duties.  Conclusion: The most appropriate action for a customs officer, based on their role, is to check other people's luggage (Option A) to ensure compliance with customs regulations.  \newline \textcolor{darkgreen}{Answer: A} \textcolor{darkgreen}{\ding{51}} \\
\midrule
\footnotesize \makecell[l]{ProntoQA} & \tiny \textbf{Rules}: Every vumpus is a rompus and a sterpus. Every rompus is a brimpus and a lempus. Zumpuses are wumpuses and dumpuses. Each grimpus is a shumpus and a tumpus. Each lorpus is muffled and an impus. Every numpus is a gorpus. Gorpuses are zumpuses and jompuses. Every jompus is a vumpus and a grimpus. \newline \textbf{Facts}: Alex is a brimpus and a lempus. Alex is a zumpus and a gorpus \newline \textcolor{DodgerBlue2}{\textbf{Question}: Alex is not a vumpus?} \newline & \tiny We need to analyze the information provided and trace Alex's properties through the relationships defined.  Given Rules:  1. Every vumpus is a rompus and a sterpus. 2. Every rompus is a brimpus and a lempus ... Given Facts:  Alex is a brimpus and a lempus.  Alex is a zumpus and a gorpus.  Step-by-Step Reasoning: 1.  Alex is a zumpus and a gorpus. From rule 7: Every gorpus is a zumpus and a jompus. Since Alex is a gorpus, Alex is a zumpus and a jompus. From rule 3: Zumpuses are wumpuses and dumpuses.  Since Alex is a zumpus, Alex is a wumpus and a dumpus. From rule 8: Every jompus is a vumpus and a grimpus. Since Alex is a jompus, Alex is a vumpus and a grimpus. 2. Alex is a vumpus.  From the above reasoning, since Alex is a jompus, and every jompus is a vumpus, Alex must be a vumpus. 3.  Conclusion: The statement "Alex is not a vumpus" would be false because we have determined that Alex is indeed a vumpus. \newline \textcolor{darkgreen}{Answer: false} \textcolor{darkgreen}{\ding{51}} \\

\bottomrule
\end{tabular}
\end{center}
\end{table}

%% file: table/negative_case_studies.tex
\begin{table}[ht]
\caption{Negative examples of \ourmethod on various datasets.}
\label{tab:negative-case-study}
 \setlength\tabcolsep{2.5pt}
\begin{center}
\begin{tabular}{m{2.2cm}m{4.5cm}m{7.5cm}}
\toprule
\multicolumn{1}{l}{\bf\small Dataset} & \bf\small \makecell[c]{Input}  &\bf\small \makecell[c]{\ourmethod}\\ 
\midrule
% \footnotesize \makecell[l]{Clutrr} & \tiny \textbf{Context}: \newline \textcolor{DodgerBlue2}{\textbf{Question}:  } & \tiny  \newline \textcolor{darkgreen}{Answer: } \textcolor{darkgreen}{\ding{51}} \\
% \midrule
% \footnotesize \makecell[l]{Folio} & \tiny \textbf{Context}:  \newline \textcolor{DodgerBlue2}{\textbf{Question}: }  & \tiny \textcolor{darkgreen}{} \newline \textcolor{darkgreen}{Answer:  } \textcolor{darkgreen}{\ding{51}} \\
% \midrule
% \footnotesize \makecell[l]{Logical \\ Deduction} & \tiny \textbf{Context}:  \newline \textbf{Options}:\newline \textcolor{DodgerBlue2}{\textbf{Question}:  } & \tiny \textcolor{darkgreen}{ \newline \textcolor{darkgreen}{Answer: } \textcolor{darkgreen}{\ding{51}} \\
% \midrule
% \footnotesize \makecell[l]{ProverQA} & \tiny \textbf{Context}:  \newline  \textbf{Options}: \newline \textcolor{DodgerBlue2}{\textbf{Question}: } & \tiny  \newline \textcolor{darkgreen}{Answer: } \textcolor{darkgreen}{\ding{51}} \\
% \midrule
\footnotesize \makecell[l]{LogiQA} & \tiny \textbf{Context}: As far as the role of individual and nation is concerned, cultural quality mainly infiltrates people's mind and improves people's realm by forming a specific cultural atmosphere; In other words, the role of cultural quality is indirect and abstract  \newline \textbf{Options}: A)Cultural quality is more important than humanistic spirit B)Humanistic spirit is more important than cultural quality C)Cultural quality and humanistic spirit are one thing D)When vigorously improving the quality of national culture, we should attach great importance to the construction of humanistic spirit \newline \textcolor{DodgerBlue2}{\textbf{Question}: The function of humanistic spirit is more direct and specific} \newline & \tiny Context Analysis: The role of cultural quality is described as indirect and abstract. It mainly infiltrates people's minds and improves their realm by forming a specific cultural atmosphere. The function of humanistic spirit is described as more direct and specific.  Question Analysis: The question asks us to compare the importance of cultural quality and humanistic spirit based on their functions. It suggests that since the humanistic spirit is more direct and specific, we should consider its importance in relation to cultural quality.  Reasoning:  The context tells us that cultural quality is indirect and abstract, while the humanistic spirit is direct and specific. Since the humanistic spirit is more direct and specific, it likely has a more immediate and tangible impact compared to cultural quality, which is more indirect and abstract. Therefore, it would be reasonable to conclude that the humanistic spirit is more important than cultural quality in terms of its direct and specific function.  Conclusion: Given that the humanistic spirit is more direct and specific, it is more important than cultural quality. \newline \textcolor{red}{Answer: B} \textcolor{red}{\ding{55}} \\
\midrule
\footnotesize \makecell[l]{AR-LSAT} & \tiny \textbf{Context}: While on vacation, Sukanya receives several e-mail messages from work, each message from one of three associates: Hilary, Jerome, and Lula. Sukanya receives at least one and no more than two messages from each of them. Sukanya receives each message on the day it is sent. No more than one message is sent each day. The messages are received in a manner consistent with the following: The first message is not from Lula. Both the first and last messages are from the same person. Exactly once Sukanya receives a message from Jerome on the day after receiving one from Hilary. Of the first three messages, exactly one is from Jerome. \newline \textbf{Options}: A) Exactly one of the messages is from Lula. B) Exactly two of the messages are from Jerome. C) The second message is from Lula. D) The third message is from Hilary. E) The fourth message is from Jerome. \newline \textcolor{DodgerBlue2} {\textbf{Question}: If Sukanya receives exactly four e-mail messages, then which one of the following must be true?} \newline & \tiny Step-by-Step Reasoning: 1. The first message is not from Lula.  Both the first and last messages are from the same person Therefore, the last message must also be from the same person as the first message. So, the first and fourth messages are from either Hilary or Jerome. 2.  Messages from Jerome: Since there is only one Jerome message in the first three messages, the Jerome message must be in one of the first three messages.  The Jerome message on the day after Hilary must be the fourth message (since there are only four messages). 3.  Messages from Lula:  Sukanya receives at least one and no more than two messages from each associate.  Since the first message is not from Lula, Lula must be in the second, third, or fourth message.  If the first and fourth messages are from Hilary, then the second and third messages must include Lula and Jerome.  If the first and fourth messages are from Jerome, then the second and third messages must include Hilary and Lula. 4.  Analyzing the Options:  Option A:  Since Sukanya receives at least one message from each associate and there are four messages, it's possible that Lula sends two messages.  This option is not necessarily true.   Option B:  The first and fourth messages are from the same person (either Hilary or Jerome). If the first and fourth messages are from Jerome, then Jerome sends two messages.  If the first and fourth messages are from Hilary, then Jerome must send one message in the first three messages and one in the fourth message. Therefore, Jerome could send either one or two messages.  This option is not necessarily true.  Option C:  The second message could be from Lula, but it's not necessarily true based on the constraints.  This option is not necessarily true.   Option D:  The third message could be from Hilary, but it's not necessarily true based on the constraints.  Option E:  Since exactly once Sukanya receives a message from Jerome on the day after receiving one from Hilary, and there are only four messages, the Jerome message on the day after Hilary must be the fourth message. Therefore, the fourth message must be from Jerome. \newline \textcolor{red}{Answer: E} \textcolor{red}{\ding{55}} \\
\midrule
\footnotesize \makecell[l]{LogicNLI} & \tiny \textbf{Rules}: If there is at least one people who is popular, then Curt is not unable. If there is someone who is loyal, then Brock is timid and Lucille is not popular. If there is someone who is timid, then Marcus is unable and Lucille is loyal. Garth being not shiny is equivalent to Jed being not unable and Marcus being light. If someone who is not timid is also loyal, then he is not unable. If there is at least one people who is light or not shiny, then Silas is timid. If Lucille is light, then Brock is not popular. If Curt is not light, then Marcus is popular. Silas is light if and only if Silas is loyal and Brock is not unable. If someone is not timid, then he is loyal. If there is someone who is both unable and not timid, then Jed is shiny. Someone who is both not shiny and unable is always light. \newline \textbf{Facts}:  Silas is light. Marcus is not timid. Marcus is popular. Marcus is light. Lucille is not unable. Curt is loyal. Brock is unable. Jed is loyal. Curt is unable. Garth is light. Brock is not shiny. Jed is popular. \newline \textcolor{DodgerBlue2}{\textbf{Question}:  Brock is unable. } \newline & \tiny  Given Facts: 1. Silas is light. 2. Marcus is not timid. ... Given Rules: 1. If there is at least one person who is popular, then Curt is not unable. 2. If there is someone who is loyal, then Brock is timid and Lucille is not popular. ...  Analysis: From the given facts, we can see that **Fact 7** directly states: Brock is unable.  This means that the statement "Brock is unable" is **entailed** by the facts provided. Conclusion: Since the fact "Brock is unable" is explicitly given in the facts, the statement \"Brock is unable\" is **entailed** by the information. \newline \textcolor{red}{Answer: entailment} \textcolor{red}{\ding{55}} \\
\bottomrule
\end{tabular}
\end{center}
\end{table}

%% file: table/ood_case_studies.tex
\begin{table}[ht]
\vspace{-1em}
\caption{OOD trajectories before and after training by \ourmethod for comparison.}
\label{tab:ood_case_studies}
 \setlength\tabcolsep{2.5pt}
\begin{center}
\begin{tabular}{m{2.2cm}m{2.5cm}m{4.3cm}m{4.3cm}}
\toprule
\multicolumn{1}{l}{\bf\small Dataset} & \bf\small \makecell[c]{Input}  & \bf\small \makecell[c]{Before Training}  & \bf\small \makecell[c]{After Training}\\ \midrule
\footnotesize \makecell[c]{ProverQA} & \tiny \textbf{Context}: Colten is coordinated. If Colten has good intentions, then he is either good-natured or well-meaning (but not both). Colten is either coordinated or clumsy, but not both. Colten is good-natured. Colten has good intentions. \newline \textbf{Options}: A) True B) False C) Uncertain \newline \textcolor{DodgerBlue2}{\textbf{Question}: Colten is lovable?} & \tiny  Step: 1. This tells us that Colten is not clumsy, as he is either coordinated or clumsy, but not both. 2. This is a conditional statement. If the condition (good intentions) is true, then one of the two outcomes (good-natured or well-meaning) must be true, but not both. 3. This is a tautology based on the first statement, so it doesn't provide new information. 4. This directly tells us that Colten is good-natured. 5. This is the condition in the second statement, so we can apply it. We know from the fourth statement that Colten is good-natured. ... \textcolor{BrickRed}{The term lovable is subjective and not directly defined in the given context. However, based on the information provided, we can infer that Colten is a good person with good intentions and a good nature. These traits generally contribute to someone being considered lovable.}  \newline \textcolor{BrickRed}{Answer: A.} \textcolor{darkred}{\ding{55}} 
& \tiny  To determine whether the statement Colten is lovable is true, false, or uncertain based on the given context, let's analyze the information step by step. From statement 1: ... From statement 3:  Colten is either coordinated or clumsy, but not both. 3.From statement 5: Colten has good intentions.4. From statement 2: If Colten has good intentions, then he is either good-natured or well-meaning (but not both). 5. From statement 4:  Colten is good-natured. Therefore, from statement 2, since Colten is good-natured, he cannot be well-meaning (because it's either good-natured or well-meaning, but not both).\textcolor{OliveGreen}{We know that Colten is good-natured and has good intentions. However, the context does not provide any direct information about whether Colten is lovable. The attributes of being coordinated, good-natured, and having good intentions do not necessarily imply that Colten is lovable, as lovable is not explicitly defined or linked to these attributes in the given context.} Therefore, based on the provided information, we cannot definitively determine whether Colten is lovable.\newline \textcolor{OliveGreen}{Answer: C.} \textcolor{darkgreen}{\ding{51}}  \\
\midrule
\footnotesize \makecell[c]{ProverQA} & \tiny \textbf{Context}: Tommy has fine fleece. For all sheep, if a sheep has fine fleece, then it is either prized by farmers or produces quality wool (or both). Nolan has fine fleece. Tommy is not prized by farmers.Question: Based on the above information, is the following statement true, false, or uncertain? Tommy produces quality wool. \newline \textbf{Options}: A) True B) False C) Uncertain \newline \textcolor{DodgerBlue2}{\textbf{Question}: Tommy produces quality wool?} & \tiny  The answer is  C) Uncertain. Step 1: Analyze the given information Tommy has fine fleece.For all sheep, if a sheep has fine fleece, then it is either prized by farmers or produces quality wool (or both). Nolan has fine fleece. Tommy is not prized by farmers.Step 2: Apply the given information to the question We know that Tommy has fine fleece, and according to the given information, if a sheep has fine fleece, it is either prized by farmers or produces quality wool (or both). We also know that Tommy is not prized by farmers.Step 3: Determine the answer  \textcolor{BrickRed}{Since Tommy is not prized by farmers, we cannot determine if he produces quality wool or not based on the given information.} \newline \textcolor{BrickRed}{Answer: Uncertain.} \textcolor{darkred}{\ding{55}} 
& \tiny  To determine whether the statement Tommy produces quality wool is true, false, or uncertain based on the given context, let's analyze the information step by step. Reasoning:1. **From statement 1:**  Tommy has fine fleece.2. **From statement 2 (general rule):**  If a sheep has fine fleece, then it is either prized by farmers or produces quality wool (or both).    \textcolor{OliveGreen}{Since Tommy has fine fleece (from statement 1), we can apply this rule to Tommy}:    Tommy is either prized by farmers or produces quality wool (or both).3. **From statement 4:**  Tommy is not prized by farmers.  This means the prized by farmers part of the rule is false for Tommy.4. **Applying the rule to Tommy:**    \textcolor{OliveGreen}{Since Tommy is not prized by farmers, the only remaining possibility is that Tommy produces quality wool.}  \newline \textcolor{OliveGreen}{Answer: A.} \textcolor{darkgreen}{\ding{51}}  \\

\midrule
\footnotesize \makecell[c]{BigBench Hard} & \tiny \textbf{Context}: The following paragraphs each describe a set of three objects arranged in a fixed order. The statements are logically consistent within each paragraph. On a shelf, there are three books: a yellow book, a brown book, and a red book. The red book is to the left of the brown book. The yellow book is to the right of the brown book. \newline \textbf{Options}: (A) The yellow book is the rightmost (B) The brown book is the rightmost (C) The red book is the rightmost \newline \textcolor{DodgerBlue2}{\textbf{Question}: Which of the following options is true?} & \tiny  \textcolor{BrickRed}{The answer is C.} Since the red book is to the left of the brown book, we can place the red book first and the brown book second. The yellow book is to the right of the brown book, so it must be placed third. Now, let's evaluate the options: \textcolor{BrickRed}{(A) The yellow book is the rightmost: This is true, as the yellow book is placed third in the arrangement.}(B) The brown book is the rightmost: This is false, as the brown book is placed second in the arrangement. (C) The red book is the rightmost: This is false, as the red book is placed first in the arrangement.  \newline \textcolor{BrickRed}{Answer: C.} \textcolor{darkred}{\ding{55}} 
& \tiny  To determine which of the options is true, \textcolor{OliveGreen}{let's analyze the given context step by step. Step 1: Understand the relationships between the books.}  From statement 2: The red book is to the left of the brown book.   This means: Red book < Brown book   From statement 3: The yellow book is to the right of the brown book. This means: Brown book < Yellow book   \textcolor{OliveGreen}{Step 2: Arrange the books in order.} Combining the two relationships:  Red book < Brown book < Yellow book   So, the order from left to right is: Red book, Brown book, Yellow book. \textcolor{OliveGreen}{Step 3: Determine the rightmost book.} In the order Red book, Brown book, Yellow book, the rightmost book is the Yellow book. \textcolor{OliveGreen}{Step 4: Evaluate the options.} (A) The yellow book is the rightmost   From our arrangement, the yellow book is indeed the rightmost.  The red book is the leftmost, not the rightmost.  This option is **false**. \newline \textcolor{OliveGreen}{Answer: A.} \textcolor{darkgreen}{\ding{51}}  \\
\bottomrule
\end{tabular}
\end{center}
\vspace{-2em}
\end{table}

%% file: iclr2026_conference.bbl
\begin{thebibliography}{91}
\providecommand{\natexlab}[1]{#1}
\providecommand{\url}[1]{\texttt{#1}}
\expandafter\ifx\csname urlstyle\endcsname\relax
  \providecommand{\doi}[1]{doi: #1}\else
  \providecommand{\doi}{doi: \begingroup \urlstyle{rm}\Url}\fi

\bibitem[{AIME}(2025)]{aime25}
{AIME}.
\newblock {AIME} problems and solutions, 2025.
\newblock URL \url{https://artofproblemsolving.com/wiki/index.php/AIME_Problems_and_Solutions}.

\bibitem[Albalak et~al.(2024)Albalak, Elazar, Xie, Longpre, Lambert, Wang, Muennighoff, Hou, Pan, Jeong, Raffel, Chang, Hashimoto, and Wang]{albalak2024a}
Alon Albalak, Yanai Elazar, Sang~Michael Xie, Shayne Longpre, Nathan Lambert, Xinyi Wang, Niklas Muennighoff, Bairu Hou, Liangming Pan, Haewon Jeong, Colin Raffel, Shiyu Chang, Tatsunori Hashimoto, and William~Yang Wang.
\newblock A survey on data selection for language models.
\newblock \emph{Transactions on Machine Learning Research}, 2024.
\newblock ISSN 2835-8856.
\newblock URL \url{https://openreview.net/forum?id=XfHWcNTSHp}.
\newblock Survey Certification.

\bibitem[An et~al.(2024)An, Ma, Lin, Zheng, Lou, and Chen]{NEURIPS2024_71c3451f}
Shengnan An, Zexiong Ma, Zeqi Lin, Nanning Zheng, Jian-Guang Lou, and Weizhu Chen.
\newblock Make your llm fully utilize the context.
\newblock In A.~Globerson, L.~Mackey, D.~Belgrave, A.~Fan, U.~Paquet, J.~Tomczak, and C.~Zhang (eds.), \emph{Advances in Neural Information Processing Systems}, volume~37, pp.\  62160--62188. Curran Associates, Inc., 2024.
\newblock URL \url{https://proceedings.neurips.cc/paper_files/paper/2024/file/71c3451f6cd6a4f82bb822db25cea4fd-Paper-Conference.pdf}.

\bibitem[{Anthropic}(2025)]{anthropic-claude-3-7-sonnet}
{Anthropic}.
\newblock Claude 3.7 sonnet.
\newblock \url{https://www.anthropic.com/news/claude-3-7-sonnet}, 2025.
\newblock Accessed: 2025-04-11.

\bibitem[Bengio et~al.(2009)Bengio, Louradour, Collobert, and Weston]{bengio2009curriculum}
Yoshua Bengio, J{\'e}r{\^o}me Louradour, Ronan Collobert, and Jason Weston.
\newblock Curriculum learning.
\newblock In \emph{Proceedings of the 26th annual international conference on machine learning}, pp.\  41--48, 2009.

\bibitem[Brown et~al.(2024)Brown, Juravsky, Ehrlich, Clark, Le, R{\'e}, and Mirhoseini]{brown2024large}
Bradley Brown, Jordan Juravsky, Ryan Ehrlich, Ronald Clark, Quoc~V Le, Christopher R{\'e}, and Azalia Mirhoseini.
\newblock Large language monkeys: Scaling inference compute with repeated sampling.
\newblock \emph{arXiv preprint arXiv:2407.21787}, 2024.

\bibitem[Chen et~al.(2025)Chen, Yang, Liu, Lee, Xu, Shoeybi, Catanzaro, and Ping]{chen2025acereason}
Yang Chen, Zhuolin Yang, Zihan Liu, Chankyu Lee, Peng Xu, Mohammad Shoeybi, Bryan Catanzaro, and Wei Ping.
\newblock Acereason-nemotron: Advancing math and code reasoning through reinforcement learning.
\newblock \emph{arXiv preprint arXiv:2505.16400}, 2025.

\bibitem[Chu et~al.(2025)Chu, Zhai, Yang, Tong, Xie, Levine, and Ma]{chu2025sft}
Tianzhe Chu, Yuexiang Zhai, Jihan Yang, Shengbang Tong, Saining Xie, Sergey Levine, and Yi~Ma.
\newblock {SFT} memorizes, {RL} generalizes: A comparative study of foundation model post-training.
\newblock In \emph{The Second Conference on Parsimony and Learning (Recent Spotlight Track)}, 2025.
\newblock URL \url{https://openreview.net/forum?id=d3E3LWmTar}.

\bibitem[Diallo et~al.(2025)Diallo, Bikakis, Dickens, Hunter, and Miller]{diallo2025rule}
A{\"\i}ssatou Diallo, Antonis Bikakis, Luke Dickens, Anthony Hunter, and Rob Miller.
\newblock Rule-guided feedback: Enhancing reasoning by enforcing rule adherence in large language models.
\newblock \emph{arXiv preprint arXiv:2503.11336}, 2025.

\bibitem[Fatemi et~al.(2025)Fatemi, Rafiee, Tang, and Talamadupula]{fatemi2025concise}
Mehdi Fatemi, Banafsheh Rafiee, Mingjie Tang, and Kartik Talamadupula.
\newblock Concise reasoning via reinforcement learning.
\newblock \emph{arXiv preprint arXiv:2504.05185}, 2025.

\bibitem[Gu et~al.(2024)Gu, Jiang, Shi, Tan, Zhai, Xu, Li, Shen, Ma, Liu, et~al.]{gu2024survey}
Jiawei Gu, Xuhui Jiang, Zhichao Shi, Hexiang Tan, Xuehao Zhai, Chengjin Xu, Wei Li, Yinghan Shen, Shengjie Ma, Honghao Liu, et~al.
\newblock A survey on llm-as-a-judge.
\newblock \emph{arXiv preprint arXiv:2411.15594}, 2024.

\bibitem[Guan et~al.(2025)Guan, Zhang, Liu, Shang, Sun, Zhu, Yang, and Yang]{guan2025rstar}
Xinyu Guan, Li~Lyna Zhang, Yifei Liu, Ning Shang, Youran Sun, Yi~Zhu, Fan Yang, and Mao Yang.
\newblock rstar-math: Small llms can master math reasoning with self-evolved deep thinking.
\newblock \emph{arXiv preprint arXiv:2501.04519}, 2025.

\bibitem[Gui et~al.(2024)Gui, Liu, Cheng, Gu, Liu, Wang, Dong, Tang, and Huang]{gui2024logicgame}
Jiayi Gui, Yiming Liu, Jiale Cheng, Xiaotao Gu, Xiao Liu, Hongning Wang, Yuxiao Dong, Jie Tang, and Minlie Huang.
\newblock Logicgame: Benchmarking rule-based reasoning abilities of large language models.
\newblock \emph{arXiv preprint arXiv:2408.15778}, 2024.

\bibitem[Guo et~al.(2025)Guo, Yang, Zhang, Song, Zhang, Xu, Zhu, Ma, Wang, Bi, et~al.]{guo2025deepseek}
Daya Guo, Dejian Yang, Haowei Zhang, Junxiao Song, Ruoyu Zhang, Runxin Xu, Qihao Zhu, Shirong Ma, Peiyi Wang, Xiao Bi, et~al.
\newblock Deepseek-r1: Incentivizing reasoning capability in llms via reinforcement learning.
\newblock \emph{arXiv preprint arXiv:2501.12948}, 2025.

\bibitem[Han et~al.(2024)Han, Schoelkopf, Zhao, Qi, Riddell, Zhou, Coady, Peng, Qiao, Benson, Sun, Wardle-Solano, Szab{\'o}, Zubova, Burtell, Fan, Liu, Wong, Sailor, Ni, Nan, Kasai, Yu, Zhang, Fabbri, Kryscinski, Yavuz, Liu, Lin, Joty, Zhou, Xiong, Ying, Cohan, and Radev]{han-etal-2024-folio}
Simeng Han, Hailey Schoelkopf, Yilun Zhao, Zhenting Qi, Martin Riddell, Wenfei Zhou, James Coady, David Peng, Yujie Qiao, Luke Benson, Lucy Sun, Alexander Wardle-Solano, Hannah Szab{\'o}, Ekaterina Zubova, Matthew Burtell, Jonathan Fan, Yixin Liu, Brian Wong, Malcolm Sailor, Ansong Ni, Linyong Nan, Jungo Kasai, Tao Yu, Rui Zhang, Alexander Fabbri, Wojciech~Maciej Kryscinski, Semih Yavuz, Ye~Liu, Xi~Victoria Lin, Shafiq Joty, Yingbo Zhou, Caiming Xiong, Rex Ying, Arman Cohan, and Dragomir Radev.
\newblock {FOLIO}: Natural language reasoning with first-order logic.
\newblock In Yaser Al-Onaizan, Mohit Bansal, and Yun-Nung Chen (eds.), \emph{Proceedings of the 2024 Conference on Empirical Methods in Natural Language Processing}, pp.\  22017--22031, Miami, Florida, USA, November 2024. Association for Computational Linguistics.
\newblock \doi{10.18653/v1/2024.emnlp-main.1229}.
\newblock URL \url{https://aclanthology.org/2024.emnlp-main.1229/}.

\bibitem[He et~al.(2024)He, Zhang, Yan, Wu, and Chen]{he2024idea}
Kaiyu He, Mian Zhang, Shuo Yan, Peilin Wu, and Zhiyu~Zoey Chen.
\newblock Idea: Enhancing the rule learning ability of large language model agent through induction, deduction, and abduction.
\newblock \emph{arXiv preprint arXiv:2408.10455}, 2024.

\bibitem[He et~al.(2025)He, Yu, and Wang]{he2025large}
Qiyuan He, Jianfei Yu, and Wenya Wang.
\newblock Large language model-enhanced symbolic reasoning for knowledge base completion.
\newblock \emph{arXiv preprint arXiv:2501.01246}, 2025.

\bibitem[Holt(2004)]{holt2004forecasting}
Charles~C Holt.
\newblock Forecasting seasonals and trends by exponentially weighted moving averages.
\newblock \emph{International journal of forecasting}, 20\penalty0 (1):\penalty0 5--10, 2004.

\bibitem[Hu et~al.(2025)Hu, Zhang, Han, Jiang, Zhang, and Shum]{hu2503open}
Jingcheng Hu, Yinmin Zhang, Qi~Han, Daxin Jiang, Xiangyu Zhang, and Heung-Yeung Shum.
\newblock Open-reasoner-zero: An open source approach to scaling up reinforcement learning on the base model, 2025.
\newblock \emph{URL https://arxiv. org/abs/2503.24290}, 2025.

\bibitem[Jaech et~al.(2024)Jaech, Kalai, Lerer, Richardson, El-Kishky, Low, Helyar, Madry, Beutel, Carney, et~al.]{jaech2024openai}
Aaron Jaech, Adam Kalai, Adam Lerer, Adam Richardson, Ahmed El-Kishky, Aiden Low, Alec Helyar, Aleksander Madry, Alex Beutel, Alex Carney, et~al.
\newblock Openai o1 system card.
\newblock \emph{arXiv preprint arXiv:2412.16720}, 2024.

\bibitem[Jia et~al.(2024)Jia, Li, Zhang, Liu, and Zheng]{jia-etal-2024-combining}
Zixia Jia, Junpeng Li, Shichuan Zhang, Anji Liu, and Zilong Zheng.
\newblock Combining supervised learning and reinforcement learning for multi-label classification tasks with partial labels.
\newblock In Lun-Wei Ku, Andre Martins, and Vivek Srikumar (eds.), \emph{Proceedings of the 62nd Annual Meeting of the Association for Computational Linguistics (Volume 1: Long Papers)}, pp.\  13553--13569, Bangkok, Thailand, August 2024. Association for Computational Linguistics.
\newblock \doi{10.18653/v1/2024.acl-long.731}.
\newblock URL \url{https://aclanthology.org/2024.acl-long.731/}.

\bibitem[Kazemi et~al.(2025)Kazemi, Fatemi, Bansal, Palowitch, Anastasiou, Mehta, Jain, Aglietti, Jindal, Chen, et~al.]{kazemi2025big}
Mehran Kazemi, Bahare Fatemi, Hritik Bansal, John Palowitch, Chrysovalantis Anastasiou, Sanket~Vaibhav Mehta, Lalit~K Jain, Virginia Aglietti, Disha Jindal, Peter Chen, et~al.
\newblock Big-bench extra hard.
\newblock \emph{arXiv preprint arXiv:2502.19187}, 2025.

\bibitem[Kwon et~al.(2023)Kwon, Li, Zhuang, Sheng, Zheng, Yu, Gonzalez, Zhang, and Stoica]{kwon2023efficient}
Woosuk Kwon, Zhuohan Li, Siyuan Zhuang, Ying Sheng, Lianmin Zheng, Cody~Hao Yu, Joseph~E. Gonzalez, Hao Zhang, and Ion Stoica.
\newblock Efficient memory management for large language model serving with pagedattention.
\newblock In \emph{Proceedings of the ACM SIGOPS 29th Symposium on Operating Systems Principles}, 2023.

\bibitem[Li et~al.(2025{\natexlab{a}})Li, Li, Wu, Zhu, Wang, Yu, Jiang, Zhu, Jia, Wu, et~al.]{li2025seek}
Hengli Li, Chenxi Li, Tong Wu, Xuekai Zhu, Yuxuan Wang, Zhaoxin Yu, Eric~Hanchen Jiang, Song-Chun Zhu, Zixia Jia, Ying~Nian Wu, et~al.
\newblock Seek in the dark: Reasoning via test-time instance-level policy gradient in latent space.
\newblock \emph{arXiv preprint arXiv:2505.13308}, 2025{\natexlab{a}}.

\bibitem[Li et~al.(2025{\natexlab{b}})Li, Dong, Liu, Yang, Wang, Wang, Zhu, Jia, and Zheng]{li2025reflectevo}
Jiaqi Li, Xinyi Dong, Yang Liu, Zhizhuo Yang, Quansen Wang, Xiaobo Wang, SongChun Zhu, Zixia Jia, and Zilong Zheng.
\newblock Reflectevo: Improving meta introspection of small llms by learning self-reflection.
\newblock \emph{arXiv preprint arXiv:2505.16475}, 2025{\natexlab{b}}.

\bibitem[Li et~al.(2025{\natexlab{c}})Li, Zou, and Liu]{li2025limr}
Xuefeng Li, Haoyang Zou, and Pengfei Liu.
\newblock Limr: Less is more for rl scaling.
\newblock \emph{arXiv preprint arXiv:2502.11886}, 2025{\natexlab{c}}.

\bibitem[Li et~al.(2025{\natexlab{d}})Li, Yue, Xu, Jiang, Niu, Lin, Ramasubramanian, and Poovendran]{li2025small}
Yuetai Li, Xiang Yue, Zhangchen Xu, Fengqing Jiang, Luyao Niu, Bill~Yuchen Lin, Bhaskar Ramasubramanian, and Radha Poovendran.
\newblock Small models struggle to learn from strong reasoners.
\newblock \emph{arXiv preprint arXiv:2502.12143}, 2025{\natexlab{d}}.

\bibitem[Liu et~al.(2023{\natexlab{a}})Liu, Liu, Cui, Teng, Duan, Zhou, and Zhang]{logiqa2023}
Hanmeng Liu, Jian Liu, Leyang Cui, Zhiyang Teng, Nan Duan, Ming Zhou, and Yue Zhang.
\newblock Logiqa 2.0—an improved dataset for logical reasoning in natural language understanding.
\newblock \emph{IEEE/ACM Transactions on Audio, Speech, and Language Processing}, 31:\penalty0 2947--2962, 2023{\natexlab{a}}.
\newblock \doi{10.1109/TASLP.2023.3293046}.

\bibitem[Liu et~al.(2023{\natexlab{b}})Liu, Zhou, Hua, Chong, Tian, Liu, Wang, You, Guo, Zhu, et~al.]{liu2023benchmarking}
Junling Liu, Peilin Zhou, Yining Hua, Dading Chong, Zhongyu Tian, Andrew Liu, Helin Wang, Chenyu You, Zhenhua Guo, Lei Zhu, et~al.
\newblock Benchmarking large language models on cmexam-a comprehensive chinese medical exam dataset.
\newblock \emph{Advances in Neural Information Processing Systems}, 36:\penalty0 52430--52452, 2023{\natexlab{b}}.

\bibitem[Liu et~al.(2024)Liu, Lin, Hewitt, Paranjape, Bevilacqua, Petroni, and Liang]{liu-etal-2024-lost}
Nelson~F. Liu, Kevin Lin, John Hewitt, Ashwin Paranjape, Michele Bevilacqua, Fabio Petroni, and Percy Liang.
\newblock Lost in the middle: How language models use long contexts.
\newblock \emph{Transactions of the Association for Computational Linguistics}, 12:\penalty0 157--173, 2024.
\newblock \doi{10.1162/tacl_a_00638}.
\newblock URL \url{https://aclanthology.org/2024.tacl-1.9/}.

\bibitem[Liu et~al.(2025{\natexlab{a}})Liu, Zheng, Muennighoff, Zeng, Dou, Pang, Jiang, and Lin]{liu2025regmix}
Qian Liu, Xiaosen Zheng, Niklas Muennighoff, Guangtao Zeng, Longxu Dou, Tianyu Pang, Jing Jiang, and Min Lin.
\newblock Regmix: Data mixture as regression for language model pre-training.
\newblock In \emph{The Thirteenth International Conference on Learning Representations}, 2025{\natexlab{a}}.
\newblock URL \url{https://openreview.net/forum?id=5BjQOUXq7i}.

\bibitem[Liu et~al.(2025{\natexlab{b}})Liu, Chen, Li, Qi, Pang, Du, Lee, and Lin]{liu2025understanding}
Zichen Liu, Changyu Chen, Wenjun Li, Penghui Qi, Tianyu Pang, Chao Du, Wee~Sun Lee, and Min Lin.
\newblock Understanding r1-zero-like training: A critical perspective.
\newblock \emph{arXiv preprint arXiv:2503.20783}, 2025{\natexlab{b}}.

\bibitem[Luo et~al.(2024)Luo, Ju, Xiong, Li, Haffari, and Pan]{luo2309chatrule}
L~Luo, J~Ju, B~Xiong, YF~Li, G~Haffari, and S~Pan.
\newblock Chatrule: Mining logical rules with large language models for knowledge graph reasoning. arxiv 2024.
\newblock \emph{arXiv preprint arXiv:2309.01538}, 2024.

\bibitem[Luo et~al.(2025)Luo, Zhang, Liu, Li, Gong, Chen, and Cheng]{luo-etal-2025-velocitune}
Zheheng Luo, Xin Zhang, Xiao Liu, Haoling Li, Yeyun Gong, Qi~Chen, and Peng Cheng.
\newblock Velocitune: A velocity-based dynamic domain reweighting method for continual pre-training.
\newblock In Wanxiang Che, Joyce Nabende, Ekaterina Shutova, and Mohammad~Taher Pilehvar (eds.), \emph{Proceedings of the 63rd Annual Meeting of the Association for Computational Linguistics (Volume 1: Long Papers)}, pp.\  16644--16656, Vienna, Austria, July 2025. Association for Computational Linguistics.
\newblock ISBN 979-8-89176-251-0.
\newblock \doi{10.18653/v1/2025.acl-long.813}.
\newblock URL \url{https://aclanthology.org/2025.acl-long.813/}.

\bibitem[Meng et~al.(2025)Meng, Du, Liu, Zhou, Lu, Fu, Han, Shi, Wang, He, et~al.]{meng2025mm}
Fanqing Meng, Lingxiao Du, Zongkai Liu, Zhixiang Zhou, Quanfeng Lu, Daocheng Fu, Tiancheng Han, Botian Shi, Wenhai Wang, Junjun He, et~al.
\newblock Mm-eureka: Exploring the frontiers of multimodal reasoning with rule-based reinforcement learning.
\newblock \emph{arXiv preprint arXiv:2503.07365}, 2025.

\bibitem[Morishita et~al.(2024)Morishita, Morio, Yamaguchi, and Sogawa]{morishita2024enhancing}
Terufumi Morishita, Gaku Morio, Atsuki Yamaguchi, and Yasuhiro Sogawa.
\newblock Enhancing reasoning capabilities of llms via principled synthetic logic corpus.
\newblock \emph{Advances in Neural Information Processing Systems}, 37:\penalty0 73572--73604, 2024.

\bibitem[Parashar et~al.(2025)Parashar, Gui, Li, Ling, Vemuri, Olson, Li, Zhang, Caverlee, Kalathil, et~al.]{parashar2025curriculum}
Shubham Parashar, Shurui Gui, Xiner Li, Hongyi Ling, Sushil Vemuri, Blake Olson, Eric Li, Yu~Zhang, James Caverlee, Dileep Kalathil, et~al.
\newblock Curriculum reinforcement learning from easy to hard tasks improves llm reasoning.
\newblock \emph{arXiv preprint arXiv:2506.06632}, 2025.

\bibitem[Paszke et~al.(2019)Paszke, Gross, Massa, Lerer, Bradbury, Chanan, Killeen, Lin, Gimelshein, Antiga, Desmaison, Kopf, Yang, DeVito, Raison, Tejani, Chilamkurthy, Steiner, Fang, Bai, and Chintala]{pytorch}
Adam Paszke, Sam Gross, Francisco Massa, Adam Lerer, James Bradbury, Gregory Chanan, Trevor Killeen, Zeming Lin, Natalia Gimelshein, Luca Antiga, Alban Desmaison, Andreas Kopf, Edward Yang, Zachary DeVito, Martin Raison, Alykhan Tejani, Sasank Chilamkurthy, Benoit Steiner, Lu~Fang, Junjie Bai, and Soumith Chintala.
\newblock Pytorch: An imperative style, high-performance deep learning library.
\newblock In H.~Wallach, H.~Larochelle, A.~Beygelzimer, F.~d\textquotesingle Alch\'{e}-Buc, E.~Fox, and R.~Garnett (eds.), \emph{Advances in Neural Information Processing Systems}, volume~32. Curran Associates, Inc., 2019.
\newblock URL \url{https://proceedings.neurips.cc/paper_files/paper/2019/file/bdbca288fee7f92f2bfa9f7012727740-Paper.pdf}.

\bibitem[Peng et~al.(2024)Peng, Xia, Yang, Xiong, Wu, and Xing]{peng2024regenesis}
Xiangyu Peng, Congying Xia, Xinyi Yang, Caiming Xiong, Chien-Sheng Wu, and Chen Xing.
\newblock Regenesis: Llms can grow into reasoning generalists via self-improvement.
\newblock \emph{arXiv preprint arXiv:2410.02108}, 2024.

\bibitem[Pomerleau(1988)]{NIPS1988_812b4ba2}
Dean~A. Pomerleau.
\newblock Alvinn: An autonomous land vehicle in a neural network.
\newblock In D.~Touretzky (ed.), \emph{Advances in Neural Information Processing Systems}, volume~1. Morgan-Kaufmann, 1988.
\newblock URL \url{https://proceedings.neurips.cc/paper_files/paper/1988/file/812b4ba287f5ee0bc9d43bbf5bbe87fb-Paper.pdf}.

\bibitem[Qi et~al.(2025)Qi, Ma, Li, Du, Hui, Wu, Laili, and He]{qi2025large}
Chengwen Qi, Ren Ma, Bowen Li, He~Du, Binyuan Hui, Jinwang Wu, Yuanjun Laili, and Conghui He.
\newblock Large language models meet symbolic provers for logical reasoning evaluation.
\newblock In \emph{The Thirteenth International Conference on Learning Representations}, 2025.
\newblock URL \url{https://openreview.net/forum?id=C25SgeXWjE}.

\bibitem[Rein et~al.(2024)Rein, Hou, Stickland, Petty, Pang, Dirani, Michael, and Bowman]{rein2024gpqa}
David Rein, Betty~Li Hou, Asa~Cooper Stickland, Jackson Petty, Richard~Yuanzhe Pang, Julien Dirani, Julian Michael, and Samuel~R. Bowman.
\newblock {GPQA}: A graduate-level google-proof q\&a benchmark.
\newblock In \emph{First Conference on Language Modeling}, 2024.
\newblock URL \url{https://openreview.net/forum?id=Ti67584b98}.

\bibitem[Saparov \& He(2023)Saparov and He]{saparov2023language}
Abulhair Saparov and He~He.
\newblock Language models are greedy reasoners: A systematic formal analysis of chain-of-thought.
\newblock In \emph{The Eleventh International Conference on Learning Representations}, 2023.
\newblock URL \url{https://openreview.net/forum?id=qFVVBzXxR2V}.

\bibitem[Saparov et~al.(2023)Saparov, Pang, Padmakumar, Joshi, Kazemi, Kim, and He]{saparov2023testing}
Abulhair Saparov, Richard~Yuanzhe Pang, Vishakh Padmakumar, Nitish Joshi, Mehran Kazemi, Najoung Kim, and He~He.
\newblock Testing the general deductive reasoning capacity of large language models using ood examples.
\newblock \emph{Advances in Neural Information Processing Systems}, 36:\penalty0 3083--3105, 2023.

\bibitem[Schulman et~al.(2017)Schulman, Wolski, Dhariwal, Radford, and Klimov]{schulman2017proximal}
John Schulman, Filip Wolski, Prafulla Dhariwal, Alec Radford, and Oleg Klimov.
\newblock Proximal policy optimization algorithms.
\newblock \emph{arXiv preprint arXiv:1707.06347}, 2017.

\bibitem[Servantez et~al.(2024{\natexlab{a}})Servantez, Barrow, Hammond, and Jain]{servantez-etal-2024-chain}
Sergio Servantez, Joe Barrow, Kristian Hammond, and Rajiv Jain.
\newblock Chain of logic: Rule-based reasoning with large language models.
\newblock In Lun-Wei Ku, Andre Martins, and Vivek Srikumar (eds.), \emph{Findings of the Association for Computational Linguistics: ACL 2024}, pp.\  2721--2733, Bangkok, Thailand, August 2024{\natexlab{a}}. Association for Computational Linguistics.
\newblock \doi{10.18653/v1/2024.findings-acl.159}.
\newblock URL \url{https://aclanthology.org/2024.findings-acl.159/}.

\bibitem[Servantez et~al.(2024{\natexlab{b}})Servantez, Barrow, Hammond, and Jain]{servantez2024chain}
Sergio Servantez, Joe Barrow, Kristian Hammond, and Rajiv Jain.
\newblock Chain of logic: Rule-based reasoning with large language models.
\newblock In Lun-Wei Ku, Andre Martins, and Vivek Srikumar (eds.), \emph{Findings of the Association for Computational Linguistics: ACL 2024}, pp.\  2721--2733, Bangkok, Thailand, August 2024{\natexlab{b}}. Association for Computational Linguistics.
\newblock \doi{10.18653/v1/2024.findings-acl.159}.
\newblock URL \url{https://aclanthology.org/2024.findings-acl.159/}.

\bibitem[Shao et~al.(2024)Shao, Wang, Zhu, Xu, Song, Bi, Zhang, Zhang, Li, Wu, et~al.]{shao2024deepseekmath}
Zhihong Shao, Peiyi Wang, Qihao Zhu, Runxin Xu, Junxiao Song, Xiao Bi, Haowei Zhang, Mingchuan Zhang, YK~Li, Y~Wu, et~al.
\newblock Deepseekmath: Pushing the limits of mathematical reasoning in open language models.
\newblock \emph{arXiv preprint arXiv:2402.03300}, 2024.

\bibitem[Sheng et~al.(2024)Sheng, Zhang, Ye, Wu, Zhang, Zhang, Peng, Lin, and Wu]{sheng2024hybridflow}
Guangming Sheng, Chi Zhang, Zilingfeng Ye, Xibin Wu, Wang Zhang, Ru~Zhang, Yanghua Peng, Haibin Lin, and Chuan Wu.
\newblock Hybridflow: A flexible and efficient rlhf framework.
\newblock \emph{arXiv preprint arXiv: 2409.19256}, 2024.

\bibitem[Shi et~al.(2025{\natexlab{a}})Shi, Wu, Song, Zhou, and Zhao]{shi2025efficient}
Taiwei Shi, Yiyang Wu, Linxin Song, Tianyi Zhou, and Jieyu Zhao.
\newblock Efficient reinforcement finetuning via adaptive curriculum learning.
\newblock \emph{arXiv preprint arXiv:2504.05520}, 2025{\natexlab{a}}.

\bibitem[Shi et~al.(2025{\natexlab{b}})Shi, Zhang, Wu, Fang, Zhang, Zhao, Chen, Zhang, Cui, Zhu, Han, Xu, and Zhou]{shi-etal-2025-dids}
Weijie Shi, Jipeng Zhang, Yaguang Wu, Jingzhi Fang, Shibo Zhang, Yao Zhao, Hao Chen, Ruiyuan Zhang, Yue Cui, Jia Zhu, Sirui Han, Jiajie Xu, and Xiaofang Zhou.
\newblock {DIDS}: Domain impact-aware data sampling for large language model training.
\newblock In Christos Christodoulopoulos, Tanmoy Chakraborty, Carolyn Rose, and Violet Peng (eds.), \emph{Proceedings of the 2025 Conference on Empirical Methods in Natural Language Processing}, pp.\  4330--4350, Suzhou, China, November 2025{\natexlab{b}}. Association for Computational Linguistics.
\newblock ISBN 979-8-89176-332-6.
\newblock \doi{10.18653/v1/2025.emnlp-main.215}.
\newblock URL \url{https://aclanthology.org/2025.emnlp-main.215/}.

\bibitem[Silver \& Sutton(2025)Silver and Sutton]{silver2025welcome}
David Silver and Richard~S Sutton.
\newblock Welcome to the era of experience.
\newblock \emph{Preprint of a chapter to appear in Designing an Intelligence, edited by George Konidaris, MIT Press (forthcoming)}, 2025.

\bibitem[Sinha et~al.(2019)Sinha, Sodhani, Dong, Pineau, and Hamilton]{sinha-etal-2019-clutrr}
Koustuv Sinha, Shagun Sodhani, Jin Dong, Joelle Pineau, and William~L. Hamilton.
\newblock {CLUTRR}: A diagnostic benchmark for inductive reasoning from text.
\newblock In Kentaro Inui, Jing Jiang, Vincent Ng, and Xiaojun Wan (eds.), \emph{Proceedings of the 2019 Conference on Empirical Methods in Natural Language Processing and the 9th International Joint Conference on Natural Language Processing (EMNLP-IJCNLP)}, pp.\  4506--4515, Hong Kong, China, November 2019. Association for Computational Linguistics.
\newblock \doi{10.18653/v1/D19-1458}.
\newblock URL \url{https://aclanthology.org/D19-1458/}.

\bibitem[Srivastava et~al.(2025)Srivastava, Cao, and Wang]{srivastava2025towards}
Gaurav Srivastava, Shuxiang Cao, and Xuan Wang.
\newblock Towards reasoning ability of small language models.
\newblock \emph{arXiv preprint arXiv:2502.11569}, 2025.

\bibitem[Sun et~al.(2024)Sun, Yu, Shen, Liu, Yang, Welleck, and Gan]{sun2024easytohard}
Zhiqing Sun, Longhui Yu, Yikang Shen, Weiyang Liu, Yiming Yang, Sean Welleck, and Chuang Gan.
\newblock Easy-to-hard generalization: Scalable alignment beyond human supervision.
\newblock In \emph{The Thirty-eighth Annual Conference on Neural Information Processing Systems}, 2024.
\newblock URL \url{https://openreview.net/forum?id=qwgfh2fTtN}.

\bibitem[Suzgun et~al.(2023)Suzgun, Scales, Sch{\"a}rli, Gehrmann, Tay, Chung, Chowdhery, Le, Chi, Zhou, and Wei]{suzgun-etal-2023-challenging}
Mirac Suzgun, Nathan Scales, Nathanael Sch{\"a}rli, Sebastian Gehrmann, Yi~Tay, Hyung~Won Chung, Aakanksha Chowdhery, Quoc Le, Ed~Chi, Denny Zhou, and Jason Wei.
\newblock Challenging {BIG}-bench tasks and whether chain-of-thought can solve them.
\newblock In Anna Rogers, Jordan Boyd-Graber, and Naoaki Okazaki (eds.), \emph{Findings of the Association for Computational Linguistics: ACL 2023}, pp.\  13003--13051, Toronto, Canada, July 2023. Association for Computational Linguistics.
\newblock \doi{10.18653/v1/2023.findings-acl.824}.
\newblock URL \url{https://aclanthology.org/2023.findings-acl.824/}.

\bibitem[Tafjord et~al.(2021)Tafjord, Dalvi, and Clark]{tafjord-etal-2021-proofwriter}
Oyvind Tafjord, Bhavana Dalvi, and Peter Clark.
\newblock {P}roof{W}riter: Generating implications, proofs, and abductive statements over natural language.
\newblock In Chengqing Zong, Fei Xia, Wenjie Li, and Roberto Navigli (eds.), \emph{Findings of the Association for Computational Linguistics: ACL-IJCNLP 2021}, pp.\  3621--3634, Online, August 2021. Association for Computational Linguistics.
\newblock \doi{10.18653/v1/2021.findings-acl.317}.
\newblock URL \url{https://aclanthology.org/2021.findings-acl.317/}.

\bibitem[Tang et~al.(2023{\natexlab{a}})Tang, Zheng, Li, Meng, Zhu, Liang, and Zhang]{tang2023icsr}
Xiaojuan Tang, Zilong Zheng, Jiaqi Li, Fanxu Meng, Song-Chun Zhu, Yitao Liang, and Muhan Zhang.
\newblock Large language models are in-context semantic reasoners rather than symbolic reasoners.
\newblock \emph{arXiv preprint arXiv: 2305.14825}, 2023{\natexlab{a}}.

\bibitem[Tang et~al.(2023{\natexlab{b}})Tang, Zheng, Li, Meng, Zhu, Liang, and Zhang]{tang2023large}
Xiaojuan Tang, Zilong Zheng, Jiaqi Li, Fanxu Meng, Song-Chun Zhu, Yitao Liang, and Muhan Zhang.
\newblock Large language models are in-context semantic reasoners rather than symbolic reasoners.
\newblock \emph{arXiv preprint arXiv:2305.14825}, 2023{\natexlab{b}}.

\bibitem[Tang et~al.(2024)Tang, Li, Liang, Zhang, and Zheng]{tang2024mars}
Xiaojuan Tang, Jiaqi Li, Yitao Liang, Muhan Zhang, and Zilong Zheng.
\newblock Mars: Situated inductive reasoning in an open-world environment.
\newblock In \emph{38th Conference on Neural Information Processing Systems (NeurIPS 2024) Track on Datasets and Benchmarks}, 2024.

\bibitem[Tian et~al.(2021)Tian, Li, Chen, Xiao, He, and Jin]{tian-etal-2021-diagnosing}
Jidong Tian, Yitian Li, Wenqing Chen, Liqiang Xiao, Hao He, and Yaohui Jin.
\newblock Diagnosing the first-order logical reasoning ability through {L}ogic{NLI}.
\newblock In Marie-Francine Moens, Xuanjing Huang, Lucia Specia, and Scott Wen-tau Yih (eds.), \emph{Proceedings of the 2021 Conference on Empirical Methods in Natural Language Processing}, pp.\  3738--3747, Online and Punta Cana, Dominican Republic, November 2021. Association for Computational Linguistics.
\newblock \doi{10.18653/v1/2021.emnlp-main.303}.
\newblock URL \url{https://aclanthology.org/2021.emnlp-main.303/}.

\bibitem[Veeraboina(2023)]{aime_1983_2024}
Hemish Veeraboina.
\newblock Aime problem set 1983-2024.
\newblock Kaggle Dataset, 2023.
\newblock URL \url{https://www.kaggle.com/datasets/hemishveeraboina/aime-problem-set-1983-2024}.
\newblock Accessed 24 November 2025.

\bibitem[Wang et~al.(2024{\natexlab{a}})Wang, Wei, Choi, and Ren]{wang-etal-2024-llms}
Siyuan Wang, Zhongyu Wei, Yejin Choi, and Xiang Ren.
\newblock Can {LLM}s reason with rules? logic scaffolding for stress-testing and improving {LLM}s.
\newblock In Lun-Wei Ku, Andre Martins, and Vivek Srikumar (eds.), \emph{Proceedings of the 62nd Annual Meeting of the Association for Computational Linguistics (Volume 1: Long Papers)}, pp.\  7523--7543, Bangkok, Thailand, August 2024{\natexlab{a}}. Association for Computational Linguistics.
\newblock \doi{10.18653/v1/2024.acl-long.406}.
\newblock URL \url{https://aclanthology.org/2024.acl-long.406/}.

\bibitem[Wang et~al.(2024{\natexlab{b}})Wang, Wei, Choi, and Ren]{wang-etal-2024-symbolic}
Siyuan Wang, Zhongyu Wei, Yejin Choi, and Xiang Ren.
\newblock Symbolic working memory enhances language models for complex rule application.
\newblock In Yaser Al-Onaizan, Mohit Bansal, and Yun-Nung Chen (eds.), \emph{Proceedings of the 2024 Conference on Empirical Methods in Natural Language Processing}, pp.\  17583--17604, Miami, Florida, USA, November 2024{\natexlab{b}}. Association for Computational Linguistics.
\newblock \doi{10.18653/v1/2024.emnlp-main.974}.
\newblock URL \url{https://aclanthology.org/2024.emnlp-main.974/}.

\bibitem[Wang et~al.(2024{\natexlab{c}})Wang, Wei, Choi, and Ren]{wang2024can}
Siyuan Wang, Zhongyu Wei, Yejin Choi, and Xiang Ren.
\newblock Can {LLM}s reason with rules? logic scaffolding for stress-testing and improving {LLM}s.
\newblock In Lun-Wei Ku, Andre Martins, and Vivek Srikumar (eds.), \emph{Proceedings of the 62nd Annual Meeting of the Association for Computational Linguistics (Volume 1: Long Papers)}, pp.\  7523--7543, Bangkok, Thailand, August 2024{\natexlab{c}}. Association for Computational Linguistics.
\newblock \doi{10.18653/v1/2024.acl-long.406}.
\newblock URL \url{https://aclanthology.org/2024.acl-long.406/}.

\bibitem[Wang et~al.(2024{\natexlab{d}})Wang, Wei, Choi, and Ren]{wang2024symbolic}
Siyuan Wang, Zhongyu Wei, Yejin Choi, and Xiang Ren.
\newblock Symbolic working memory enhances language models for complex rule application.
\newblock \emph{arXiv preprint arXiv:2408.13654}, 2024{\natexlab{d}}.

\bibitem[Wang et~al.(2023)Wang, Wei, Schuurmans, Le, Chi, Narang, Chowdhery, and Zhou]{wang2023selfconsistency}
Xuezhi Wang, Jason Wei, Dale Schuurmans, Quoc~V Le, Ed~H. Chi, Sharan Narang, Aakanksha Chowdhery, and Denny Zhou.
\newblock Self-consistency improves chain of thought reasoning in language models.
\newblock In \emph{The Eleventh International Conference on Learning Representations}, 2023.
\newblock URL \url{https://openreview.net/forum?id=1PL1NIMMrw}.

\bibitem[Wei et~al.(2022{\natexlab{a}})Wei, Bosma, Zhao, Guu, Yu, Lester, Du, Dai, and Le]{wei2022finetuned}
Jason Wei, Maarten Bosma, Vincent Zhao, Kelvin Guu, Adams~Wei Yu, Brian Lester, Nan Du, Andrew~M. Dai, and Quoc~V Le.
\newblock Finetuned language models are zero-shot learners.
\newblock In \emph{International Conference on Learning Representations}, 2022{\natexlab{a}}.
\newblock URL \url{https://openreview.net/forum?id=gEZrGCozdqR}.

\bibitem[Wei et~al.(2022{\natexlab{b}})Wei, Wang, Schuurmans, Bosma, brian ichter, Xia, Chi, Le, and Zhou]{wei2022chain}
Jason Wei, Xuezhi Wang, Dale Schuurmans, Maarten Bosma, brian ichter, Fei Xia, Ed~H. Chi, Quoc~V Le, and Denny Zhou.
\newblock Chain of thought prompting elicits reasoning in large language models.
\newblock In Alice~H. Oh, Alekh Agarwal, Danielle Belgrave, and Kyunghyun Cho (eds.), \emph{Advances in Neural Information Processing Systems}, 2022{\natexlab{b}}.
\newblock URL \url{https://openreview.net/forum?id=_VjQlMeSB_J}.

\bibitem[Wei et~al.(2025)Wei, Duchenne, Copet, Carbonneaux, Zhang, Fried, Synnaeve, Singh, and Wang]{wei2025swe}
Yuxiang Wei, Olivier Duchenne, Jade Copet, Quentin Carbonneaux, Lingming Zhang, Daniel Fried, Gabriel Synnaeve, Rishabh Singh, and Sida~I Wang.
\newblock Swe-rl: Advancing llm reasoning via reinforcement learning on open software evolution.
\newblock \emph{arXiv preprint arXiv:2502.18449}, 2025.

\bibitem[Xie et~al.(2023)Xie, Pham, Dong, Du, Liu, Lu, Liang, Le, Ma, and Yu]{xie2023doremi}
Sang~Michael Xie, Hieu Pham, Xuanyi Dong, Nan Du, Hanxiao Liu, Yifeng Lu, Percy Liang, Quoc~V Le, Tengyu Ma, and Adams~Wei Yu.
\newblock Doremi: Optimizing data mixtures speeds up language model pretraining.
\newblock In \emph{Thirty-seventh Conference on Neural Information Processing Systems}, 2023.
\newblock URL \url{https://openreview.net/forum?id=lXuByUeHhd}.

\bibitem[Xie et~al.(2025)Xie, Gao, Ren, Luo, Hong, Dai, Zhou, Qiu, Wu, and Luo]{xie2025logic}
Tian Xie, Zitian Gao, Qingnan Ren, Haoming Luo, Yuqian Hong, Bryan Dai, Joey Zhou, Kai Qiu, Zhirong Wu, and Chong Luo.
\newblock Logic-rl: Unleashing llm reasoning with rule-based reinforcement learning.
\newblock \emph{arXiv preprint arXiv:2502.14768}, 2025.

\bibitem[Xiong et~al.(2024)Xiong, Jin, Lu, and Zhang]{xiong2024benchmarking}
Guangzhi Xiong, Qiao Jin, Zhiyong Lu, and Aidong Zhang.
\newblock Benchmarking retrieval-augmented generation for medicine.
\newblock In \emph{Findings of the Association for Computational Linguistics ACL 2024}, pp.\  6233--6251, 2024.

\bibitem[Xu et~al.(2024{\natexlab{a}})Xu, Fei, Pan, Liu, Lee, and Hsu]{xu-etal-2024-faithful}
Jundong Xu, Hao Fei, Liangming Pan, Qian Liu, Mong-Li Lee, and Wynne Hsu.
\newblock Faithful logical reasoning via symbolic chain-of-thought.
\newblock In Lun-Wei Ku, Andre Martins, and Vivek Srikumar (eds.), \emph{Proceedings of the 62nd Annual Meeting of the Association for Computational Linguistics (Volume 1: Long Papers)}, pp.\  13326--13365, Bangkok, Thailand, August 2024{\natexlab{a}}. Association for Computational Linguistics.
\newblock \doi{10.18653/v1/2024.acl-long.720}.
\newblock URL \url{https://aclanthology.org/2024.acl-long.720/}.

\bibitem[Xu et~al.(2024{\natexlab{b}})Xu, Fei, Pan, Liu, Lee, and Hsu]{xu2024faithful}
Jundong Xu, Hao Fei, Liangming Pan, Qian Liu, Mong-Li Lee, and Wynne Hsu.
\newblock Faithful logical reasoning via symbolic chain-of-thought.
\newblock In Lun-Wei Ku, Andre Martins, and Vivek Srikumar (eds.), \emph{Proceedings of the 62nd Annual Meeting of the Association for Computational Linguistics (Volume 1: Long Papers)}, pp.\  13326--13365, Bangkok, Thailand, August 2024{\natexlab{b}}. Association for Computational Linguistics.
\newblock \doi{10.18653/v1/2024.acl-long.720}.
\newblock URL \url{https://aclanthology.org/2024.acl-long.720/}.

\bibitem[Yang et~al.(2024)Yang, Zhang, Hui, Gao, Yu, Li, Liu, Tu, Zhou, Lin, et~al.]{yang2024qwen2}
An~Yang, Beichen Zhang, Binyuan Hui, Bofei Gao, Bowen Yu, Chengpeng Li, Dayiheng Liu, Jianhong Tu, Jingren Zhou, Junyang Lin, et~al.
\newblock Qwen2. 5-math technical report: Toward mathematical expert model via self-improvement.
\newblock \emph{arXiv preprint arXiv:2409.12122}, 2024.

\bibitem[Yang et~al.(2025{\natexlab{a}})Yang, Li, Yang, Zhang, Hui, Zheng, Yu, Gao, Huang, Lv, et~al.]{qwen3}
An~Yang, Anfeng Li, Baosong Yang, Beichen Zhang, Binyuan Hui, Bo~Zheng, Bowen Yu, Chang Gao, Chengen Huang, Chenxu Lv, et~al.
\newblock Qwen3 technical report.
\newblock \emph{arXiv preprint arXiv:2505.09388}, 2025{\natexlab{a}}.

\bibitem[Yang et~al.(2025{\natexlab{b}})Yang, Li, Yang, Zhang, Hui, Zheng, Yu, Gao, Huang, Lv, et~al.]{yang2025qwen3}
An~Yang, Anfeng Li, Baosong Yang, Beichen Zhang, Binyuan Hui, Bo~Zheng, Bowen Yu, Chang Gao, Chengen Huang, Chenxu Lv, et~al.
\newblock Qwen3 technical report.
\newblock \emph{arXiv preprint arXiv:2505.09388}, 2025{\natexlab{b}}.

\bibitem[Yang et~al.(2025{\natexlab{c}})Yang, Ma, Lin, and Wei]{yang2025towards}
Wenkai Yang, Shuming Ma, Yankai Lin, and Furu Wei.
\newblock Towards thinking-optimal scaling of test-time compute for llm reasoning.
\newblock \emph{arXiv preprint arXiv:2502.18080}, 2025{\natexlab{c}}.

\bibitem[Yeo et~al.(2025)Yeo, Tong, Niu, Neubig, and Yue]{yeo2025demystifying}
Edward Yeo, Yuxuan Tong, Xinyao Niu, Graham Neubig, and Xiang Yue.
\newblock Demystifying long chain-of-thought reasoning in {LLM}s.
\newblock In \emph{ICLR 2025 Workshop on Navigating and Addressing Data Problems for Foundation Models}, 2025.
\newblock URL \url{https://openreview.net/forum?id=AgtQlhMQ0V}.

\bibitem[Yu et~al.(2025)Yu, Zhang, Zhu, Yuan, Zuo, Yue, Fan, Liu, Liu, Liu, et~al.]{yu2025dapo}
Qiying Yu, Zheng Zhang, Ruofei Zhu, Yufeng Yuan, Xiaochen Zuo, Yu~Yue, Tiantian Fan, Gaohong Liu, Lingjun Liu, Xin Liu, et~al.
\newblock Dapo: An open-source llm reinforcement learning system at scale.
\newblock \emph{arXiv preprint arXiv:2503.14476}, 2025.

\bibitem[Zhang et~al.(2025{\natexlab{a}})Zhang, Mitchell, Ren, Lu, Schwarzer, Pokrass, Zhao, Sanders, Kalai, Passos, Sokolowsky, Le, Ritter, Sheng, Wang, Kostrikov, Lee, Ferstad, Lampe, Radhakrishnan, Fitzgerald, Bubeck, Dubois, Bai, Applebaum, Proehl, Mays, Parish, Liu, Maksin, Ho, Wang, Wang, Watkins, Chao, Miserendino, Patwardhan, Woodford, Hoover, Brill, Stirman, Ajjarapu, Turley, Handa, Godement, Nathan, Huang, Wang, Gohel, Eggers, Yu, Ashley, Huang, Bogan, Sokolova, Horacek, Such, Cohen, Gross, Becker, Wu, Lv, Byron, Liodakis, Johnson, Trpcic, Yesildal, Rygaard, Marsan, Ram-chandani, Kshirsagar, Conlon, Xia, Fu, Narayanan, Choudhry, Kaftan, Creech, Vallone, Duberstein, Sert, Wallace, Zhao, Kofman, Yu, Candela, laine Boyd, Yatbaz, McClay, Wang, Agarwal, Jain, Toizer, Hern{\'a}ndez, Mostovoy, Li, Cha, Wang, Ahmad, Peterson, Chang, Ying, Clark, Stuckey, Tworek, Pachocki, hannes Heidecke, Weil, Fedus, Chen, Altman, and Zaremba]{ZhangOpenAIOS}
Brian Zhang, Eric Mitchell, Hongyu Ren, Kevin Lu, Max Schwarzer, Michelle Pokrass, Shengjia Zhao, Ted Sanders, Adam Kalai, Alexandre Passos, Benjamin Sokolowsky, Elaine~Ya Le, Erik Ritter, Hao Sheng, Hanson Wang, Ilya Kostrikov, James Lee, Johannes Ferstad, Michael Lampe, Prashanth Radhakrishnan, Sean Fitzgerald, S{\'e}bastien Bubeck, Yann Dubois, Yu~Bai, Andy Applebaum, Elizabeth Proehl, Evan Mays, Joel Parish, Kevin Liu, Leon Maksin, Leyton Ho, Miles Wang, Michele Wang, Olivia Watkins, Patrick Chao, Samuel Miserendino, Tejal~A. Patwardhan, Antonia Woodford, Beth Hoover, Jake Brill, Kelly Stirman, Neel Ajjarapu, Nick Turley, Nikunj Handa, Olivier Godement, Akshay Nathan, Alyssa Huang, Andy Wang, Ankit Gohel, Ben Eggers, Brian Yu, Bryan Ashley, Chengdu Huang, Davin Bogan, Emily Sokolova, Eric Horacek, Felipe~Petroski Such, Jonah Cohen, Joshua Gross, Justin Becker, Kan Wu, Larry Lv, Lee Byron, Manoli Liodakis, Max Johnson, Mike Trpcic, Murat Yesildal, Rasmus Rygaard, R.~J. Marsan, Rohit Ram-chandani, Rohan
  Kshirsagar, Sara Conlon, Tony Xia, Siyuan Fu, Srinivas Narayanan, Sulman Choudhry, Tomer Kaftan, Trevor Creech, Andrea Vallone, Andrew Duberstein, Enis Sert, Eric Wallace, Grace Zhao, Irina Kofman, Jieqi Yu, Joaquin~Qui{\~n}onero Candela, Made laine Boyd, Mehmet~Ali Yatbaz, Mike McClay, Mingxuan Wang, Sandhini Agarwal, Saachi Jain, Sam Toizer, Santiago Hern{\'a}ndez, Steve Mostovoy, Tao Li, Young Cha, Yunyun Wang, Lama Ahmad, Troy Peterson, Carpus Chang, Kristen Ying, Aidan Clark, Dane Stuckey, Jerry Tworek, Jakub~W. Pachocki, Jo~hannes Heidecke, Kevin Weil, Liam Fedus, Mark Chen, Sam Altman, and Wojciech Zaremba.
\newblock Openai o3-mini system card.
\newblock In \emph{OpenAI Blog Post}, 2025{\natexlab{a}}.
\newblock URL \url{https://cdn.openai.com/o3-mini-system-card-feb10.pdf}.

\bibitem[Zhang et~al.(2025{\natexlab{b}})Zhang, Wang, Cheng, Zhuang, Lin, Zhang, Wang, Cui, Wang, Peng, et~al.]{zhang2025srpo}
Xiaojiang Zhang, Jinghui Wang, Zifei Cheng, Wenhao Zhuang, Zheng Lin, Minglei Zhang, Shaojie Wang, Yinghan Cui, Chao Wang, Junyi Peng, et~al.
\newblock Srpo: A cross-domain implementation of large-scale reinforcement learning on llm.
\newblock \emph{arXiv preprint arXiv:2504.14286}, 2025{\natexlab{b}}.

\bibitem[{Zhang} et~al.(2025){Zhang}, {Liu}, {Yuan}, {Yuan}, {Gu}, and {Yao}]{2025arXiv250517508Z}
Yifan {Zhang}, Yifeng {Liu}, Huizhuo {Yuan}, Yang {Yuan}, Quanquan {Gu}, and Andrew~C {Yao}.
\newblock {On the Design of KL-Regularized Policy Gradient Algorithms for LLM Reasoning}.
\newblock \emph{arXiv e-prints}, art. arXiv:2505.17508, May 2025.

\bibitem[Zhao et~al.(2025)Zhao, Wu, Yue, Wu, Xu, Lin, Wang, Wu, Zheng, and Huang]{zhao2025absolute}
Andrew Zhao, Yiran Wu, Yang Yue, Tong Wu, Quentin Xu, Matthieu Lin, Shenzhi Wang, Qingyun Wu, Zilong Zheng, and Gao Huang.
\newblock Absolute zero: Reinforced self-play reasoning with zero data.
\newblock \emph{arXiv preprint arXiv:2505.03335}, 2025.

\bibitem[Zhao et~al.(2023)Zhao, Gu, Varma, Luo, Huang, Xu, Wright, Shojanazeri, Ott, Shleifer, Desmaison, Balioglu, Damania, Nguyen, Chauhan, Hao, Mathews, and Li]{pytorchfsdp}
Yanli Zhao, Andrew Gu, Rohan Varma, Liang Luo, Chien-Chin Huang, Min Xu, Less Wright, Hamid Shojanazeri, Myle Ott, Sam Shleifer, Alban Desmaison, Can Balioglu, Pritam Damania, Bernard Nguyen, Geeta Chauhan, Yuchen Hao, Ajit Mathews, and Shen Li.
\newblock Pytorch fsdp: Experiences on scaling fully sharded data parallel.
\newblock \emph{Proc. VLDB Endow.}, 16\penalty0 (12):\penalty0 3848–3860, August 2023.
\newblock ISSN 2150-8097.
\newblock \doi{10.14778/3611540.3611569}.
\newblock URL \url{https://doi.org/10.14778/3611540.3611569}.

\bibitem[Zhong et~al.(2022)Zhong, Wang, Tang, Xu, Guo, Chen, Wang, Yin, Zhou, and Duan]{zhong-etal-2022-analytical}
Wanjun Zhong, Siyuan Wang, Duyu Tang, Zenan Xu, Daya Guo, Yining Chen, Jiahai Wang, Jian Yin, Ming Zhou, and Nan Duan.
\newblock Analytical reasoning of text.
\newblock In Marine Carpuat, Marie-Catherine de~Marneffe, and Ivan~Vladimir Meza~Ruiz (eds.), \emph{Findings of the Association for Computational Linguistics: NAACL 2022}, pp.\  2306--2319, Seattle, United States, July 2022. Association for Computational Linguistics.
\newblock \doi{10.18653/v1/2022.findings-naacl.177}.
\newblock URL \url{https://aclanthology.org/2022.findings-naacl.177/}.

\bibitem[Zhou et~al.(2024)Zhou, Hua, Pan, Cheng, Wu, Yu, and Wang]{zhou2024rulearena}
Ruiwen Zhou, Wenyue Hua, Liangming Pan, Sitao Cheng, Xiaobao Wu, En~Yu, and William~Yang Wang.
\newblock Rulearena: A benchmark for rule-guided reasoning with llms in real-world scenarios.
\newblock \emph{arXiv preprint arXiv:2412.08972}, 2024.

\bibitem[Zhu et~al.(2023)Zhu, Xue, Chen, Zhou, Tang, Schuurmans, and Dai]{zhu2023large}
Zhaocheng Zhu, Yuan Xue, Xinyun Chen, Denny Zhou, Jian Tang, Dale Schuurmans, and Hanjun Dai.
\newblock Large language models can learn rules.
\newblock \emph{arXiv preprint arXiv:2310.07064}, 2023.

\bibitem[Zucchet et~al.(2025)Zucchet, Bornschein, Chan, Lampinen, Pascanu, and De]{zucchet2025language}
Nicolas Zucchet, Jorg Bornschein, Stephanie~C.Y. Chan, Andrew~Kyle Lampinen, Razvan Pascanu, and Soham De.
\newblock How do language models learn facts? dynamics, curricula and hallucinations.
\newblock In \emph{Second Conference on Language Modeling}, 2025.
\newblock URL \url{https://openreview.net/forum?id=vBcGnragkr}.

\bibitem[Zuo et~al.(2025)Zuo, Zhang, Qu, Sheng, Zhu, Qi, Sun, Cui, Ding, and Zhou]{zuo2025ttrl}
Yuxin Zuo, Kaiyan Zhang, Shang Qu, Li~Sheng, Xuekai Zhu, Biqing Qi, Youbang Sun, Ganqu Cui, Ning Ding, and Bowen Zhou.
\newblock Ttrl: Test-time reinforcement learning.
\newblock \emph{arXiv preprint arXiv:2504.16084}, 2025.

\end{thebibliography}
